\documentclass{article}

\usepackage{microtype}
\usepackage{graphicx}
\usepackage{subfigure}
\usepackage{booktabs} 

\usepackage{amsmath}
\usepackage{amssymb}
\usepackage{amsthm}
\usepackage{enumitem}
\newcommand{\vect}[1]{{\rm{\mathbf{#1}}}}
\newtheorem{example}{Example}
\newtheorem{proposition}{Proposition}
\newtheorem{theorem}{Theorem}
\newtheorem{definition}{Definition}

\newcommand{\squishlisttwo}{
 \begin{list}{$\bullet$}
  { \setlength{\itemsep}{5pt}
     \setlength{\parsep}{0pt}
    \setlength{\topsep}{0pt}
    \setlength{\partopsep}{0pt}
    \setlength{\leftmargin}{1em}
    \setlength{\labelwidth}{1.5em}
    \setlength{\labelsep}{0.5em} } }

\newcommand{\squishend}{
  \end{list}  }

\usepackage{hyperref}
\usepackage{xurl}
\usepackage{cleveref}
\crefformat{footnote}{#2\footnotemark[#1]#3}

\usepackage[accepted]{icml2020}

\icmltitlerunning{Collaborative Machine Learning with Incentive-Aware Model Rewards}

\begin{document}

\twocolumn[
\icmltitle{Collaborative Machine Learning with Incentive-Aware Model Rewards}

\icmlsetsymbol{equal}{*}

\begin{icmlauthorlist}
\icmlauthor{Rachael Hwee Ling Sim}{nus}
\icmlauthor{Yehong Zhang}{nus}
\icmlauthor{Mun Choon Chan}{nus}
\icmlauthor{Bryan Kian Hsiang Low}{nus}
\end{icmlauthorlist}

\icmlaffiliation{nus}{Department of Computer Science, National University of Singapore, Republic of Singapore}

\icmlcorrespondingauthor{Bryan Kian Hsiang Low}{lowkh@comp.nus.edu.sg}

\icmlkeywords{Data Valuation, Incentives in Machine Learning}

\vskip 0.3in
]



\printAffiliationsAndNotice{}  

\begin{abstract}
Collaborative \emph{machine learning} (ML) is an appealing paradigm to build high-quality ML models by training on the aggregated data from many parties. However, these parties are only willing to share their data when given enough incentives, such as a guaranteed fair reward based on their contributions. This motivates the need for measuring a party's contribution and designing an incentive-aware reward scheme accordingly. This paper proposes to value a party's 
reward 
based on Shapley value and information gain on model parameters given its data. Subsequently, we give each party a model as a reward. To formally incentivize the collaboration, we define some desirable properties (e.g., fairness and stability) which are inspired by cooperative game theory but adapted for our model reward that is uniquely \emph{freely replicable}. Then, we propose a novel model reward scheme to satisfy 
fairness and trade off between the desirable properties via an adjustable parameter. The value of each party's model reward determined by our scheme is attained by injecting Gaussian noise to the aggregated training data with an optimized noise variance. We empirically demonstrate interesting properties of our scheme and evaluate its performance using synthetic and real-world datasets.
\end{abstract}

\section{Introduction}
\label{sect:intro}
Collaborative \emph{machine learning} (ML) is an appealing paradigm to build high-quality ML models. While an individual party may have limited data,
it is possible to build improved, high-quality ML models by training on the aggregated data from many parties.
For example, in healthcare, a hospital or healthcare firm whose data diversity and quantity are limited due to its small patient base can draw on data from other hospitals and firms to improve the prediction of some disease progression (e.g., diabetes) \cite{share-health}. This collaboration can be encouraged by a government agency, such as the National Institute of Health in the United States.
In precision agriculture, a farmer with limited land area and sensors can combine his collected data with the other farmers to improve the modeling of the effect of various influences (e.g., weather, pest) on his crop yield \cite{share-agriculture}. Such data sharing also benefits other application domains, including real estate in which a property agency
can pool together its limited transactional data with that of the other agencies to improve the prediction of property prices 
\cite{share-real-estate}.

However, any party would have incurred some nontrivial cost to collect its data. So, they would not altruistically donate their data and risk losing their competitive edge. 
These parties will be motivated to share their data when given enough incentives, 
such as a guaranteed benefit from the collaboration and a \emph{fair} higher reward from contributing more valuable data.
To this end, we propose to value each party's contributed data and design an incentive-aware reward scheme to give each party a separate ML model as a reward (in short, \emph{model reward}) accordingly.
We use only model rewards and exclude monetary compensations as (a) in the above-mentioned applications, every party such as a hospital is mainly interested in improving model quality for unlimited future test predictions;
(b) there may not be a feasible and clear source of monetary revenue to compensate participants (e.g., due to restrictions, the government may not be able to pay firms using tax revenues);
and (c) if parties have to pay to participate in the collaboration, then the total payment is debatable and many may lack funding while still having valuable data.

\emph{How then should we value a party's data and its effect on model quality?}
To answer this first question, 
we propose a valuation method based on the \emph{informativeness} of data.
In particular, more informative data induces a greater reduction in the uncertainty of the model parameters, hence improving model quality.
In contrast, existing data valuation methods \cite{ghorbani2019data,ghorbani2020,jia2019efficient,jia2019towards,yoon2020} measure model quality via its validation accuracy, which calls for a
tedious or even impossible process of needing all parties to agree on a common validation dataset.
Inaccurate valuation can also arise due to how a party's test queries, which are likely to change over time, differ from the validation dataset.
Our data valuation method does not make any assumption about the current or future distribution of test queries.

Next, \emph{how should we design a reward scheme to decide the values of model rewards for incentivizing a collaboration?}
Intuitively, a party will be motivated to collaborate if it can receive a better ML model than others who have contributed less valuable data, and than what it can build alone or get from collaborating separately with some parties.
Also, the parties often like to maximize the total benefit from the collaboration.
These incentives appear related to solution concepts (\emph{fairness}, \emph{individual rationality}, \emph{stability},  and \emph{group welfare}) from \emph{cooperative game theory} (CGT),  respectively.
However, as CGT assumptions are restrictive for the uniquely \emph{freely replicable}\footnote{Data and model reward, like digital goods, can be replicated at zero marginal cost and given to more parties.} nature of our model reward, these CGT concepts have to be adapted for defining incentives for our model reward.
We then design a novel model reward scheme to provide these incentives.

Finally, \emph{how can we realize the values of the model rewards decided by our scheme?}
How should we modify the model rewards or the data used to train them?
An obvious approach to control the values of the model rewards is to select and only train on subsets of the aggregated data. However, this requires considering an exponential number of \emph{discrete} subsets of data, which is intractable for large datasets such as medical records.
We avoid this issue by injecting noise into the aggregated data from multiple parties instead.
The value of each party's model reward can then be realized by simply optimizing the \emph{continuous} noise variance parameter.

The specific contributions of our work here include:
\squishlisttwo
	\item Proposing a data valuation method using the \emph{information gain} (IG) on model parameters given the data (Sec.~\ref{MI});

	\item Defining new conditions for incentives (i.e., \emph{Shapley fairness}, \emph{stability}, \emph{individual rationality}, and \emph{group welfare}) that are suitable for the \emph{freely replicable} nature of our model reward (Sec.~\ref{incentives}). 
	As these incentives cannot be provided all at the same time, we design a novel model reward scheme with an adjustable parameter to trade off between them while maintaining fairness (Sec.~\ref{Tying});

	\item Injecting Gaussian noise into the aggregated data from multiple parties and optimizing the noise variance parameter for realizing the values of the model rewards decided by our scheme (Sec.~\ref{noise}); and

	\item Demonstrating interesting properties of our model reward scheme empirically and evaluating its performance with synthetic and real-world datasets (Sec.~\ref{experi}).
\squishend
To the best of our knowledge, our work here is the first to propose a collaborative ML scheme that formally considers incentives
beyond fairness and relies solely on model rewards to  
realize them.
Existing works \cite{jia2019efficient,jia2019towards,ohrimenko2019collaborative} have only looked at fairness and have to resort to monetary compensations if considered.
\section{Problem Formulation}
\label{problem}
In our problem setting, we consider $n$ honest and non-malicious parties, each of whom owns some data and assume the availability of a trusted \emph{central party}\footnote{In reality, such a central party can be found in established data sharing platforms like Ocean Protocol \cite{ocean-protocol} and Data Republic (\url{https://datarepublic.com}).}
who aggregates data from these parties, measures the value of their data, and distributes a resulting trained ML model to each party.
We first introduce the notations and terminologies used in this work: Let 
$N \triangleq \{1, \ldots, n\}$ denote a set of $n$ parties.
Any subset $C \subseteq N$ is called a \emph{coalition} of parties.
The \emph{grand coalition} is the set $N$ of all parties.
Parties will team up and partition themselves into a \emph{coalition structure} $CS$. Formally, $CS$ is a set of coalitions such that $\bigcup_{C \in CS} C = N$ and $C \cap C' = \emptyset$ for any $C, C' \in CS$ and $C \neq C'$.
The data of party $i\in N$ is represented by $D_i \triangleq (\vect{X}_i, \vect{y}_i)$ where $\vect{X}_i$ and $\vect{y}_i$ are the input matrix and output vector, respectively.
Let $v_C$ denote the \emph{value of the (aggregated) data} $D_C \triangleq \{D_i\}_{i \in C}$ of any coalition $C \subseteq N$. We use $v_i$ to represent $v_{\{i\}}$ to ease  notation. For each party $i \in N$,  $r_i$ denotes the \emph{value of its received model reward}.

The objective is to design a collaborative ML scheme for the central party to decide and realize the values $(r_i)_{i \in N}$ of model rewards distributed to parties $1, \ldots, n$. The scheme should satisfy certain incentives (e.g., fairness and stability) to encourage the collaboration.
Some works \cite{ghorbani2019data,jia2019efficient,jia2019towards} have considered similar problems and can fairly \emph{partition} the (monetary) value $v_N$ of the entire aggregated data into $r_i$ for $i \in N$. 
They achieved this using results from \emph{cooperative game theory} (CGT).
%
Our problem, however, differs and cannot be addressed directly using CGT. This is due to the \emph{freely replicable} nature of our \emph{model reward} 
-- the total value $\sum_{i \in N}r_i$ of received model rewards 
over all parties $i \in N$ can exceed $v_N$.

We will next show how to assess the value of data (Sec.~\ref{MI}) and, more importantly, how to design the reward scheme to decide the values $(r_i)_{i \in N}$ of model rewards accordingly and realize these values for achieving our incentive-aware objective (Sec.~\ref{reward}).
\section{Data Valuation with Information Gain}\label{MI}
A set $D_C$ of data is considered more valuable (i.e., higher $v_C$) if it can be used to train a higher-quality (hence more valuable) ML model.
Existing data valuation methods \cite{ghorbani2019data,ghorbani2020,jia2019towards,yoon2020} measure the quality of a trained ML model via its validation accuracy.
However, these methods require the central party to carefully select a validation dataset that all parties must agree on.
This is often a tedious or even impossible process, especially if every party's test queries, which are likely to change over time, differ from the validation dataset.\footnote{In the unlikely event that all parties can agree on a common validation dataset, validation accuracy would be the preferred measure of model quality and hence data valuation method.}
For example, two private hospitals $\mathcal{H}_1$ and $\mathcal{H}_2$ aggregate their data for diabetes prediction. 
$\mathcal{H}_1$ and $\mathcal{H}_2$ prefer accurate test predictions for female and young patients, respectively.
Due to the differences in their data sizes, data qualities, and preferred test queries, it is difficult for the central party to decide the demographics of the patients in the validation dataset such that the data valuation is unbiased and accurate.

To circumvent the above-mentioned issues, our proposed data valuation method instead considers an information-theoretic measure of the quality of a trained model in terms of the reduction in uncertainty of the model parameters, denoted by vector $\boldsymbol{\theta}$, after training on data $D_C$.
We use the prior entropy $\mathbb{H}(\boldsymbol{\theta})$ and posterior entropy $\mathbb{H}(\boldsymbol{\theta}| D_C)$ to represent the uncertainty of $\boldsymbol{\theta}$ before and after training on $D_C$, respectively.
So, if the data $D_C$ for $C \subseteq N$ can induce a greater reduction in the uncertainty/entropy of $\boldsymbol{\theta}$ or, equivalently, \emph{information gain} (IG) $\mathbb{I}(\boldsymbol{\theta}; D_C)$ on $\boldsymbol{\theta}$:
\begin{equation}\label{data_IG}
v_C \triangleq \mathbb{I}(\boldsymbol{\theta}; D_C) = \mathbb{H}(\boldsymbol{\theta}) - \mathbb{H}(\boldsymbol{\theta}| D_C)\ ,
\end{equation}
then a higher-quality (hence more valuable) model can be trained using this more valuable/informative data $D_C$.
IG is an appropriate data valuation method as it is often used as a surrogate measure of the test/predictive accuracy of a trained model
\cite{krause2007,kruschke2008} since the test queries are usually not known \emph{a priori}.
We empirically demonstrate such a surrogate effect in Appendix~\ref{surrogate}.
The predictive distribution of the output $y^*$ at the test input $\vect{x}^*$ given data $D_C$ is calculated by averaging over all possible model parameters $\boldsymbol{\theta}$ weighted by their posterior belief $p(\boldsymbol{\theta}|D_C)$, i.e., 
$
p(y^*|\vect{x}^*, D_C) = \int p(y^*|\vect{x}^*, \boldsymbol{\theta})\ p(\boldsymbol{\theta}|D_C) \ \text{d}\boldsymbol{\theta}.
$
By reducing the uncertainty in $\boldsymbol{\theta}$, we can further rule out models that are unlikely given $D_C$ and place higher weights (i.e.,  by increasing $p(\boldsymbol{\theta}|D_C)$) on models that are closer to the true model parameters, thus improving the predictive accuracy for any 
test query
in expectation. 
The value $v_C$~\eqref{data_IG} of data $D_C$ has the following properties: 
\squishlisttwo
	\item Data of an \emph{empty} coalition has \emph{no} value:
	$v_{\emptyset} = 0\ .$
	\item Data of any coalition $C \subseteq N$ has \emph{non-negative} value:
	$\forall C\subseteq N\ \ v_C \geq 0 \ .$
	\item \textbf{Monotonicity}. Adding more parties to a coalition cannot decrease the value of its data: $\forall C\subseteq C'\subseteq N\ \ v_{C'} \geq v_C\ .$
	\item \textbf{Submodularity}. Data of any party $i$ is less valuable to a larger coalition which has more parties and data:\vspace{-1.7mm}
\squishend
	$\forall i \in N \ \ \forall C\subseteq C'\subseteq N \setminus \{i\}\ \ v_{C' \cup \{i\}} - v_{C'} \leq v_{C \cup \{i\}} - v_C\ .$
	
The second and third properties are due to the ``information never hurts" bound for entropy \cite{Cover91}. 
The submodular property of IG~\eqref{data_IG} assumes conditional independence of the data $D_i$ and $D_j$ given $\boldsymbol{\theta}$ for any $i, j \in N$ and $i \neq j$; its proof is in Appendix~\ref{a.IG}.

The first two properties fulfill standard assumptions of CGT. The latter two properties will influence the design and properties of our model reward scheme in Sec.~\ref{reward}. For example, the monotonic property ensures that the value $r_i$ of any party $i$'s model reward is never negative (Sec.~\ref{reward}).
%
%
\section{Incentive-Aware Reward Scheme with Model Rewards}
\label{reward}
Recall our problem setting in  Sec.~\ref{problem} that the central party will train a model for each party as a reward. The reward should be decided based on the value of each party's 
data relative to that of the other parties'. 
Let $D_i^{r}$ be the data (i.e., derived from $D_N$) used to train party $i$'s model reward. The value of party $i$'s model reward is $r_i \triangleq \mathbb{I}(\boldsymbol{\theta}; D_i^{r})$ according to Sec.~\ref{MI}.
In this section, we will first present the incentives to encourage collaboration (Sec.~\ref{incentives}), then describe how our incentive-aware reward scheme will satisfy them (Sec.~\ref{Tying}), and finally show how  to vary $D_i^{r}$ to realize the values of the model rewards decided by our scheme (Sec.~\ref{noise}).

We desire a collaboration involving the \emph{grand coalition} (i.e., $CS = \{N\}$) as it results in the largest aggregated data $D_N$ and hence allows the highest value of model reward to be achieved, which  eliminates the tedious process of deciding how the parties should team up and partition themselves.
In Sec.~\ref{addition}, we will discuss how to incentivize the grand coalition formation and increase the total benefit from the collaboration.
\subsection{Incentives} \label{incentives}
Our reward scheme has to be valid, fair to all the parties, and guarantee an improved model for each party.
To achieve these, we exploit and adapt key assumptions and constraints about rewards in CGT. 
As has been mentioned in Sec.~\ref{problem}, the modifications are necessary 
as the model reward 
is uniquely freely replicable.
We require the following incentive conditions to hold for the values $(r_i)_{i \in N}$ of model rewards based on the chosen coalition structure $CS$: 
\begin{enumerate}[label=R\arabic*,leftmargin=*,itemsep=5pt,topsep=0pt,parsep=0pt,partopsep=0pt]
    \item \label{R1} \textbf{Non-negativity.} $\forall i \in N\ \ r_i \geq 0\ .$
    
    \item \label{feasibility} \textbf{Feasibility.} The model reward received by each party in any coalition $C\in CS$ cannot be more valuable than the model 
    trained on their aggregated data $D_C$:
    ${\forall C \in CS}\ \ {\forall i \in C}\ \ {r_i \leq v_{C}}\ .$

    \item \label{weak-efficiency} \textbf{Weak Efficiency.} In each coalition $C\in CS$, the model reward received by at least a party $i \in C$ is as valuable as the model 
    trained on the aggregated data $D_C$ of $C$:
    ${\forall C \in CS}\ \ {\exists i \in C}\ \ {r_i = v_{C}}\ .$

    \item \label{rational} \textbf{Individual Rationality.} Each party should receive a model reward that is at least as valuable as the model trained on its own data:
    $\forall i \in N\ \ r_i \geq v_i\ . $
\end{enumerate}
\ref{R1} and \ref{rational} are the same as the solution concepts in CGT while \ref{feasibility} and \ref{weak-efficiency} have been adapted.\footnote{\ref{feasibility} and \ref{weak-efficiency} are, respectively, adapted from ${\forall C \in CS}\ {\sum_{i \in C}{r_i} \leq v_{C}}$ and ${\forall C \in CS}\ {\sum_{i \in C}{r_i} = v_{C}}$ in CGT~\cite{coop-game-theory}.}
When $CS = \{N\}$ (i.e., grand coalition), \ref{feasibility} and \ref{weak-efficiency} become ${\forall i \in N}\ \ {r_i \leq v_{N}}$ and ${\exists i \in N}\ \ {r_i = v_{N}}$,  respectively. So, each party cannot receive a more valuable model reward than the model trained on $D_N$ as it would involve creating data artificially.
\subsubsection{Fairness}
In addition, when $CS = \{N\}$ (i.e., grand coalition), to guarantee that the reward scheme is \emph{fair} to all $n$ parties, the values $(r_i)_{i \in N}$ of their model rewards must satisfy the following properties which are inspired by the fairness concepts in CGT \cite{masc-kernel,shapley_value,Young1985} and have also been experimentally studied from a normative perspective \cite{declippel2013}: 
 \begin{enumerate}[label=F\arabic*,leftmargin=*,itemsep=5pt,topsep=0pt,parsep=0pt,partopsep=0pt]
	\item\label{FAIR1} \textbf{Uselessness.}\footnote{\label{C12ref}These properties are axioms of Shapley Value \cite{shapley_value} and have been widely used in existing ML works \cite{ghorbani2019data,jia2019towards,ohrimenko2019collaborative} for data valuation.} 
	If including the data of party $i$ does not improve the quality of a model trained on the aggregated data of any coalition (e.g., when $D_i = \emptyset$), 
	then party $i$  should receive a valueless model reward: For all $i\in N$,
	\[(\forall C\subseteq N \setminus \{i\}\ \ v_{C \cup \{i\}} = v_C) \Rightarrow r_i = 0 \ .\]

	\item\label{FAIR2} \textbf{Symmetry.}\textsuperscript{\ref{C12ref}} 
	If  
	including the data of party $i$ 
	yields the same improvement as that of party $j$ in the quality of a model trained on the aggregated data of any coalition  (e.g., when $D_i = D_j$), then they should receive  equally valuable model rewards: For all $i,j\in N$ s.t.~$i \neq j$,
	\[ (\forall C \subseteq N\setminus\{i,j\}\ \ v_{C \cup \{i\}} = v_{C \cup \{j\}}) \Rightarrow r_i = r_j \ .\]

    \item\label{FAIR3} \textbf{Strict Desirability~\cite{masc-kernel}.} 
    If the quality of a model trained on the aggregated data of at least a coalition improves more by including the data of party $i$ than that of party $j$, but the reverse is not true,
    then party $i$  should receive a more valuable model reward than party $j$: For all $i,j\in N$ s.t.~$i \neq j$,
	\[
	\begin{array}{l} 
	(\exists B \subseteq N\setminus\{i,j\}\ \ v_{B \cup \{i\}} > v_{B \cup \{j\}})\ \land \vspace{1mm}\\
     (\forall C \subseteq N\setminus\{i,j\}\ \ v_{C \cup \{i\}} \geq v_{C \cup \{j\}})	
	\Rightarrow r_i > r_j\  . 
	\end{array} 
	\]

 	\item\label{FAIR4} \textbf{Strict Monotonicity.}\footnote{Our definition is similar to coalitional monotonicity 
	in~\cite{Young1985} except that we consider $>$ instead of $\geq$. This rules out the scenario where the value of party $i$'s data improves but not that of its model reward.
	We further check the feasibility of  improvement in the value of its model reward.} 
	If the quality of a model trained on the aggregated data of at least a coalition containing party $i$ improves (e.g., by including more data of party $i$), \emph{ceteris paribus},
    then party $i$  should receive a more valuable model reward than before:  
	%
	Let $\{v_C\}_{C \in 2^N}$ and $\{v'_C\}_{C \in 2^N}$ denote any two sets of values of data over all coalitions $C\subseteq N$,
	and $r_i$ and $r'_i$ be the corresponding values of model rewards received by party $i$. For all $i\in N$,
	\[
	\begin{array}{l} 
	(\exists B \subseteq  N\setminus\{i\}\ \ v'_{B \cup \{i\}} > v_{B \cup \{i\}})
	\ \land \vspace{1mm}\\
	(\forall C \subseteq  N\setminus\{i\}\ \ v'_{C \cup \{i\}} \geq v_{C \cup \{i\}})
	\ \land \vspace{1mm}\\
	(\forall A \subseteq  N\setminus\{i\}\ \ v'_{A} = v_{A}) 
\land (v'_N > r_i) \Rightarrow r'_i > r_i\ . 
	\end{array} 
	\]
	%
\end{enumerate}
We have the following incentive condition: 
\begin{enumerate}[resume,label=R\arabic*,leftmargin=*,itemsep=5pt,topsep=0pt,parsep=0pt,partopsep=0pt]
	\item\label{min-fairness} \textbf{Fairness.} The values $(r_i)_{i \in N}$ of model rewards must satisfy \ref{FAIR1} to \ref{FAIR4}.
\end{enumerate}

Both \ref{FAIR3} and \ref{FAIR4} imply that marginal contribution ($v_{C \cup \{i\}} - v_{C}$) matters.
\ref{FAIR4} is the only property guaranteeing that if party $i$ adds more valuable/informative data to a coalition containing it, \emph{ceteris paribus},  then it should receive a more valuable model reward than before. 
Additionally,~\ref{FAIR3} establishes a relationship between  parties $i$ and $j$ that if their marginal contributions only differ for coalition $C$ (i.e., $v_{C \cup \{i\}} > v_{C \cup \{j\}}$ w.l.o.g.), then party $i$ should receive a more valuable model reward than party $j$. 

To illustrate their significance, we consider two simpler reward schemes where we directly set the value of model reward received by every party $i\in N$ as (a) the value of its data (i.e., $r_i \triangleq v_i$) or 
(b) the decrease in the value of its data if it leaves the grand coalition 
(i.e., $r_i \triangleq v_{N} - v_{N \setminus \{i\}}$).
Both schemes violate \ref{FAIR3} and \ref{FAIR4} as they ignore the values of the other parties' data: The value of model reward received by party $i$ does not change when its marginal contribution to any non-empty coalition $C \subset N \setminus \{i\}$ increases.
Intuitively, a party with a small value $v_i$ of data (e.g., few data points) can be highly valuable to the other parties if its data is distinct. Conversely, a party with a high value $v_j$ of data should be less valuable to the other parties with similar data as its marginal contribution is lower.

Hence, we consider the Shapley value which captures the idea of marginal contribution precisely as it uses the expected marginal contribution of a party $i$ when it joins the parties preceding it in any permutation:
\begin{equation}\label{shapley}
\text{Shapley}_v(i) = \frac{1}{n!} \sum_{\pi \in \Pi_N} {\left(v_{{S_{\pi,i}} \cup \{i\}} - v_{S_{\pi,i}}\right)}
\end{equation}
where $\Pi_N$ is the set of all possible permutations of $N$ and $S_{\pi,i}$ is the coalition of parties preceding $i$ in permutation $\pi$.
\begin{proposition}\label{p1}
	$r_i = \text{Shapley}_v(i)$ for all $i \in N$ satisfy \ref{min-fairness}.
\end{proposition}
Its proof is in Appendix~\ref{a.p1}.
A party will have a larger Shapley value and hence value $r_i$ of  model reward when its data is highly valuable on its own (e.g., with low inherent noise) and/or to the other parties (e.g., due to low correlation).

\textbf{Fairness with Weak Efficiency (\ref{weak-efficiency}).}
We simplify the notation of $\text{Shapley}_v(i)$ to $\phi_i$.
Such a reward scheme may not satisfy \ref{weak-efficiency} as the total value $\sum_{i \in N}r_i$ of model rewards is only $v_N$~\cite{coop-game-theory}.
Due to the freely replicable nature of our model reward, we want the total value to exceed $v_N$ and the value of some party's model reward to be $v_N$.
To satisfy other incentive conditions, we consider a function $g$ to map $(\phi_i)_{i \in N}$ to $(r_i)_{i \in N}$ (i.e., $r_i \triangleq g(\phi_i)$). 
To achieve fairness in \ref{min-fairness}, $g$ must be strictly increasing 
with $g(0) = 0$.
These motivate us to propose the following:
\begin{definition}[\textbf{Shapley Fairness}] \label{shapley-fair}
Given $\{v_C\}_{C \in 2^N}$, if there exists a constant $k > 0$ s.t.~$r_i = k\phi_i$ for all $i \in N$, then the values $(r_i)_{i \in N}$ of model rewards are Shapley fair.
\end{definition}
Note that CGT sets $k = 1$ (Proposition~\ref{p1}). Also, we can control constant $k$ to satisfy other incentive conditions such as~\ref{weak-efficiency}.
A consequence of Definition~\ref{shapley-fair} is that $\phi_i/\phi_j = r_i/r_j$.
So, increasing the ratio of expected marginal contributions $\phi_i$ of party $i$ vs.~$\phi_j$ of party $j$
(e.g., by including more valuable/informative data of party $i$, \emph{ceteris paribus}) 
results in the same increase in the ratio of
values of model rewards $r_i$ received by party $i$ vs.~$r_j$ received by party $j$. 

\textbf{Fairness with Individual Rationality (\ref{rational}).} However, another issue persists in the above modified definition of fairness (Definition~\ref{shapley-fair}): 
\ref{rational} may not be satisfied 
when the value of data is submodular (e.g., IG). We show an example below and leave the detailed discussion to Appendix~\ref{violate}:
\begin{example}
Suppose that there is a coalition of two parties whose values of data are $v_1 = 7$, $v_2 = 5$, and $v_{\{1, 2\}}=8$. Using \eqref{shapley}, the Shapley values of the two parties are $\phi_1 = 5$ and $\phi_2 = 3$. To satisfy \ref{weak-efficiency}, we set $r_1 = 8$. Then, to achieve Shapley fairness, $r_2 = (8/5)\times 3 = 4.8$. Since $r_2 < v_2$,  \ref{rational} is not satisfied.
\end{example}
Since our model reward is freely replicable, it is possible to give all parties in a larger coalition more valuable model rewards.
%
How then should we redefine $g$ to derive larger values $r_i$ of model rewards to achieve \ref{rational} while still satisfying \ref{min-fairness} and \ref{weak-efficiency} (i.e., $g(\max_{i \in N}\phi_i) = v_N$)? To answer this question,  we further modify the above definition of Shapley fairness (Definition~\ref{shapley-fair}) to the following:
%
\begin{definition}[$\rho$\textbf{-Shapley Fairness}] \label{ratio-fairness}
Given $\{v_C\}_{C \in 2^N}$, if there exist constants $k > 0$ and $\rho > 0$ s.t.~$r_i = k \phi_i^\rho$ for all $i \in N$, then the values $(r_i)_{i \in N}$ of model rewards are $\rho$-Shapley fair.
\end{definition}
It follows from Definition~\ref{ratio-fairness} that when $\rho < 1$, $\phi_i > \phi_j$ implies $\phi_i/\phi_j> r_i/r_j>1$.
Furthermore,
reducing $\rho$ from $1$ ``weakens'' proportionality of $r_i$ to $\phi_i$ and decreases the ratio of $r_i$ vs.~$r_j$ while preserving $r_i> r_j$.
So, if the value of a party's model reward is $v_N$ (hence satisfying \ref{weak-efficiency}), then the values of the other parties' model rewards will become closer to $v_N$, thereby potentially satisfying \ref{rational}.  
We will discuss the choice of $k$ and the effect of $\rho$ in Sec.~\ref{Tying}.
%
\subsubsection{Stability and Group Welfare}\label{addition}
At the beginning of Sec.~\ref{reward}, we have provided justifications for desiring the grand coalition. We will now introduce two other solution concepts in CGT (i.e., stability and group welfare) to incentivize its formation and increase the total benefit from the collaboration, respectively. To the best of our knowledge, these solution concepts have not been considered in existing collaborative ML works.

A coalition structure $CS$ with a given set $(r_i)_{i \in N}$ of values of model rewards is said to be \emph{stable} if no subset of parties has a common incentive to abandon it to form another coalition on their own. With stability, parties can be assured that the collaboration will not fall apart and they cannot get a more valuable model reward from adding or removing parties.
Similar to  
other solution concepts in CGT introduced previously, the definition of stability in CGT\footnote{In CGT, a coalition structure $CS$ with given $(r_i)_{i \in N}$ is \emph{stable} if $\forall C \subseteq N\ \ v_C \leq \sum_{i \in C} r_i\ .$
} does not suit the freely replicable nature of our model reward and the redefined incentive condition on feasibility \ref{feasibility}.
Therefore, we propose the following new definition of stability:
\begin{definition}[\textbf{Stability}]\label{stable} A coalition structure $CS$ with a given set $(r_i)_{i \in N}$ of values of model rewards is stable if $\forall C \subseteq N\ \ \exists i \in C\ \ {v_C \leq r_i}\ .$
\end{definition}
Conversely, supposing $\exists C \subseteq N\ \ \forall i \in C\ \ r_i < v_C$, all parties in $C$ may be willing to deviate to form coalition $C$ as they can feasibly increase the values of their model rewards (up) to $v_C$.
The condition in Definition~\ref{stable} is computationally costly to check as it involves an exponential (in $n$) number of constraints. To ease computation and since we are mainly interested in the grand coalition (i.e., $CS = \{N\}$), we look at the following sufficient condition instead:
\begin{enumerate}[resume,label=R\arabic*,leftmargin=*,itemsep=5pt,topsep=0pt,parsep=0pt,partopsep=0pt]
	\item \label{stability} \textbf{Stability of  Grand Coalition.} Suppose that the value of data is monotonic. The grand coalition is stable if for every coalition $C$, the value of the model reward received by the party with largest Shapley value is at least $v_C$:\\
	$
	\forall C \subseteq N\ \ \forall i \in C\ \ \phi_i = \max_{j \in C} \phi_j \Rightarrow\ v_C \leq r_i \ .
	$
\end{enumerate}
The values $(r_i)_{i \in N}$ of model rewards that satisfy \ref{stability} will also satisfy Definition~\ref{stable}. 
To ensure \ref{stability}, for each party $i \in N$, we will only need to check one constraint involving $r_i$: $v_{C_i} \leq r_i$ where $C_i$ includes all parties whose Shapley value is 
at most $\phi_i$. The monotonic property of the value of data guarantees that $v_{C} \leq v_{C_i} \leq r_i$ for any $C \subseteq C_i$. The total number of constraints is linear in the number $n$ of parties.
Unlike \ref{R1} to \ref{min-fairness}, \ref{stability} is an optional incentive condition. 

The last incentive condition stated below does not have to be \emph{fully} provided:
\begin{enumerate}[resume,label=R\arabic*,leftmargin=*,itemsep=5pt,topsep=0pt,parsep=0pt,partopsep=0pt]
    \item \label{group-welfare} \textbf{Group Welfare.} The values $(r_i)_{i \in N}$ of model rewards should maximize the group welfare $\sum_{i \in N} r_i$. 
\end{enumerate}
\subsection{Reward Scheme Considering All Incentives}\label{Tying}
In this subsection, we will present a reward scheme which considers all the incentives in Sec.~\ref{incentives} and assumes that the grand coalition will form. 
It uses parameter $\rho$ in Definition~\ref{ratio-fairness} to trade off between achieving Shapley fairness vs.~satisfying the other incentives, as detailed below:
\begin{theorem} \label{mi-power}
Let $0 < \rho \leq 1$.
For each party $i \in N$, let $\phi_i \triangleq \text{Shapley}_v(i)$
and reward $ {r_i \triangleq {{({\phi_i}/{\phi^*} ) }^\rho} \times v_N}$ where $\phi^* = \max_{i \in N}\phi_i$.\footnote{In practice, $\phi_i$ can be based on other  solution concepts in CGT that satisfy \ref{FAIR1} to \ref{FAIR4}.
}
The values $(r_i)_{i \in N}$ of model rewards are $\rho$-Shapley fair and satisfy \ref{R1} to \ref{weak-efficiency} and \ref{min-fairness} when $\rho > 0$ . Also, when
\squishlisttwo
    \item $\rho = 1$, $(r_i)_{i \in N}$ are (pure) Shapley fair (Definition~\ref{shapley-fair});
    \item $\rho \leq \rho_r \triangleq \min_{i \in N}{{\log(v_{i}/v_N)}/{\log(\phi_i/\phi^*)}}$, $(r_i)_{i \in N}$ satisfy individual rationality (\ref{rational});
    \item $\rho \leq \rho_s \triangleq \min_{i \in N}{{\log(v_{C_i}/v_N)}/{\log(\phi_i/\phi^*)}}$ where coalition $C_i \triangleq \{j\in N\mid \phi_j \leq \phi_i\}$, $(r_i)_{i \in N}$ achieve stability of the grand coalition (\ref{stability}) and individual rationality (\ref{rational}) as $\rho_s \leq \rho_r$;
    \item $\rho = 0$, $(r_i)_{i \in N}$ provide maximum group welfare (\ref{group-welfare}) but do not satisfy fairness (\ref{min-fairness}). 
\squishend
\end{theorem}
Its proof 
is in Appendix~\ref{a.power}.
As $\rho$ decreases from 1, the values $(r_i)_{i \in N}$ of model rewards deviate further from pure Shapley fairness (i.e., less proportional to $(\phi_i)_{i \in N}$) and the value $r_i$ of any party $i$'s model reward with $0 < \phi_i/\phi^* < 1$ will increase. This increases group welfare (\ref{group-welfare}) and the values $(r_i)_{i \in N}$ of model rewards can potentially satisfy individual rationality (\ref{rational}) and stability (\ref{stability}). Thus, a smaller $\rho$ addresses the limitations of pure Shapley fairness.
We do not consider (a) $\rho > 1$ as it reduces group welfare and is not needed to satisfy other incentives, nor (b) $\rho < 0$ as it demands that 
more valuable/informative data and higher Shapley value lead to less valuable model reward, hence not satisfying fairness (\ref{min-fairness}). 

\begin{figure}
    \includegraphics[scale=0.13]{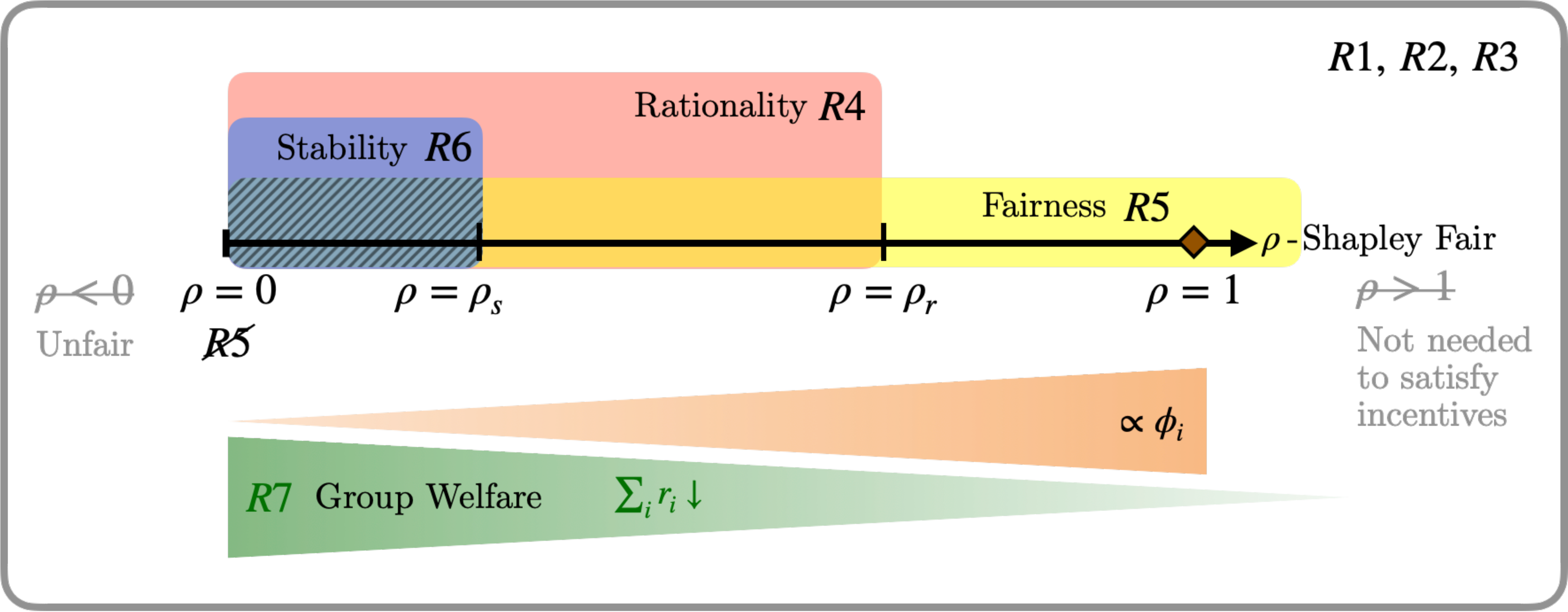} \vspace{-3mm}
    \caption{An overview of the incentive conditions and how they are satisfied by our reward scheme in Theorem~\ref{mi-power}. There may be alternative desirable reward schemes not using Definition~\ref{ratio-fairness}. Note that in some scenarios, $1$ may be less than $\rho_r$ or $\rho_s$. As a result, 
    the values $(r_i)_{i \in N}$ of model rewards that are purely Shapley fair 
    naturally satisfy individual rationality and stability, respectively.} 
    \label{overview}\vspace{-3.2mm}
\end{figure}

Fig.~\ref{overview} gives an overview of the incentive conditions and how Theorem~\ref{mi-power} uses $\rho$ to satisfy them and trade off between the various incentives. Note that \ref{stability} guarantees \ref{rational}. Ideally, we want the values $(r_i)_{i \in N}$ of model rewards to satisfy fairness (\ref{min-fairness}), stability (\ref{stability}), and individual rationality (\ref{rational}) (i.e., shaded region in Fig.~\ref{overview}). However, the 
values $(r_i)_{i \in N}$ of model rewards that are purely Shapley fair 
may not always lie in the desired shaded region. So, a smaller $\rho$ is needed. 
%
%

Consider how much each party $i$ benefits from the collaboration by choosing a $\rho$ smaller than $1$, specifically, by measuring the ratio of the values $r_i$ of its received model reward that are $\rho$-Shapley fair ($\rho< 1$) vs.~purely Shapley fair ($\rho=1$). 
Such a ratio can be evaluated to  $(\phi^*/\phi_i)^{1-\rho}\geq 1$
and is smaller (larger) for a party $i$ with a larger (smaller) $\phi_i$.
So, when a smaller $\rho$  is chosen, a party $i$ with $\phi_i$ close to $\phi^*$ needs to be ``altruistic"  as it cannot benefit as much due to a smaller ratio than any party $j$ with a smaller $\phi_j$.  However, note that party $i$ is already getting close to the maximum value $v_N$ of model reward. Regardless of the chosen $\rho$, the party $i$ with the largest $\phi_i$ (i.e., $\phi^*$) always receives the most valuable model reward with the highest value $v_N$.
In practice, the parties may agree to use a smaller $\rho$ if they
want to 
increase their total benefit from the collaboration
or if they do not know their relative expected marginal contributions beforehand. 
%
\subsection{Realization of Model Rewards} \label{noise}
Finally, we will have to train a model for each party as a reward to realize the values $(r^*_i)_{i \in N}$ of model rewards decided by the above reward scheme.
Recall from the beginning of Sec.~\ref{reward} that the value $r_i = \mathbb{I}(\boldsymbol{\theta}; D^r_i)$ of model reward received by each party $i$ is the IG on model parameters $\theta$ given the data $D^r_i$.
How is the training data $D^r_i$ for each party $i$ selected to realize $r_i = r^*_i$?

A direct approach is to select and only train on a subset of the aggregated data from all parties: $D^r_i \subseteq D_N$. However, this is infeasible due to the need to consider an \emph{exponential} number of \emph{discrete} subsets of data, all of which may not be able to realize the decided value $r^*_i$ of model reward exactly.


To avoid the above issue, we instead consider injecting Gaussian noise $\mathcal{N}(\vect{0},\eta_i\vect{I})$ into the aggregated data 
from multiple parties 
and optimizing the  \emph{continuous} noise variance parameter $\eta_i$
for realizing the decided value $r^*_i$ of model reward.
Specifically, the central party collects data $D_i \triangleq (\vect{X}_i, \vect{y}_i)$ from every party $i \in N$ and constructs $D^r_i$ 
by concatenating the input matrices $\vect{X}_i$ for $i\in N$ and the output vector $\vect{y}_i$ with $\vect{z}_i \triangleq 
(\vect{y}_j)_{j \in N\setminus \{i\}}
+ \mathcal{N}(\vect{0},\eta_i\vect{I})$.
When training a model for party $i$, we use the original $\vect{y}_i$ so that each party gets all the information from its own data $D_i$ and cannot improve its received model reward by subsequently training on it. We inject noise with variance $\eta_i$ only to the other parties' data (i.e., $(\vect{y}_j)_{j \in N\setminus \{i\}}$) to reduce the information from parties $N \setminus \{i\}$ conveyed to party $i$. 
In particular, $r_i = v_N$ when $\eta_i = 0$ and $r_i = v_i$ when $\eta_i = \infty$. By varying $\eta_i$, we can span different values of $r_i$. We use an efficient root-finding algorithm to find the optimal $\eta_i$ such that $r_i = r^*_i$ for each party $i\in N$.
The effect of adding noise to the data will be reflected in the variance of party $i$'s model reward parameters and its predictive distribution. We will show that the added noise affects predictive accuracy reasonably in Sec.~\ref{experi}.
\section{Experiments and Discussion}\label{experi}
This section empirically evaluates the performance  and properties of our reward scheme (Sec.~\ref{Tying}) on Bayesian regression models
with the
(a) synthetic Friedman dataset with $6$ input features \cite{friedman},
(b) \emph{diabetes progression} (DiaP) dataset
on the diabetes progression of $442$ patients
with $9$ 
input features 
\cite{diabetes-dataset}, and
(c) \emph{Californian housing} (CaliH) dataset 
on the value\ of $20640$ houses
with $8$ input features   \cite{cali-dataset}.
We use a Gaussian likelihood and assume that the model hyperparameters 
are known or learned using maximum likelihood estimation.

The performance of our reward scheme is evaluated using the 
IG and \emph{mean negative log probability} (MNLP) metrics:\vspace{-0.4mm}
\begin{equation}
\hspace{-1.7mm}
\begin{array}{rl}
\text{MNLP} \triangleq &\hspace{-2.4mm} \displaystyle\frac{1}{|D^*|} \sum_{(\vect{x}^*, y^*) \in D^*} {-\log p(y^* |\vect{x}^*, D_i^r)} \\
 = & \hspace{-2.4mm}\displaystyle\frac{1}{|D^*|} \sum_{(\vect{x}^*, y^*) \in D^*} {\frac{1}{2} \left( \log(2 \pi \sigma_*^2) + \frac{(\mu_* - y^*)^2}{\sigma_*^2} \right)}\vspace{-1mm} 
\end{array}
\label{mnlp}
\end{equation}
where $D^*$ denotes a test dataset,  
and $\mu_*$ and $\sigma_*^2$ denote,  respectively, the predictive mean and variance of the predictive distribution $p(y^* |\vect{x}^*, D_i^r)$.
We then use these metrics to show how an improvement in IG can affect the predictive accuracy of a trained model on a common test dataset.\footnote{We use 
a randomly sampled test dataset to illustrate how IG is a suitable surrogate measure of the test/predictive accuracy of a trained model. In Sec.~\ref{MI}, we have discussed that in practice, it is often tedious or even impossible for all parties to agree on a common validation dataset, let alone a common set of test queries.}
More details of the experimental settings such as how to select the test dataset are in Appendix~\ref{detail}.
%
%
The performance of our reward scheme cannot be directly compared against that of the existing works  \cite{ghorbani2019data,jia2019towards} as they mainly focus on non-Bayesian classification models and assume monetary compensations.
For all experiments, we consider a collaboration of $n=3$ parties as it suffices to show interesting results. 

\textbf{Gaussian process (GP) regression with synthetic Friedman dataset.} 
For each party, we generate $250$ data points from the Friedman function.
We assign party $1$ the most valuable data by spanning the first input feature over the entire input domain $[0, 1]$.
In contrast, for parties $2$ and $3$, the same input feature spans only the non-overlapping ranges of $[0,0.5]$ and $[0.5, 1]$, respectively. This makes the data of parties $2$ and $3$ highly valuable to each other, i.e.,  $v_{\{2,3\}}$ is high relative to $v_2$. 
Using \eqref{shapley}, the Shapley values of the three parties are
$\phi_1 = 34.57$, $\phi_2 = 29.24$, and $\phi_3 = 30.78$.
Party $1$ has the largest $\phi_1$ and will always receive the most valuable model reward with the highest IG or value $v_N$.
\begin{figure}
    \begin{tabular}{@{}c@{\hspace{1.5mm}}c@{}}
       \includegraphics[scale=0.195]{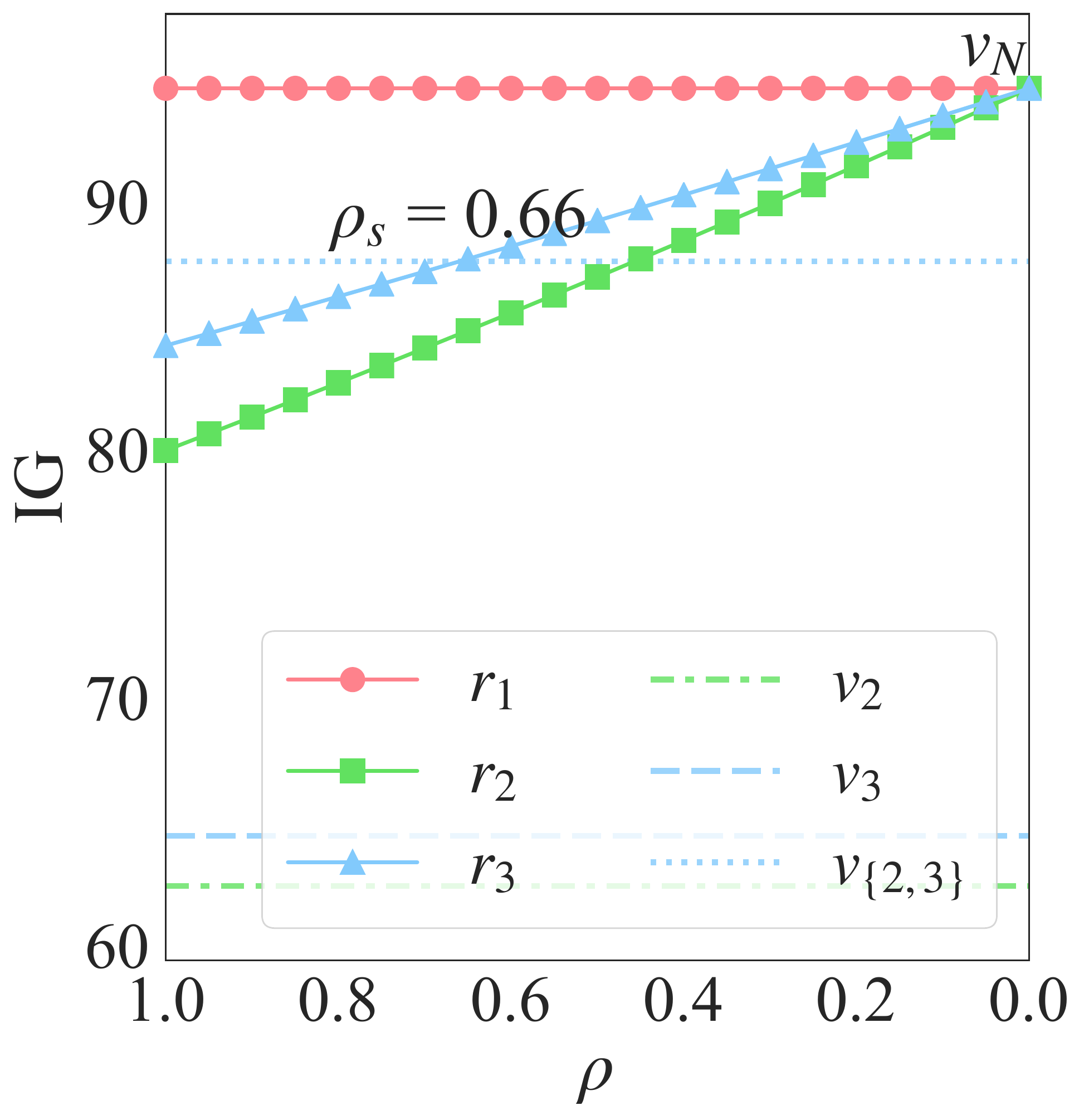} & \includegraphics[scale=0.195]{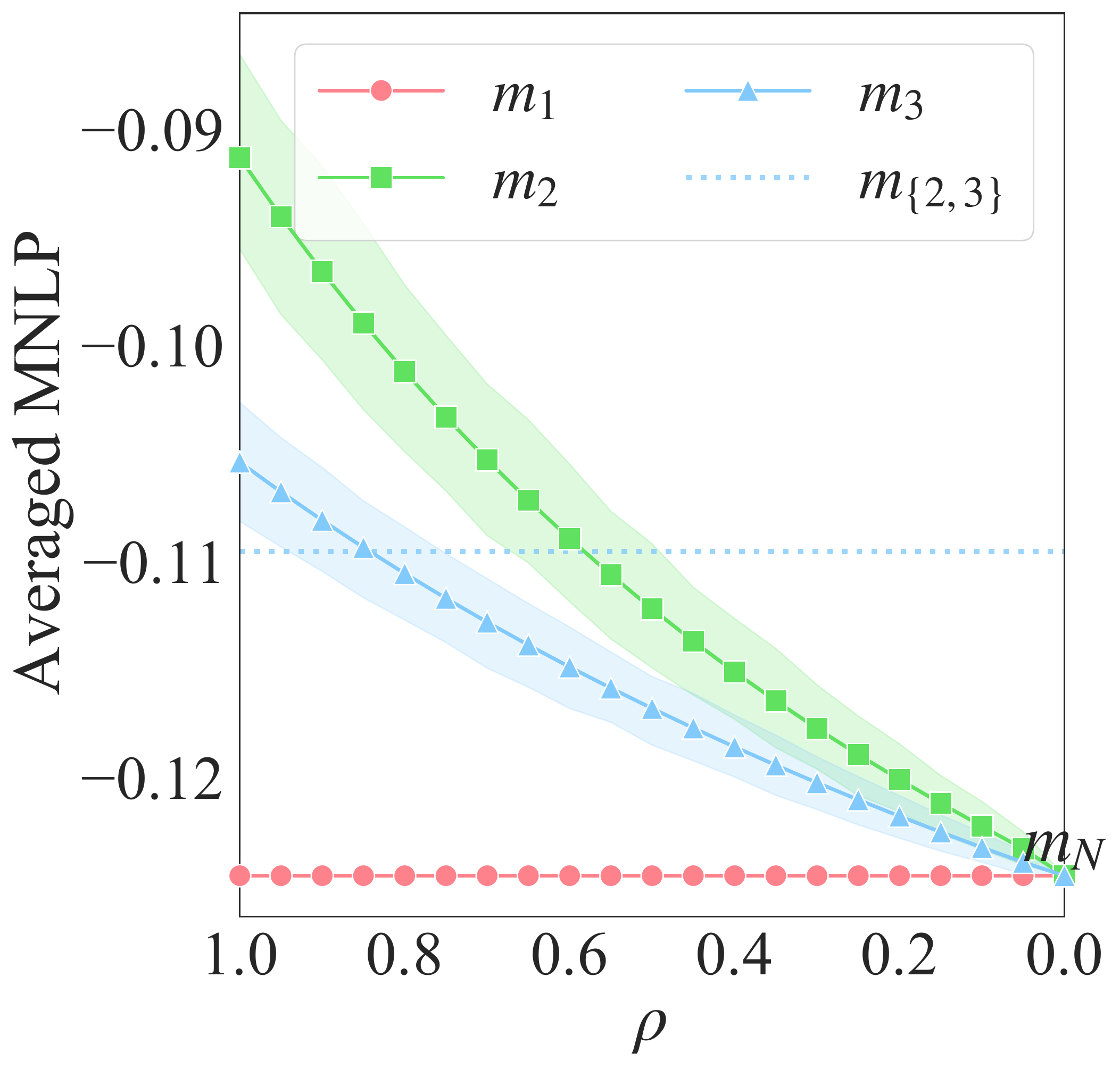} 
       \\
       {(a) IG} & {(b) MNLP} 
\end{tabular}
    \caption{Graphs of (a) IG and (b) MNLP vs.~the adjustable parameter $\rho$ for GP regression with synthetic Friedman dataset where $m_i$ and $m_C$ (with $|C| > 1$) denote the MNLPs of the models, respectively, rewarded to party $i$ and trained using data $D_C$.}
    \label{fig:power}\vspace{-3mm}
\end{figure}

Fig.~\ref{fig:power}a shows that our reward scheme indeed satisfies fairness (\ref{min-fairness}): A party $i$ with a larger $\phi_i$ always receives a higher $r_i$ (i.e., more valuable model reward).
%
Also, since $\rho_r \geq 1$ here, our reward scheme always satisfies individual rationality (\ref{rational}): Every party $i$ receives a more valuable model reward than that trained on its own data, i.e., its IG $r_i$ is higher than $v_i$ (see dotted lines in Fig.~\ref{fig:power}a).
As $\rho$ decreases (rightwards), party $1$'s most valuable model reward is unaffected but it has to be ``altruistic'' to parties $2$ and $3$ with increasing $r_2$ and $r_3$ (i.e., increasingly valuable model rewards) but smaller $\phi_2$ and $\phi_3$, as discussed in the last paragraph of Sec.~\ref{Tying}.
When $\rho$ decreases to $\rho_s$, stability (\ref{stability}) is reached: Party $3$'s model reward matches what it will receive by only collaborating with party $2$.
When $\rho = 0$, all parties receive an equally valuable model reward with the same IG/value $v_N$ despite their varying expected marginal contributions.
This shows how $\rho$ can be reduced to trade off  proportionality in pure Shapley fairness for satisfying other incentives such as stability (\ref{stability}) and group welfare (\ref{group-welfare}).

In Fig.~\ref{fig:power}b, we report the averaged MNLP and shade the $95\%$ confidence interval over $20$ different realizations of $\vect{z}_i$. 
As $\rho$ decreases and each party receives an increasingly valuable model reward (i.e., increasing IG/$r_i$), MNLP decreases,  
thus showing that an improvement in IG translates to a higher predictive accuracy of its trained model.
\begin{figure}
        \begin{tabular}{@{}c@{\hspace{1.5mm}}c@{}} \includegraphics[width=0.49\columnwidth]{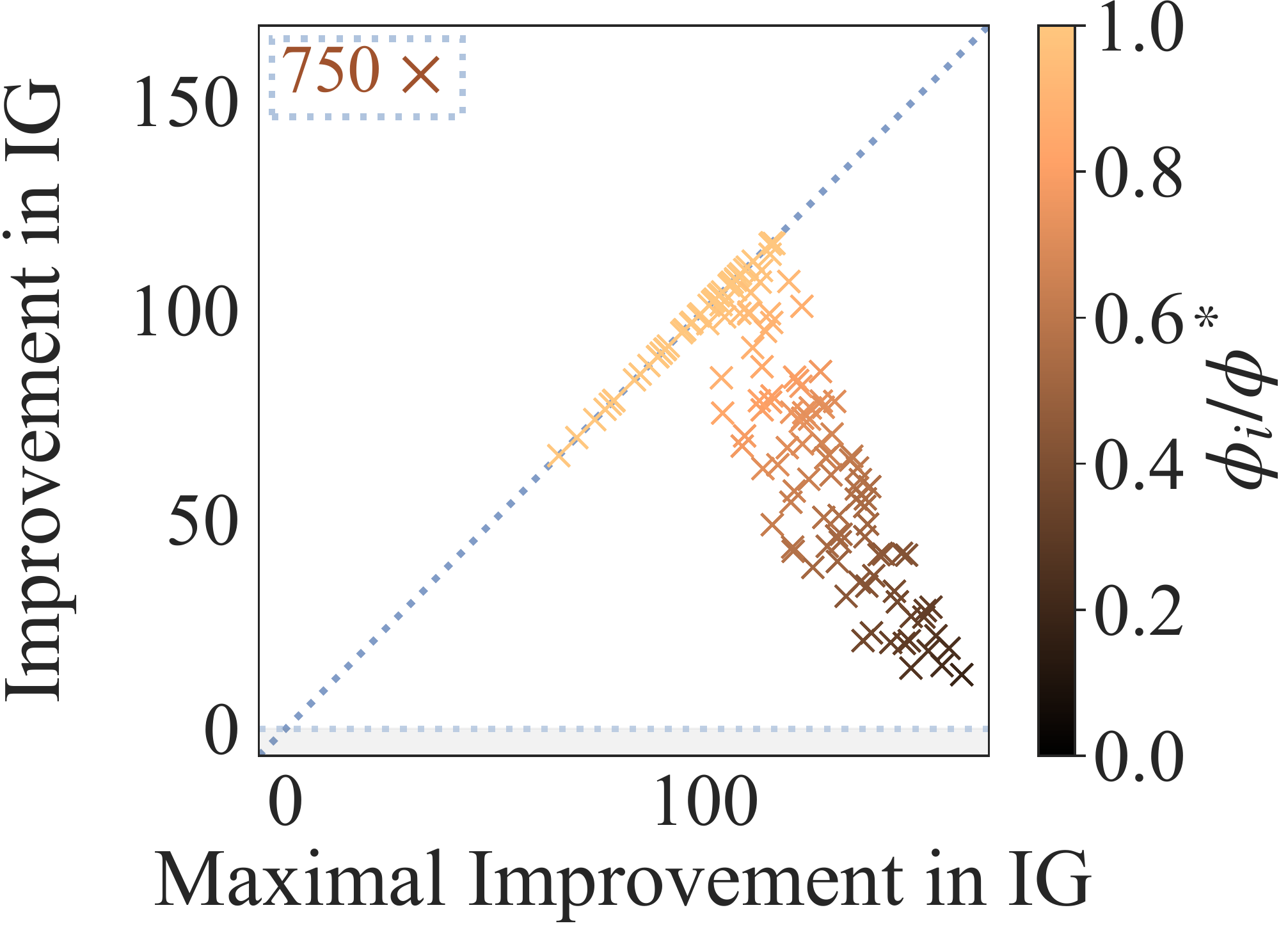} &  \includegraphics[width=0.49\columnwidth]{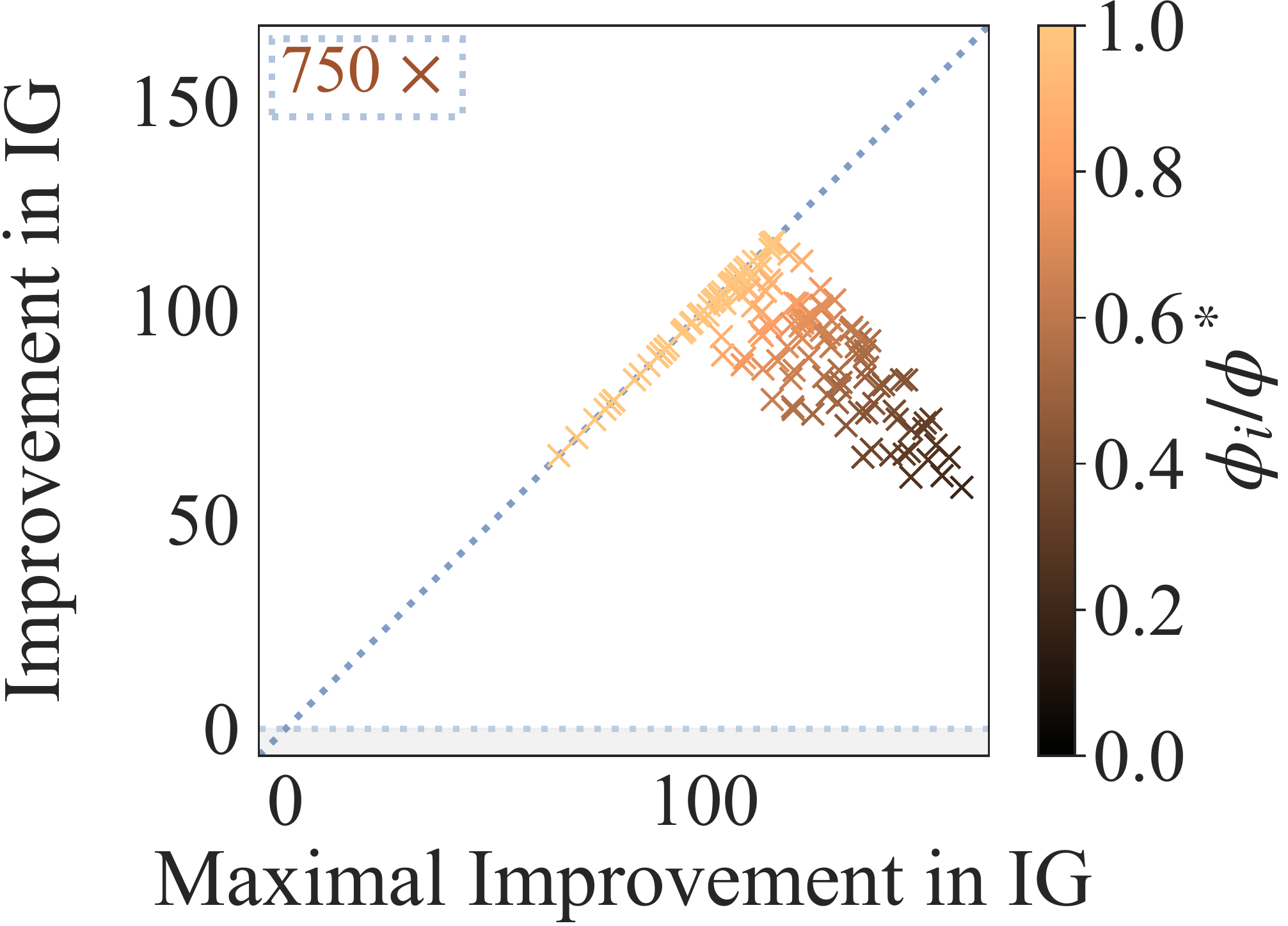} \\
        {(a) IG, $\rho = 1$} & {(b) IG, $\rho = 0.5$} \vspace{1mm}\\
        \includegraphics[width=0.49\columnwidth]{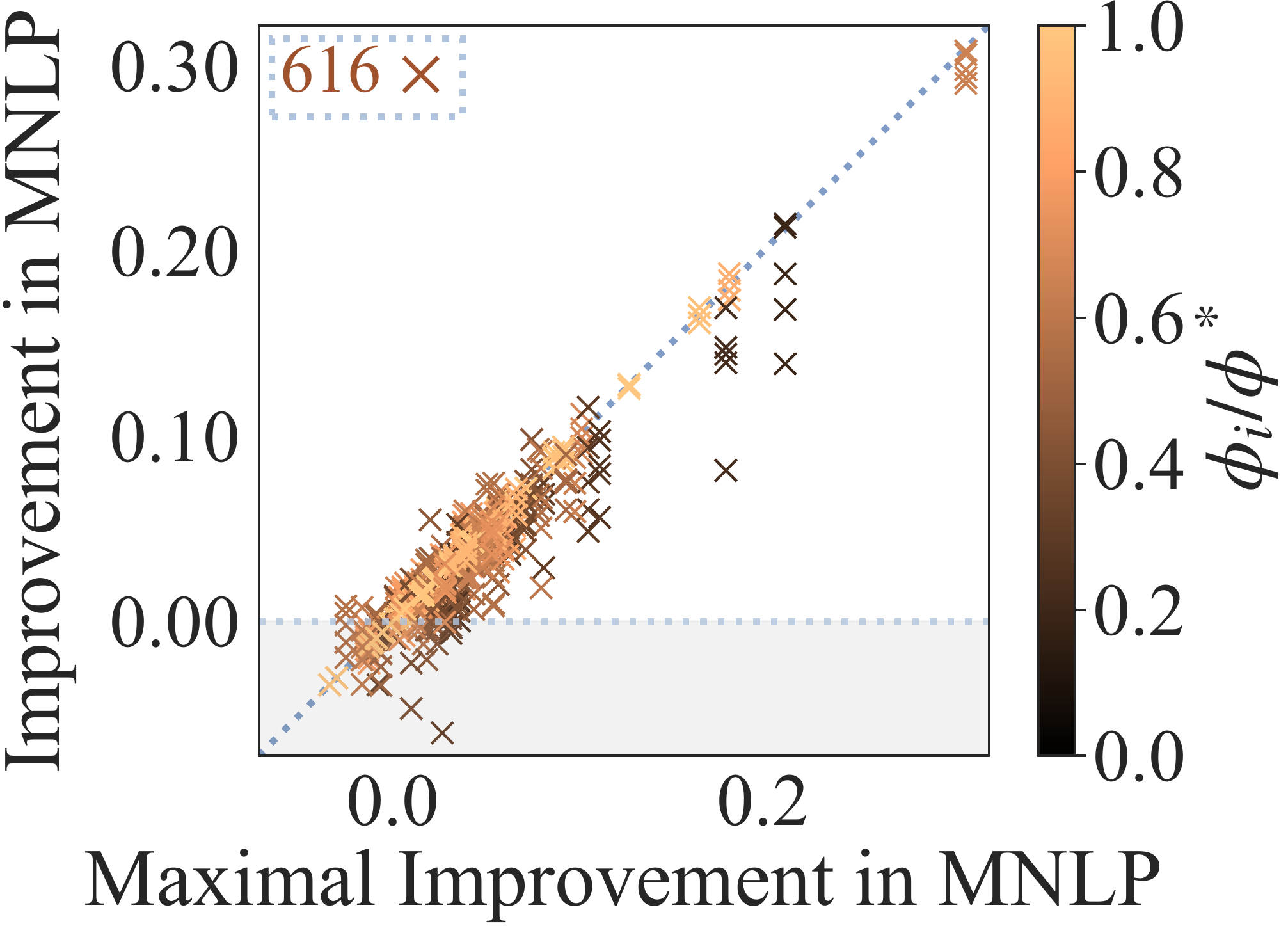} &  \includegraphics[width=0.49\columnwidth]{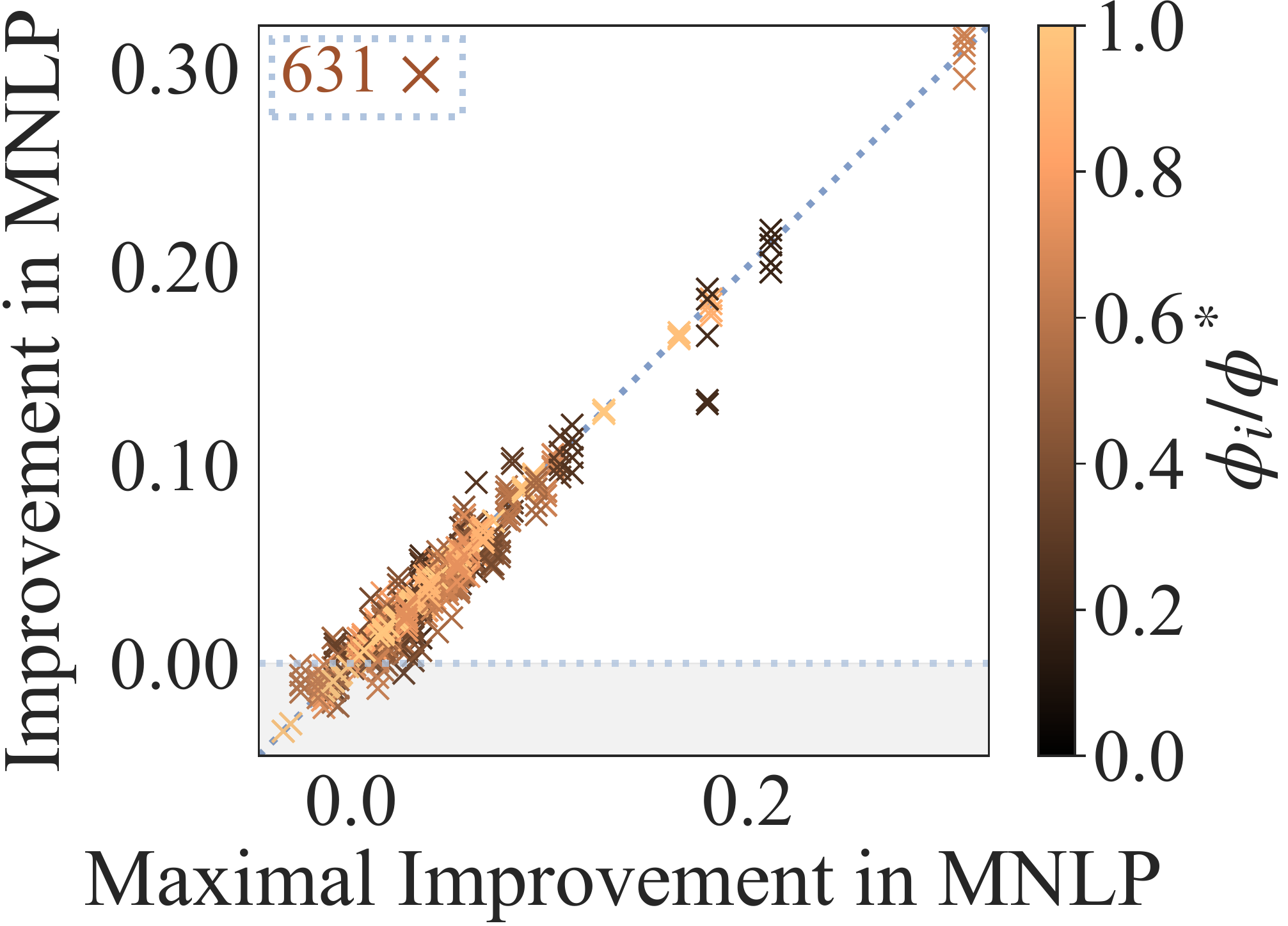} \\
        {(c) MNLP, $\rho = 1$} & {(d) MNLP, $\rho = 0.5$}\vspace{-3mm}
        \end{tabular}
        \caption{Scatter graph of the (a-b) improvement in IG ($r_i - v_i$) vs.~maximal improvement in IG ($v_N - v_i$) and (c-d) improvement in MNLP vs.~maximal improvement in MNLP when $\rho=1, 0.5$ for multiple partitions of DiaP dataset among $n=3$ parties.
        }
      \label{fig:diabetes}\vspace{-2.6mm}
\end{figure}

\textbf{GP regression with DiaP dataset.}
We train a GP regression model with a composite kernel (i.e., squared exponential kernel $+$ exponential kernel) and  
partition the training dataset among $n=3$ parties, as detailed later.
The results are shown in Fig.~\ref{fig:diabetes}. 

\begin{figure}
        \begin{tabular}{@{}c@{\hspace{1.5mm}}c@{}}                               \includegraphics[width=0.49\columnwidth]{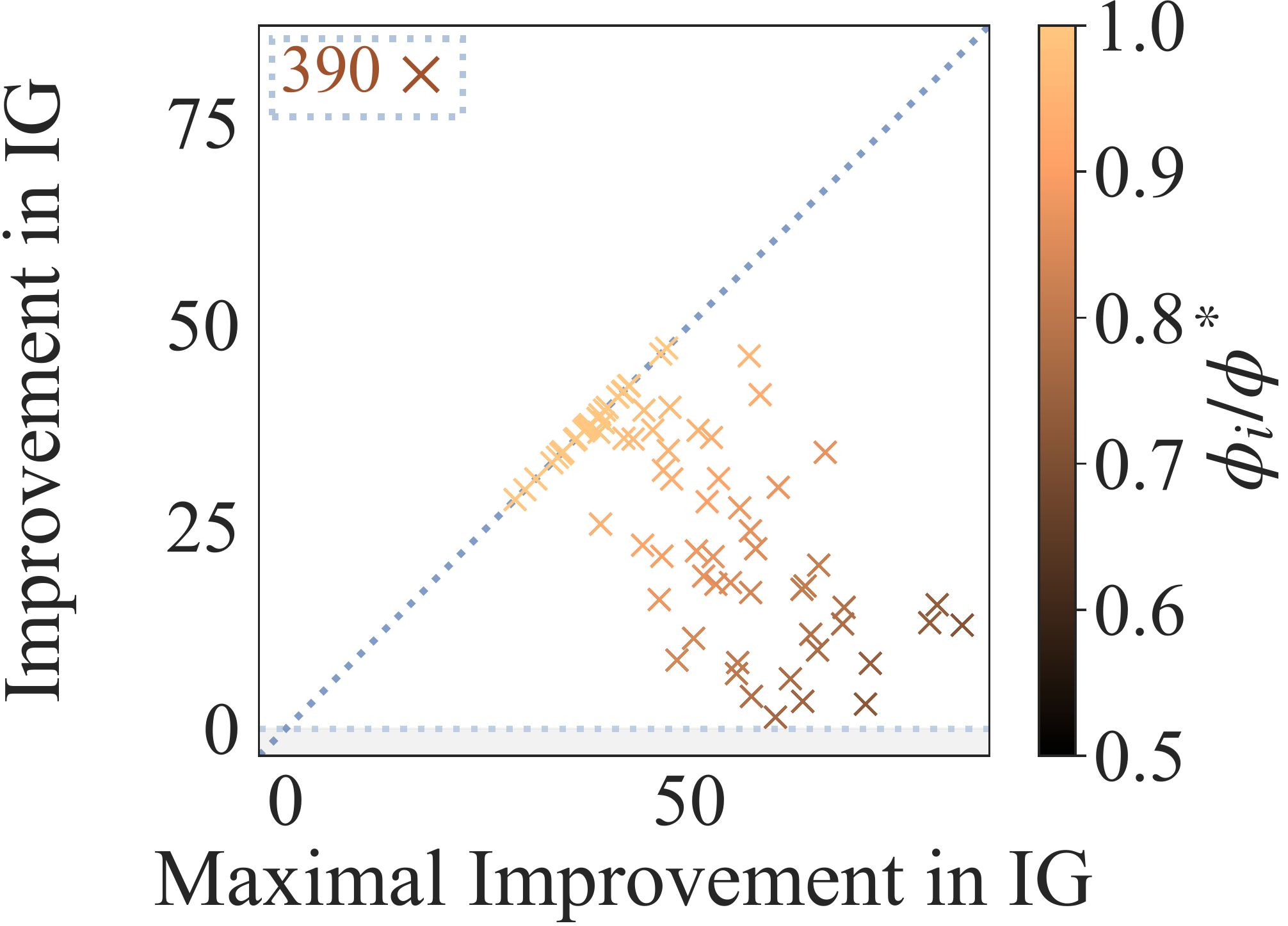} & \includegraphics[width=0.49\columnwidth]{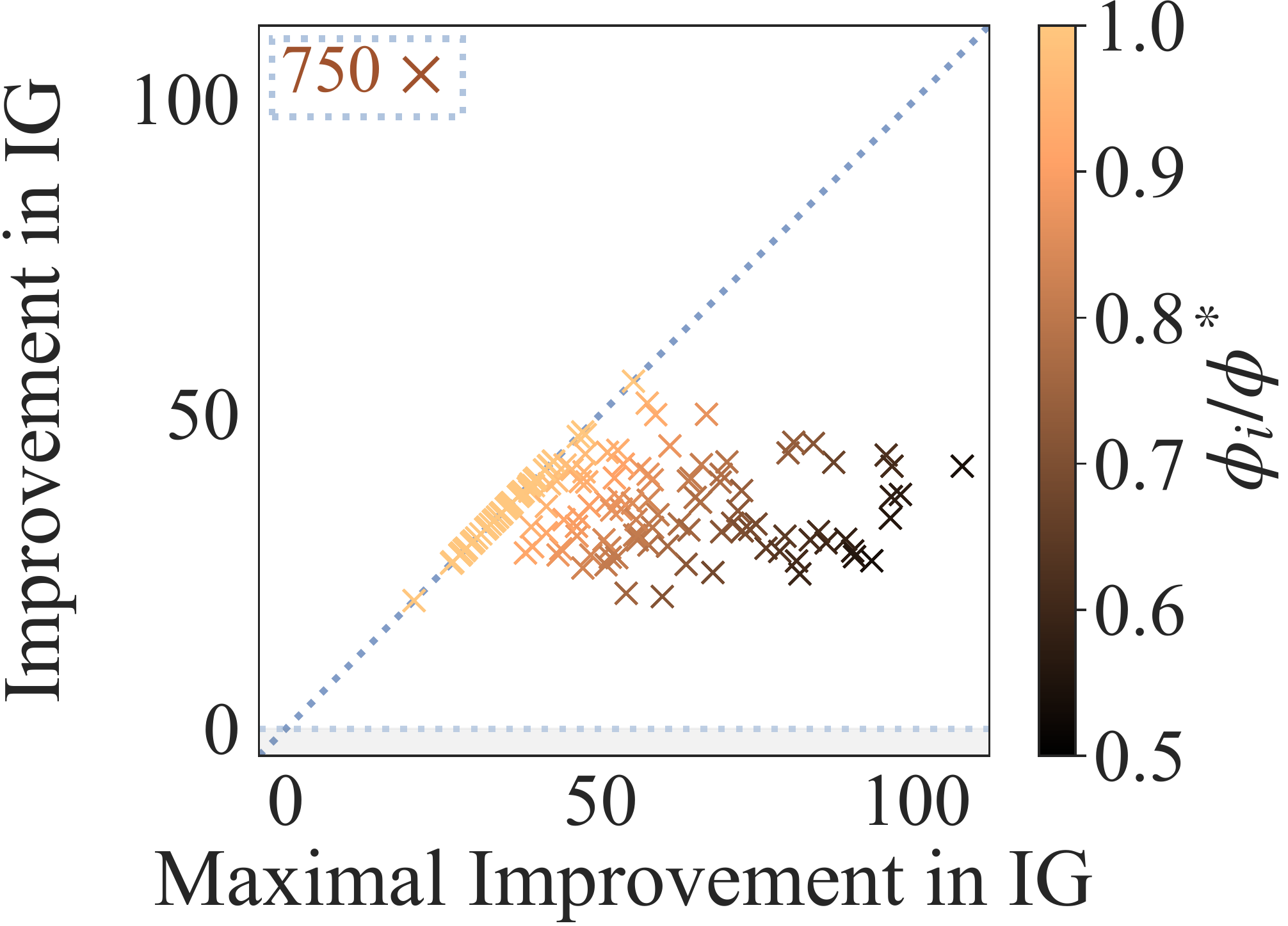} \\
        {(a) IG, $\rho = 1$} &{(b) IG, $\rho = 0.5$} \vspace{1mm}\\
        \includegraphics[width=0.49\columnwidth]{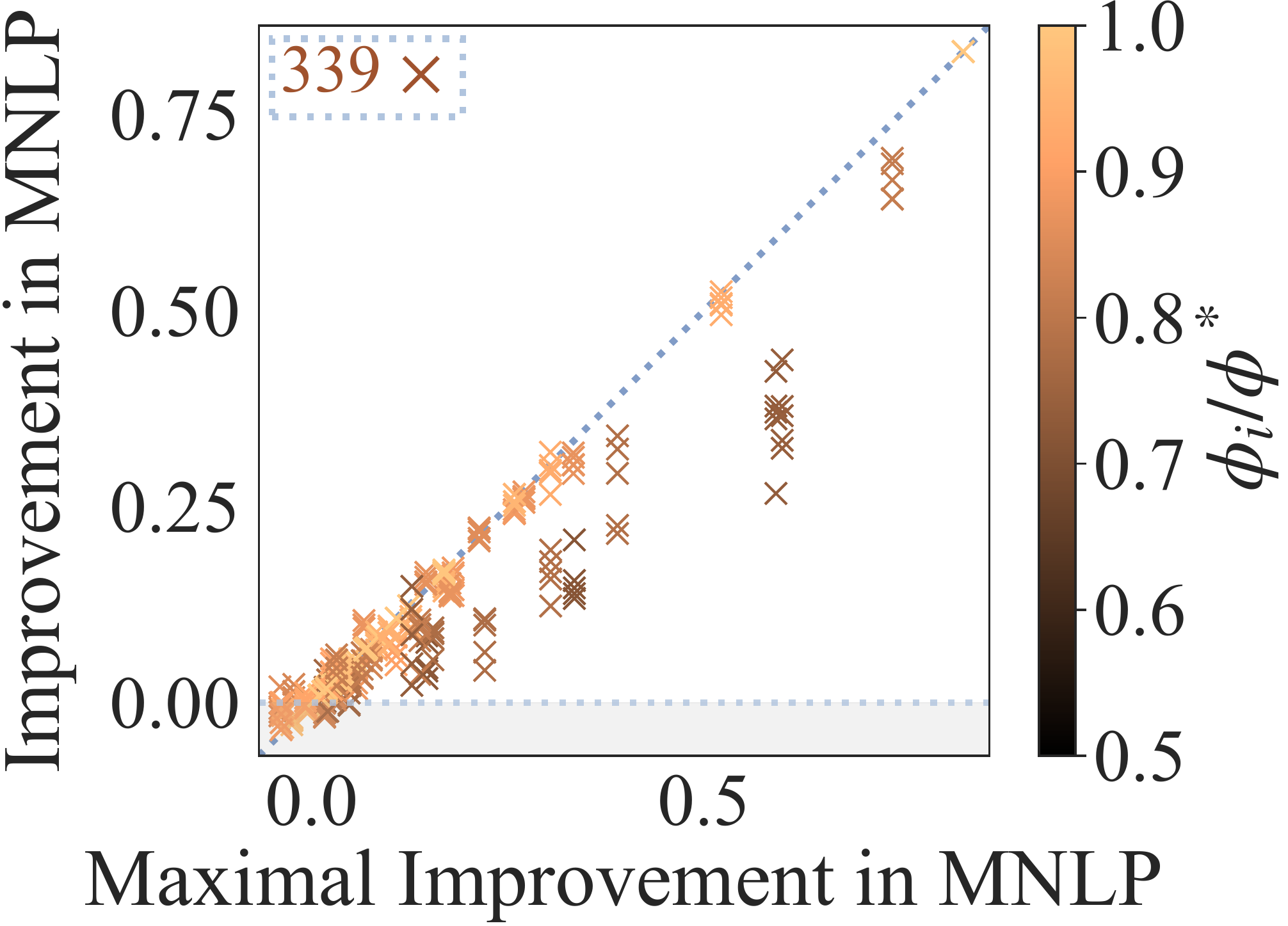} & \includegraphics[width=0.49\columnwidth]{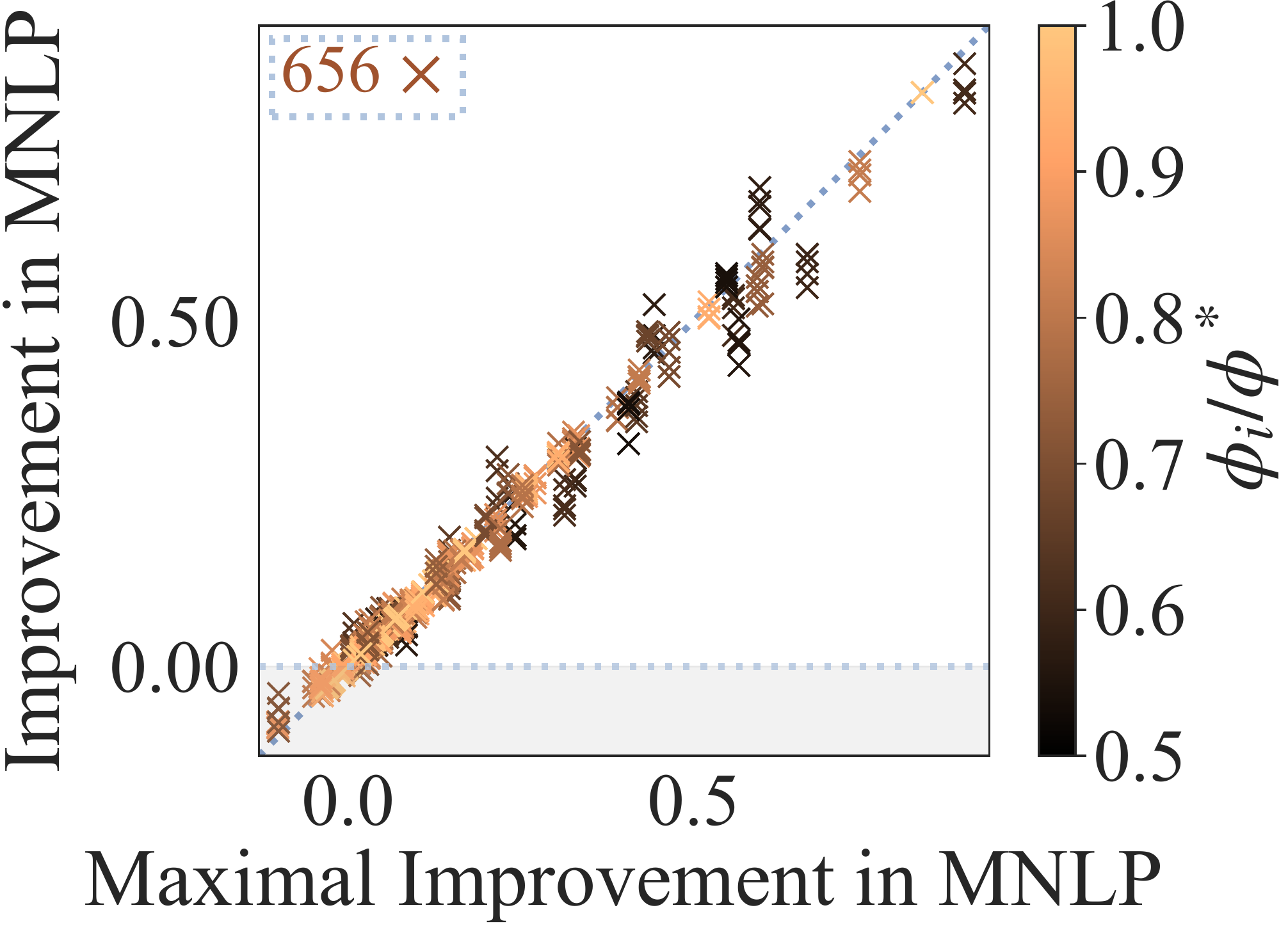} \\
        {(c) MNLP, $\rho = 1$} & {(d) MNLP, $\rho = 0.5$}
        \end{tabular}
        \caption{Scatter graph of the (a-b) improvement in IG ($r_i - v_i$) vs.~maximal improvement in IG ($v_N - v_i$) and (c-d) improvement in MNLP vs.~maximal improvement in MNLP when $\rho=1, 0.5$ for multiple partitions of CaliH dataset among $n=3$ parties.
        }
        \label{fig:cali}
\end{figure}

\textbf{Neural network regression with CaliH dataset.}
We consider \emph{Bayesian linear regression} (BLR) on the last layer of a \emph{neural network} (NN).
$60\%$ of the CaliH data is designated as the ``public'' dataset\footnote{
In practice, the ``public" dataset can be provided by the government to kickstart the collaborative ML.
Alternatively, it can be historical data that companies are more willing to share.} and used to pretrain a NN.
Since the correlation of the house value with the input features in the data of the parties may differ from that in the ``public" dataset, we perform transfer learning and selectively retrain only the last layer using BLR with a standard Gaussian prior. 
So, IG is only computed based on  BLR.
From the remaining data, we select $400$ data points for the test dataset and $1600$ data  points\footnote{We restrict the total number of data points so that every individual party cannot train a high-quality model alone.} for the training dataset to be partitioned among $3$ parties, as detailed later.
The results are shown in Fig.~\ref{fig:cali}. 

\begin{figure}
        \begin{tabular}{@{}c@{\hspace{1.5mm}}c@{}} \includegraphics[width=0.49\columnwidth]{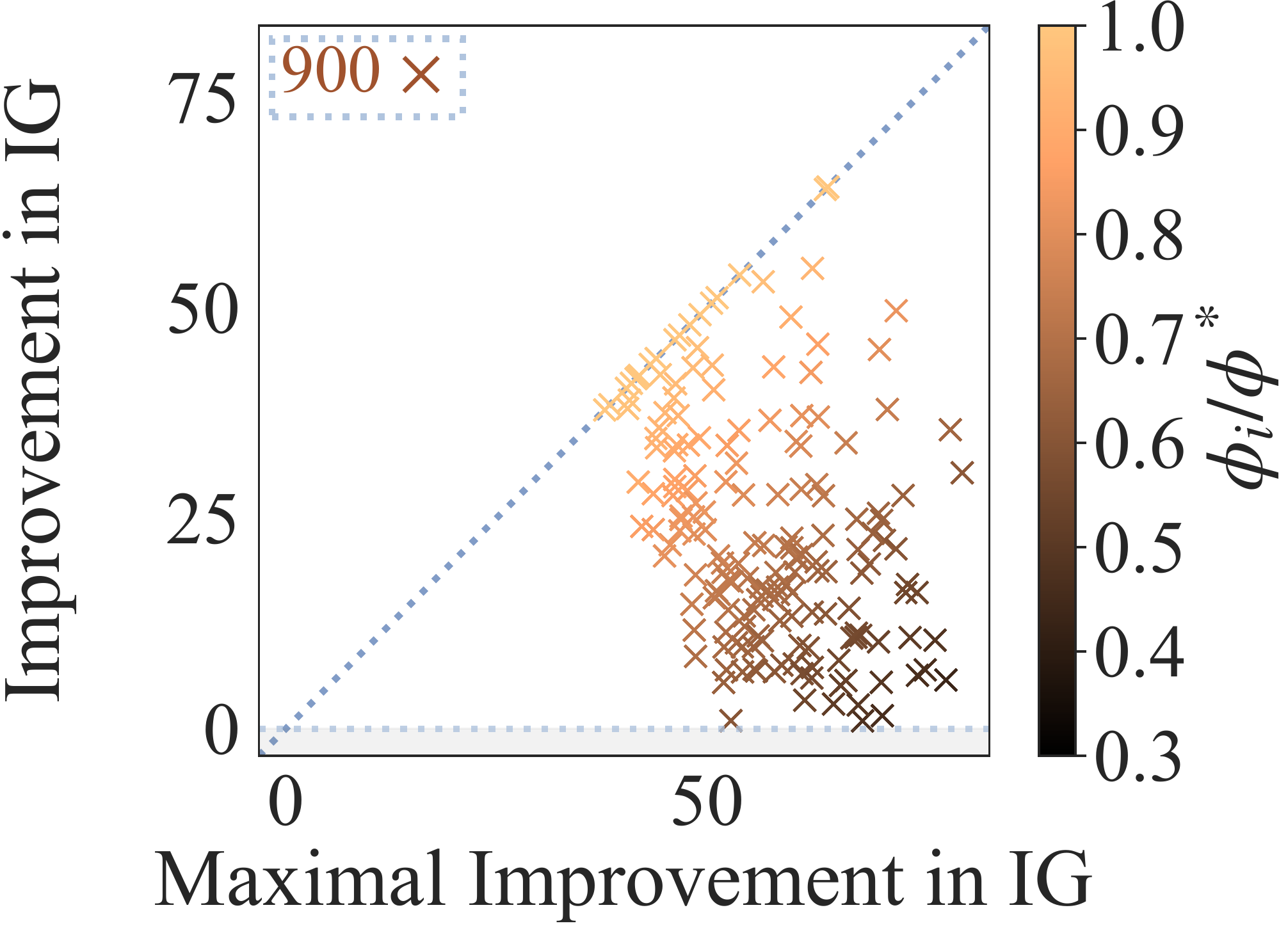} &  \includegraphics[width=0.49\columnwidth]{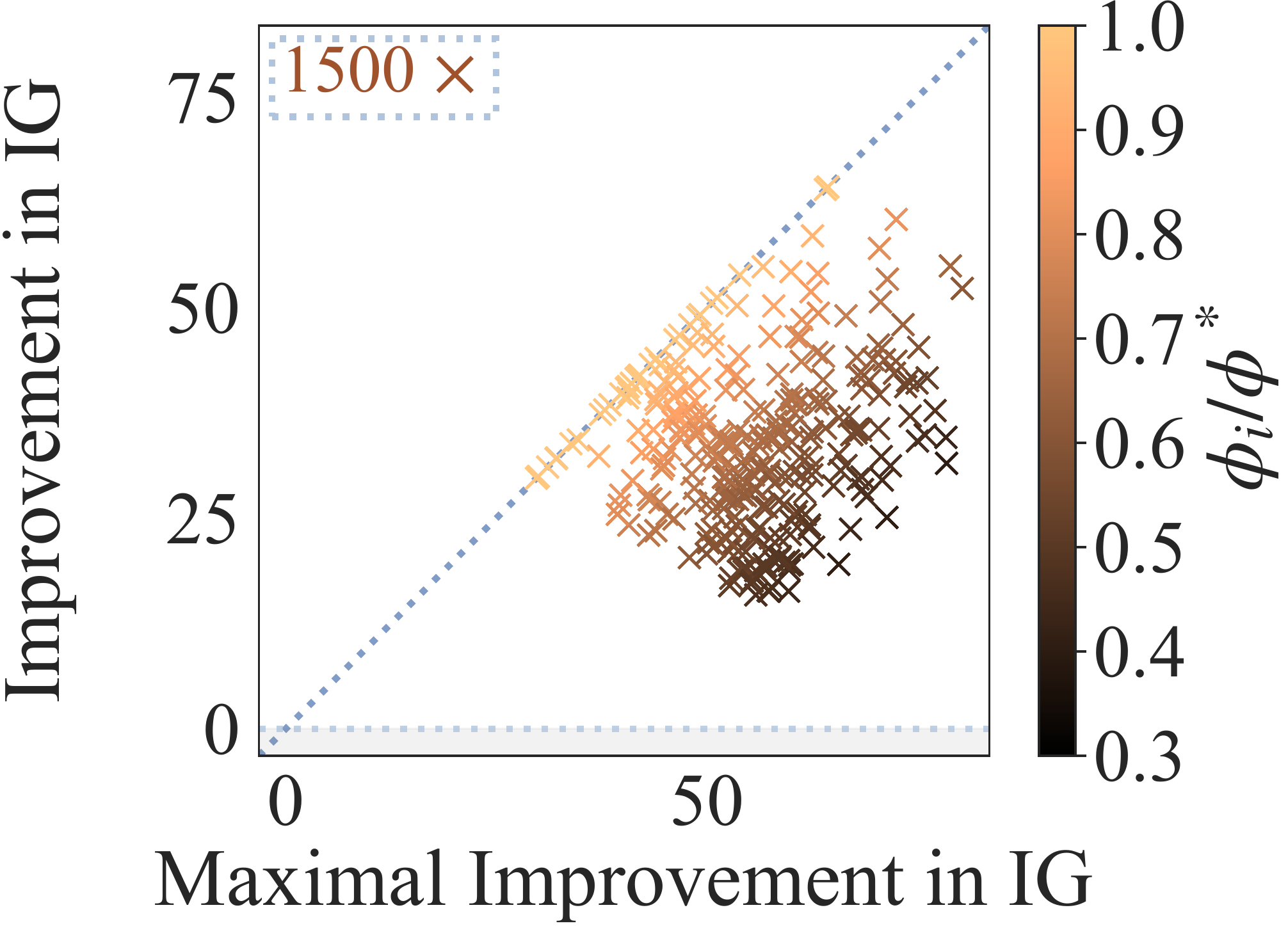} \\
        {(a) IG, $\rho = 1$} & {(b) IG, $\rho = 0.5$} \vspace{1mm}\\
        \includegraphics[width=0.49\columnwidth]{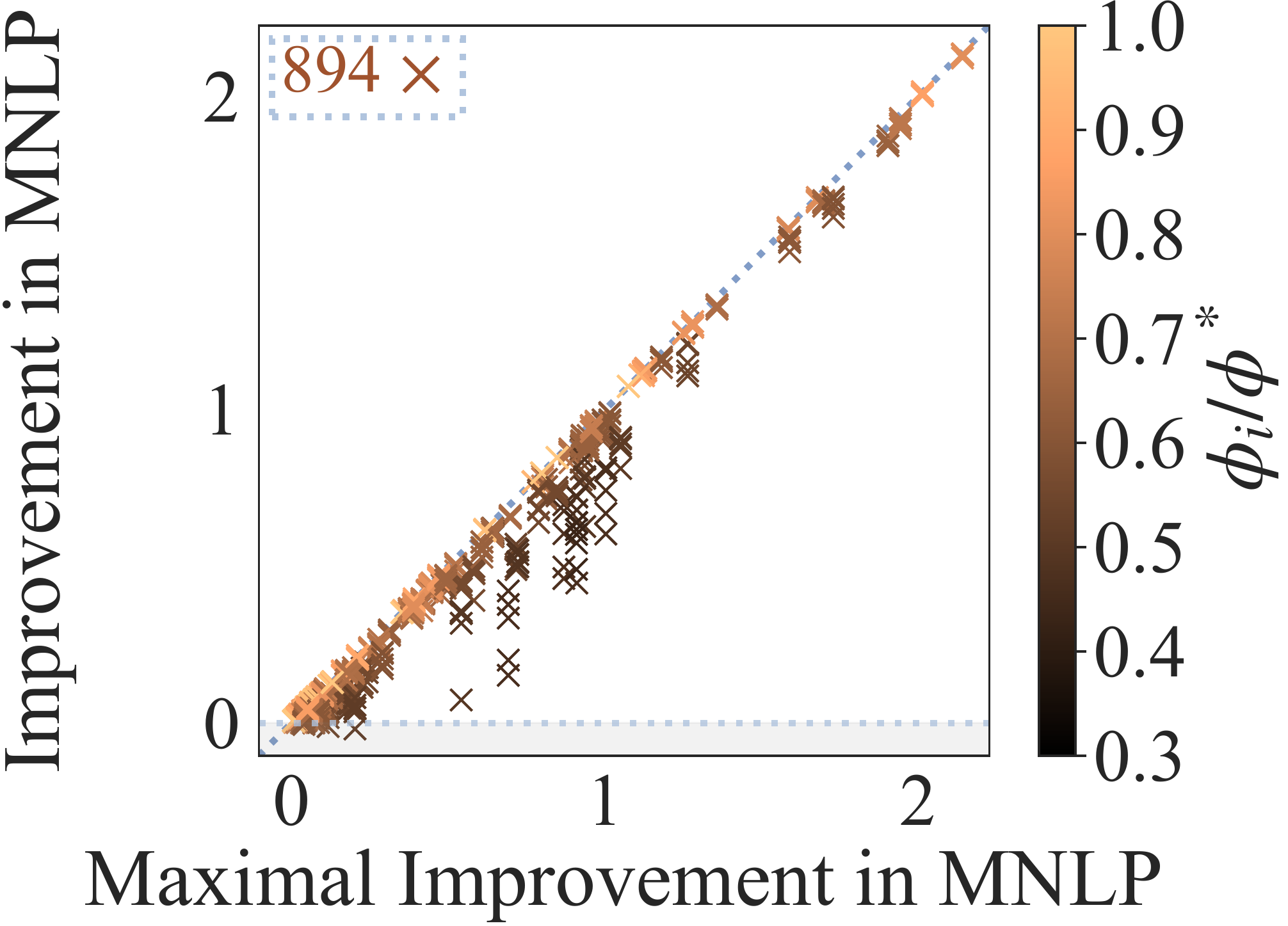} &  \includegraphics[width=0.49\columnwidth]{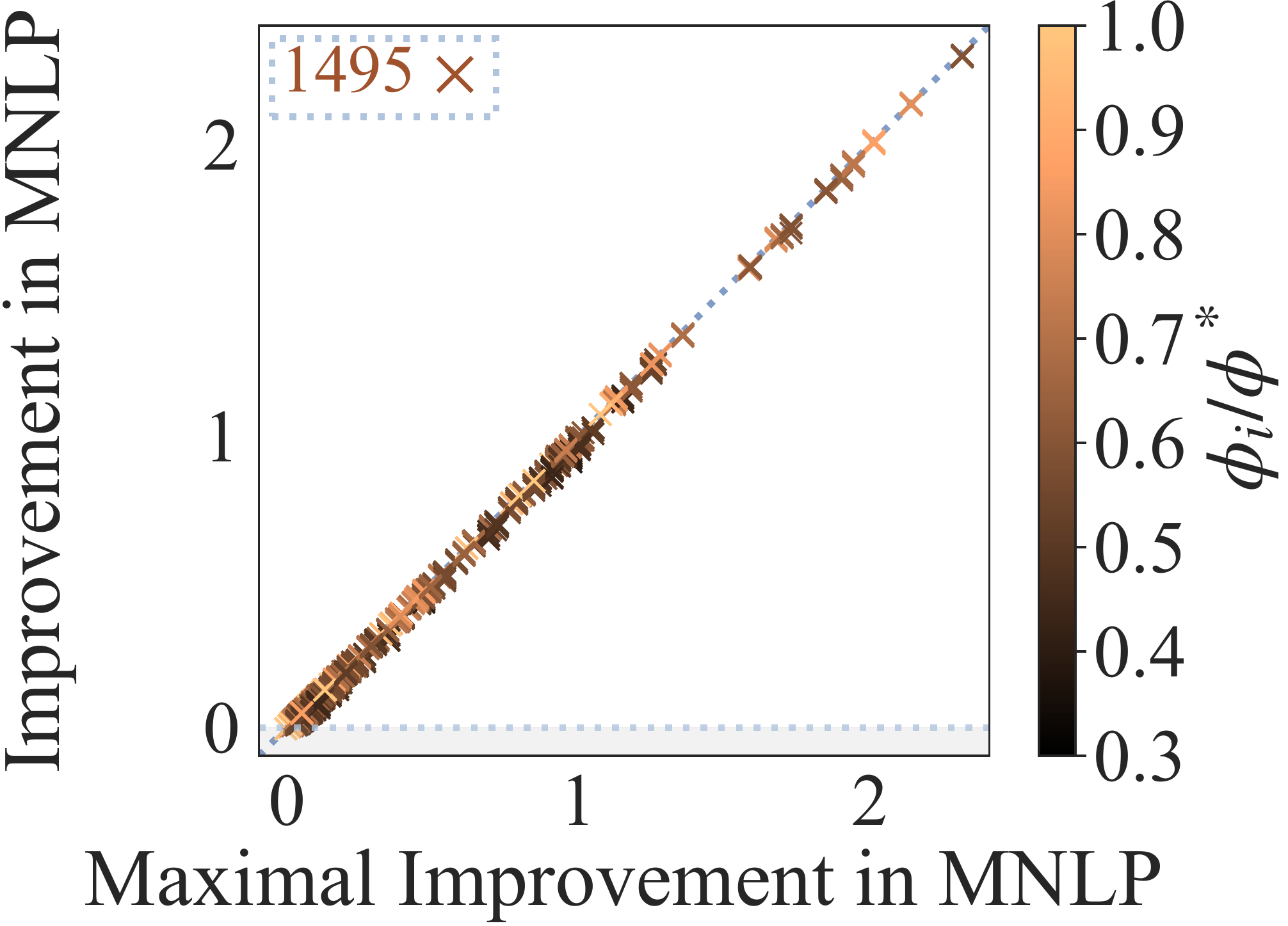} \\
        {(c) MNLP, $\rho = 1$} & {(d) MNLP, $\rho = 0.5$}
        \end{tabular}
        \caption{Scatter graph of the (a-b) improvement in IG ($r_i - v_i$) vs.~maximal improvement in IG ($v_N - v_i$) and (c-d) improvement in MNLP vs.~maximal improvement in MNLP when $\rho=1, 0.5$ for multiple partitions of Friedman dataset among $n=10$ parties.
        }
        \label{fig:friedman-10-parties-gp}
\end{figure}

\textbf{Sparse GP regression with synthetic Friedman dataset (10 parties).}
We also evaluate the performance of our reward scheme on the collaboration between a larger number $n=10$ of parties.
We consider a larger synthetic Friedman dataset of size $5000$ and partition the training dataset among $10$ parties such that each party owns at least $5\%$ of the dataset. We train a sparse GP model based on the \emph{deterministic training conditional} approximation~\cite{NghiaICML16,HoangICML16} and a squared exponential kernel. Since the exact Shapley value~\eqref{shapley} is expensive to evaluate with a large number of parties, we approximate it using the \emph{simple random sampling} algorithm in \cite{maleki2013}.
The results are shown in Fig.~\ref{fig:friedman-10-parties-gp}. 

\begin{figure}
    \centering
    \includegraphics[scale=0.26]{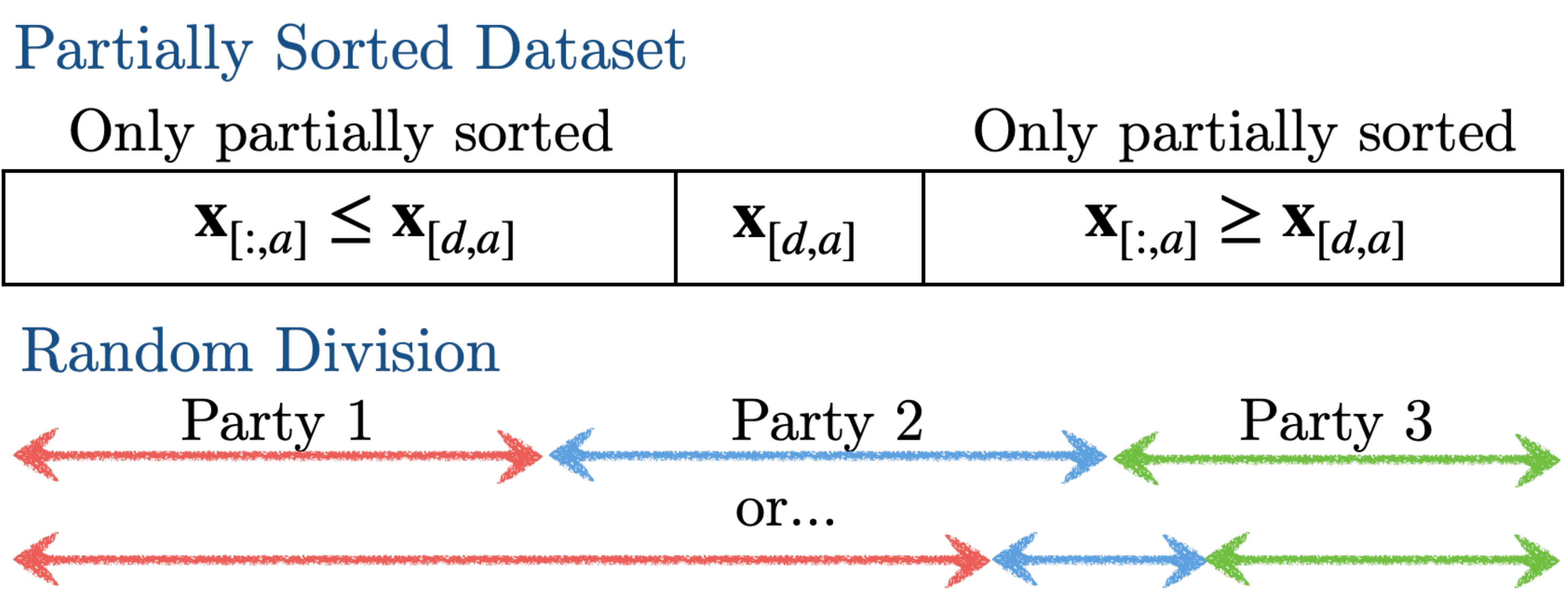}\vspace{-3mm}
    \caption{Only data point $d$ will be at its sorted position based on input feature $a$. We then divide this partially sorted training dataset into $3$ consecutive blocks of random sizes with the constraint that each party owns at least $10\%$ of the dataset.}\vspace{-3.2mm}
    \label{fig:datapart}
\end{figure}

\textbf{Partitioning Algorithm.}
To divide any training dataset among parties and vary their contributed data, we first choose an input feature $a$ uniformly at random from all the input features.
Next, the data point $d$ is chosen uniformly at random.
We partially sort the dataset and divide it into continuous blocks of random sizes, as detailed in Fig.~\ref{fig:datapart}.
Such a setup allows parties to have different quantities of unique or partially overlapping data.
The input feature $a$ may have real-world significance: If $a$ is the age of patients, then this models the scenario where hospitals have data of patients with different demographics.

\textbf{Summary of Observations.}
For any dataset, we consider different train-test splits and partitions of the training dataset using the above algorithm.
For each partition, we compute the optimized noise variance $\eta_i$ to realize $r^*_i$ and consider $5$ realizations of $\vect{z}_i$ for each party $i\in N$.
For a given partition and choice of $\rho$, if the values $(r_i)_{i \in N}$ of model rewards satisfy individual rationality (\ref{rational}), then we compute the ratio $\phi_i/\phi^*$,
improvement in IG ($r_i - v_i$) and MNLP, and 
maximal (possible) improvement in IG ($v_N - v_i$) and MNLP for each party $i\in N$.
The improvement is measured relative to a model trained only on party $i$'s data.
The best/lowest MNLP $m_N$ is assumed to be achieved by a model trained on the aggregated data of all parties.
Each realization of $\vect{z}_i$ (from each party $i$) will produce a point in the scatter graphs in Figs.~\ref{fig:diabetes},~\ref{fig:cali}, and~\ref{fig:friedman-10-parties-gp}.
If a point lies on the diagonal identity line, then the corresponding party (e.g., with the largest $\phi_i$, i.e., $\phi^*$) receives a model reward with MNLP $m_N$.
We also report the number of points with positive improvement in IG and MNLP in the top left hand corner of each graph.

In Figs.~\ref{fig:diabetes}c-d, \ref{fig:cali}c-d, and \ref{fig:friedman-10-parties-gp}c-d, the improvement in MNLP is usually positive: The predictive performance of each party benefits from the collaboration.
Occasionally but reasonably, this does not hold when the maximal improvement in MNLP is extremely small or even negative. 
Lighter colored points lie closer to the diagonal line: Parties with  
larger $\phi_i$ receive more valuable model rewards (Figs.~\ref{fig:diabetes}a-b,~\ref{fig:cali}a-b,~\ref{fig:friedman-10-parties-gp}a-b) which translates to lower MNLP (Figs.~\ref{fig:diabetes}c-d,~\ref{fig:cali}c-d,~\ref{fig:friedman-10-parties-gp}c-d).

In Figs.~\ref{fig:diabetes},~\ref{fig:cali} and~\ref{fig:friedman-10-parties-gp}, when $\rho$ decreases from $1$ to $0.5$, darker colored points move closer to the diagonal line, which implies that parties with smaller $\phi_i$ can now receive more valuable model rewards with higher predictive accuracy.
The number of points (reported in the top left corner) also increase as more of them satisfy \ref{rational}.

In Fig.~\ref{fig:friedman-10-parties-gp}d with $\rho=0.5$, most points are close to the diagonal identity line
because it may be the case that not all training data points are needed to achieve the lowest MNLP. Instead, fewer or noisier data points are sufficient. In this scenario, a larger $\rho$ may be preferred to reward the parties differently.
More experimental results for different Bayesian models and datasets are shown in Appendix~\ref{ap.bosh}.

\textbf{Limitations}. Though IG is a suitable surrogate measure of the predictive accuracy of a trained model, a higher IG does not always translate to lower MNLP.
In Figs.~\ref{fig:diabetes}c-d and~\ref{fig:cali}c-d, occasionally, the improvement in MNLP is negative (around $15\%$ of the time) and darker points lie above the diagonal line. The latter suggests that 
parties with smaller $\phi_i$
may receive model rewards with MNLP lower than $m_N$  (i.e., awarded to the party with largest $\phi_i$, i.e., $\phi^*$).
This may be purely due to randomness, but we have also investigated factors
(e.g., model selection)
leading to a weak relationship between IG and MNLP and reported the results in Appendix~\ref{ap.ig-mnlp}.
Our reward scheme works best when a suitable model 
is selected 
and the model prior is not sufficiently informative for any party to achieve a high predictive accuracy by training a model on its own data.
In practice, this is reasonable as collaboration precisely happens only when every individual party cannot train a high-quality model alone but can do so from working together.
\section{Conclusion}
This paper describes a collaborative ML scheme that distributes only model rewards to the parties. We perform data valuation based on the IG on the model parameters given the data and compute each party's marginal contribution using the Shapley value.
To decide the appropriate rewards to incentivize the collaboration, we adapt solution concepts from CGT (i.e., fairness, Shapley fairness, stability, individual rationality, and group welfare) for our freely replicable model rewards. We propose a novel reward scheme (Theorem~\ref{mi-power}) with an adjustable parameter that can trade off between these incentives while maintaining fairness.
Empirical evaluations show that a smaller $\rho$ and more valuable model rewards translate to higher predictive accuracy.
Current efforts on designing incentive mechanisms for federated learning can build upon our work presented in this paper.
For future work, we plan to address privacy preservation in our reward scheme. The noise injection method used for realizing the model rewards in this work is closely related to the Gaussian mechanism of differential privacy \cite{dwork2014}. This motivates us to explore how the injected noise will affect privacy in the model rewards.

\section*{Acknowledgements}
This research/project is supported by the National Research Foundation, Singapore under its Strategic Capability Research Centres Funding Initiative and its AI Singapore Programme (Award Number: AISG-GC-$2019$-$002$). Any opinions, findings, and conclusions or recommendations expressed in this material are those of the author(s) and do not reflect the views of National Research Foundation, Singapore.

\bibliography{sources}
\bibliographystyle{icml2020}

\clearpage
\appendix
\section{Proof of  Submodularity of IG~\eqref{data_IG}}
\label{a.IG}
Firstly, we assume that the data $D_i$ and $D_j$ are conditionally independent given model parameters $\boldsymbol{\theta}$ for any $i, j \in N$ and $i \neq j$.
Then, $\forall i \in N \ \ \forall C\subseteq C'\subseteq N \setminus \{i\}\ ,$
\begin{equation*}
\begin{array}{l}
v_{C\cup \{i\}} - v_{C} \vspace{1mm}\\
= \mathbb{I}(\boldsymbol{\theta};D_{C\cup \{i\}}) - \mathbb{I}(\boldsymbol{\theta}; D_{C}) \vspace{1mm}\\
= \mathbb{H}(\boldsymbol{\theta}| D_{C}) - \mathbb{H}(\boldsymbol{\theta}|D_{C\cup \{i\}}) \vspace{1mm}\\
= \mathbb{H}(D_i| D_{C}) - \mathbb{H}(D_i|\boldsymbol{\theta},D_{C}) \vspace{1mm}\\
= \mathbb{H}(D_i| D_{C}) - \mathbb{H}(D_i|\boldsymbol{\theta}) \vspace{1mm}\\
\geq \mathbb{H}(D_i| D_{C'}) - \mathbb{H}(D_i|\boldsymbol{\theta}) \vspace{1mm}\\
= \mathbb{H}(D_i| D_{C'})  - \mathbb{H}(D_i|\boldsymbol{\theta},D_{C'}) \vspace{1mm}\\
= \mathbb{H}(\boldsymbol{\theta}| D_{C'})  - \mathbb{H}(\boldsymbol{\theta}|D_{C'\cup \{i\}}) \vspace{1mm}\\
= \mathbb{I}(\boldsymbol{\theta};D_{C'\cup \{i\}}) - \mathbb{I}(\boldsymbol{\theta}; D_{C'}) \vspace{1mm}\\
= v_{C' \cup \{i\}} - v_{C'}
\end{array}
\end{equation*}
where the third and sixth equalities are due to the symmetric property of mutual information, the fourth and fifth equalities are due to the conditional independence of $D_i$ and $D_C$ given $\boldsymbol{\theta}$, and the  inequality is due to the property that conditioning on more information (i.e., $D_{C'\setminus C}$) never increases the entropy (i.e., ``information never hurts" bound for entropy) \cite{Cover91}.
\section{Proof of Proposition~\ref{p1}} 
\label{a.p1}
%
We prove below that $r_i = \text{Shapley}_v(i)$ for all $i \in N$ satisfy each of the properties \ref{FAIR1} to~\ref{FAIR4}:
\squishlisttwo
	\item \ref{FAIR1} \textbf{Uselessness.}
	For all $i \in N$, if
	$$\forall C\subseteq N \setminus \{i\}\ \ v_{C \cup \{i\}} = v_C\ ,
	$$
	then
	\begin{align*}
	r_i = \text{Shapley}_v(i) & = \frac{1}{n!} \sum_{\pi \in \Pi_N} {\left(v_{{S_{\pi,i}} \cup \{i\}} - v_{S_{\pi,i}}\right)} \\
	&= \frac{1}{n!} \sum_{\pi \in \Pi_N} {0} \\
	&= 0 \ . 
	\end{align*}
	
	\item \ref{FAIR2} \textbf{Symmetry.} For all $i,j\in N$ s.t.~$i \neq j$, if 
	\begin{equation}
	\forall C \subseteq N\setminus\{i,j\}\ \ v_{C \cup \{i\}} = v_{C \cup \{j\}}\ ,
	\label{dodo}
	\end{equation}
	then
	\begin{align*}
	r_i = \text{Shapley}_v(i) & = \frac{1}{n!} \sum_{\pi \in \Pi_N} {\left(v_{{S_{\pi,i}} \cup \{i\}} - v_{S_{\pi,i}}\right)} \\
	& = \frac{1}{n!} \sum_{\pi' \in \Pi_N} {\left(v_{{S_{\pi',j}} \cup \{j\}} - v_{S_{\pi',j}}\right)} \\
	&= \text{Shapley}_v(j) = r_j \ . 
	\end{align*}
	Note that $\pi'$ is obtained from $\pi$ by swapping the positions of $i$ and $j$ in $\pi$.
	Then, we prove the third equality by considering the following two cases: (a) When $i$ precedes $j$ in $\pi$, $S_{\pi,i} = S_{\pi',j}$ (i.e., both excluding $i$ and $j$) and thus $v_{S_{\pi,i}} = v_{S_{\pi',j}}$. It also follows from~\eqref{dodo} that by setting $C = S_{\pi,i} = S_{\pi',j}$,  $v_{{S_{\pi,i}} \cup \{i\}} = v_{C \cup \{i\}} = v_{C \cup \{j\}} = v_{{S_{\pi',j}} \cup \{j\}}$;
	(b) when $j$ precedes $i$ in $\pi$, ${S_{\pi,i} \cup \{i\}} = {S_{\pi',j} \cup \{j\}}$ (i.e., both containing $i$ and $j$) and thus $v_{{S_{\pi,i}} \cup \{i\}} = v_{{S_{\pi',j}} \cup \{j\}}$. It also follows from~\eqref{dodo} that by setting $C = S_{\pi,i} \setminus \{j\} = S_{\pi',j} \setminus \{i\}$,  $v_{S_{\pi,i}} = v_{C \cup \{j\}} = v_{C \cup \{i\}} = v_{S_{\pi',j}}$. 
    
    \item \ref{FAIR3} \textbf{Strict Desirability.}
    For all $i,j\in N$ s.t.~$i \neq j$, if
    \begin{equation}
	\begin{array}{l}
    (\exists B \subseteq N\setminus\{i,j\}\ \ v_{B \cup \{i\}} > v_{B \cup \{j\}})\ \land \vspace{1mm}\\
    (\forall C \subseteq N\setminus\{i,j\}\ \ v_{C \cup \{i\}} \geq v_{C \cup \{j\}})\ ,
    \end{array}
    \label{bobo}
    \end{equation}
    then
    \begin{align*}
	r_i = \text{Shapley}_v(i) & = \frac{1}{n!} \sum_{\pi \in \Pi_N} {\left(v_{{S_{\pi,i}} \cup \{i\}} - v_{S_{\pi,i}}\right)} \\
	& > \frac{1}{n!} \sum_{\pi' \in \Pi_N} {\left(v_{{S_{\pi',j}} \cup \{j\}} - v_{S_{\pi',j}}\right)} \\
	&= \text{Shapley}_v(j) = r_j \ . 
	\end{align*}
	Note that $\pi'$ is obtained from $\pi$ by swapping the positions of $i$ and $j$ in $\pi$. Similar to the proof of \ref{FAIR2}, we prove the inequality by considering the following two cases:
	(a) When $i$ precedes $j$ in $\pi$, $S_{\pi,i} = S_{\pi',j}$ (i.e., both excluding $i$ and $j$) and thus $v_{S_{\pi,i}} = v_{S_{\pi',j}}$. It also follows from~\eqref{bobo} that by setting $C = S_{\pi,i} = S_{\pi',j}$,  $v_{{S_{\pi,i}} \cup \{i\}} = v_{C \cup \{i\}} \geq v_{C \cup \{j\}} =  v_{{S_{\pi',j}} \cup \{j\}}$;
	(b) when $j$ precedes $i$ in $\pi$,  ${S_{\pi,i} \cup \{i\}} = {S_{\pi',j} \cup \{j\}}$ (i.e., both containing $i$ and $j$) and thus $v_{{S_{\pi,i}} \cup \{i\}} = v_{{S_{\pi',j}} \cup \{j\}}$.  It also follows from~\eqref{bobo} that 
	by setting $C = S_{\pi,i} \setminus \{j\} = S_{\pi',j} \setminus \{i\}$, $-v_{S_{\pi,i}} = -v_{C \cup \{j\}} \geq -v_{C \cup \{i\}} = -v_{S_{\pi',j}}$.
	%
	By~\eqref{bobo}, a strict inequality must hold for either case a or b.
	
	\item \ref{FAIR4} \textbf{Strict Monotonicity.}
	Let $\{v_C\}_{C \in 2^N}$ and $\{v'_C\}_{C \in 2^N}$ denote any two sets of values of data over all coalitions $C\subseteq N$,
	and $r_i$ and $r'_i$ be the corresponding values of model rewards received by party $i$. For all $i\in N$, if  
	\begin{equation}
	\begin{array}{l} 
	(\exists B \subseteq  N\setminus\{i\}\ \ v'_{B \cup \{i\}} > v_{B \cup \{i\}})
	\ \land \vspace{1mm}\\
	(\forall C \subseteq  N\setminus\{i\}\ \ v'_{C \cup \{i\}} \geq v_{C \cup \{i\}})
	\ \land \vspace{1mm}\\
	(\forall A \subseteq  N\setminus\{i\}\ \ v'_{A} = v_{A}) \land (v'_N > r_i)
	\ , 
	\end{array} 
	\label{jobo}
	\end{equation}
	then
	\begin{align*}
	r_i = \text{Shapley}_v(i) & = \frac{1}{n!} \sum_{\pi \in \Pi_N} {\left(v_{{S_{\pi,i}} \cup \{i\}} - v_{S_{\pi,i}}\right)} \\
    & = \frac{1}{n!} \sum_{\pi \in \Pi_N} {\left(v_{{S_{\pi,i}} \cup \{i\}} - v'_{S_{\pi,i}}\right)} \\
    & < \frac{1}{n!} \sum_{\pi \in \Pi_N} {\left(v'_{{S_{\pi,i}} \cup \{i\}} - v'_{S_{\pi,i}}\right)} \\
&= \text{Shapley}_{v'}(i) = r'_i \ . 
	\end{align*}
	Note that for any $\pi$, $S_{\pi,i}$ does not contain $i$. Hence, it follows from~\eqref{jobo} that $v_{S_{\pi,i}} = v'_{S_{\pi,i}}$ which results in the third equality. The inequality also follows from~\eqref{jobo}. 
\squishend
\section{Enforcing Shapley Fairness may not satisfy Individual Rationality (\ref{rational}) for Submodular Value of Data}~\label{violate}
When the value of data is submodular, for any party $i \in N$ and any coalition $C \subseteq N\setminus \{i\}$, 
${v_{i} - v_{\emptyset}} = v_{i} \geq { v_{C \cup \{i\}} - v_C}$. Then,
\begin{align}\label{sub}
\phi_i \triangleq \text{Shapley}_v(i) & = \frac{1}{n!} \sum_{\pi \in \Pi_N} {\left(v_{{S_{\pi,i}} \cup \{i\}} - v_{S_{\pi,i}}\right)} \notag\\
& \leq \frac{1}{n!} \sum_{\pi \in \Pi_N} v_i \notag\\
& = v_i\ .
\end{align}
Also, for any $i \in N$, $\phi_i \leq \sum_{i \in N}{\phi_i} = v_N$ due to the efficiency of Shapley value \cite{coop-game-theory}.

To satisfy \ref{weak-efficiency} and \ref{min-fairness}   simultaneously, we need to set $k = v_N / \phi^*$ in Definition~\ref{shapley-fair} where $\phi^*=\max_{i\in N}{\phi_i}$. 
So, for each party $i \in N$,
$$
r_i = (v_N /\phi^*) \times \phi_i \
$$
according to Definition~\ref{shapley-fair} and the party with the largest $\phi_i$ (i.e., $\phi^*$) always receives the most valuable model reward with the highest value $v_N$. 

When $v_i/\phi_i > v_N / \phi^*$, \ref{rational} will not be satisfied (i.e., $r_i = (v_N /\phi^*) \times \phi_i < v_i$).
Since $v_N / \phi^* \geq 1$, this will never happen if $v_i/\phi_i < 1$.
However, as the value of data is submodular, for each party $i \in N$, $v_i \nless \phi_i$ due to \eqref{sub}. Thus,~\ref{rational} may not be satisfied.
\section{Proof of Theorem~\ref{mi-power}} \label{a.power}
Firstly, the values $(r_i)_{i \in N}$ of model rewards satisfy non-negativity (\ref{R1}): For each party $i\in N$, $\phi_i \geq 0$ due to~\eqref{shapley} and the non-negative and monotonic  value of data using IG~\eqref{data_IG}.

Next, the values $(r_i)_{i \in N}$ of model rewards satisfy feasibility (\ref{feasibility}): For each party $i \in N$, $r_i \triangleq {\left( \phi_i /\phi^* \right) }^\rho \times v_N \leq 1^\rho \times v_N = v_N$ when $\rho \geq 0$.
Also, weak efficiency (\ref{weak-efficiency}) is satisfied since $r_i = v_N$ for the party $i$ with the largest Shapley value $\phi^*$.

When $\rho > 0$, the values $(r_i)_{i \in N}$ of model rewards are $\rho$-Shapley fair with $k = (1/\phi^*)^\rho \times v_N$ in Definition~\ref{ratio-fairness}. 
\subsection{Proof of Individual Rationality (\ref{rational})}
For each party $i \in N$, since
\[
\rho \leq \rho_r \triangleq \min_{j \in N}{\frac{\log(v_{j}/v_N)}{\log(\phi_j/\phi^*)}}  \leq \frac{\log(v_{i}/v_N)}{\log(\phi_i/\phi^*)}
\]
and $\log(\phi_i/\phi^*) \leq 0\ ,$ 
\begin{align*}
\log(v_{i}/v_N) & \leq \log{\left(\phi_i/\phi^* \right) }^\rho \\
 v_{i}/v_N  & \leq {\left( \phi_i/\phi^* \right) }^\rho \\
 v_i & \leq {\left(\phi_i/\phi^* \right) }^\rho \times v_N = r_i\ .
\end{align*}
As $r_i \geq v_i$ for each party $i \in N$, the values $(r_i)_{i \in N}$ of model rewards satisfy \ref{rational}.

For each party $i \in N$, when $\rho = 0$, $r_i = v_N$. Since the value of data is monotonic, $v_i \leq v_N$. So, $\rho_r \geq 0$. It is possible that $\rho_r > 1$.
\subsection{Proof of Stability (\ref{stability})}
%
For each party $i \in N$, since
\[
\rho \leq \rho_s \triangleq \min_{j \in N}{\frac{\log(v_{C_j}/v_N)}{\log(\phi_j/\phi^*)}}  \leq \frac{\log(v_{{C_i}}/v_N)}{\log(\phi_i/\phi^*)}
\]
and $\log(\phi_i/\phi^*) \leq 0\ ,$
\begin{align*}
\log(v_{C_i}/v_N) & \leq \log{\left(\phi_i/\phi^* \right) }^\rho \\
v_{C_i}/v_N  & \leq {\left( \phi_i/\phi^* \right) }^\rho \\
v_{C_i} & \leq {\left(\phi_i/\phi^* \right) }^\rho \times v_N = r_i
\ .
\end{align*}
As $r_i \geq v_{C_i}$ for each party $i \in N$, the values $(r_i)_{i \in N}$ of  model rewards satisfy \ref{stability}.

For each party $i \in N$, when $\rho = 0$, $r_i = v_N$. Since the value of data is monotonic, $v_i \leq v_{C_i} \leq v_N$. So, $\rho_r \geq \rho_s \geq 0$.
It is possible that $\rho_s > 1$.
\subsection{Proof of Fairness (\ref{min-fairness})}
%
Firstly, we show that the Shapley value satisfies the following {\bf party monotonicity} property,\footnote{Another solution concept that satisfies the party monotonicity property is the Banzhaf index.} which will be used in our proof of~\ref{FAIR4}: For all $i\in N$, if
\begin{equation}
\begin{array}{l} 
\displaystyle(\exists B \subseteq  N\setminus\{i\}\ \ v'_{B \cup \{i\}} > v_{B \cup \{i\}})\ \land \vspace{1mm}\\
(\forall C \subseteq  N\setminus\{i\}\ \ v'_{C \cup \{i\}} \geq v_{C \cup \{i\}})\ \land \vspace{1mm}\\
(\forall A \subseteq  N\setminus\{i\}\ \ v'_{A} = v_{A})\ ,
\end{array}
\label{qaqa}
\end{equation}
then $\forall j \in N\setminus\{i\} \ \ \phi'_i - \phi_i \geq \phi'_j - \phi_j\ .$
\begin{proof}
Let $v^\Delta_{C} \triangleq \left(v'_{C} - v_{C}\right)$ for any coalition $C\subseteq N$. 
It follows from~\eqref{shapley} that for any party $j \in N\setminus\{i\}$,
\begin{align*}
\phi'_j - \phi_j &= \frac{1}{n!} \sum_{\pi \in \Pi_{i \prec j}} {\left(v^\Delta_{S_{\pi,j} \cup \{j\}} - v^\Delta_{S_{\pi,j}}\right)} \\
&\leq \frac{1}{n!} \sum_{\pi \in \Pi_{i \prec j}} {\left(v^\Delta_{S_{\pi,j} \cup \{j\}}\right)} \\
&= \frac{1}{n!} \sum_{\pi' \in \Pi_{j \prec i}} {\left(v^\Delta_{S_{\pi',i} \cup \{i\}}\right)} \\   
&\leq \frac{1}{n!} \sum_{\pi \in \Pi_N} {\left(v^\Delta_{S_{\pi,i} \cup \{i\}}\right)} \\
&= \phi'_i - \phi_i 
\end{align*}
where $\Pi_{i \prec j}$ is a subset of $\Pi_N$ containing all permutations with $i$ preceding $j$.
We consider only the subset $\Pi_{i \prec j}$ of $\Pi_N$ in the first equality due to~\eqref{qaqa} where 
the quality of a model trained on the aggregated data of a coalition can only change/increase when
containing party $i$.
The first inequality
is also due to~\eqref{qaqa} where $v'_C \geq v_C$ for every $C \subseteq N$ and thus $v^\Delta_{S_{\pi,j}} \geq 0$. 
%
The second equality
can be understood by swapping the positions of $i$ and $j$ in permutation $\pi$ to obtain $\pi'$ such that $S_{\pi,j} \cup \{j\} = S_{\pi',i} \cup \{i\}$.
%
The second inequality
holds because there may exist a permutation $\pi$ where $i \prec j\ \text{in}\ \pi$ (i.e., $j \notin S_{\pi,i}$) and $v^\Delta_{S_{\pi,i} \cup \{i\}} \geq 0$. 
The last equality
holds since $v^\Delta_{S_{\pi,i}} = 0$ due to~\eqref{qaqa} where $\forall A \subseteq  N\setminus\{i\}\ \ v'_{A} = v_{A}$.
\end{proof}
%

Let us now consider $r_i \triangleq {\left( \phi_i /\phi^* \right) }^\rho \times v_N$.
We will prove the following properties for \ref{min-fairness}: 
\squishlisttwo
	\item \ref{FAIR1} \textbf{Uselessness.}
	\[
	\begin{array}{l} 
	\displaystyle
	(\forall C\subseteq N \setminus \{i\}\ \ v_{C \cup \{i\}} = v_C)\ \vspace{1mm}\\ 
	\Rightarrow \phi_i = 0 \ \vspace{1mm}\\ 
	\Rightarrow r_i = 0 \ . 
	\end{array} 
	\]
	
	\item \ref{FAIR2} \textbf{Symmetry.} Since each $\phi_i^\rho$ is multiplied by the same factor $k = v_N/{\phi^*}^\rho$,
	\[
	\begin{array}{l} 
	\displaystyle
	(\forall C \subseteq N\setminus\{i,j\}\ \ v_{C \cup \{i\}} = v_{C \cup \{j\}})\ \vspace{1mm}\\  
	\Rightarrow \phi_i = \phi_j \vspace{1mm}\\ 
	\Rightarrow r_i = r_j \ .
	\end{array} 
	\]
    
    \item \ref{FAIR3} \textbf{Strict Desirability.} Since each $\phi_i^\rho$ is multiplied by the same factor $k = v_N/{\phi^*}^\rho$ and $k > 0$,
	\[
	\begin{array}{l} 
	\displaystyle
	(\exists B \subseteq N\setminus\{i,j\}\ \ v_{B \cup \{i\}} > v_{B \cup \{j\}})\ \land \vspace{1mm}\\
    (\forall C \subseteq N\setminus\{i,j\}\ \ v_{C \cup \{i\}} \geq v_{C \cup \{j\}}) \vspace{1mm}\\
	\Rightarrow \phi_i > \phi_j\ \vspace{1mm}\\ 
	\Rightarrow r_i > r_j \ .
	\end{array} 
	\]
	The first implication in the proof of \ref{FAIR1}, \ref{FAIR2}, and \ref{FAIR3} follows from Proposition~\ref{p1}.

    \item \ref{FAIR4} \textbf{Strict Monotonicity.} Let $\{v_C\}_{C \in 2^N}$ and $\{v'_C\}_{C \in 2^N}$ denote any two sets of values of data over all coalitions $C\subseteq N$,
	and $\phi_i$ and $\phi'_i$ be the corresponding Shapley values of party $i$.
    Let $\ell\in N$ be the party with the largest Shapley value $\phi'_\ell$ based on $\{v'_C\}_{C \in 2^N}$.
    For any party $i \in N$ satisfying~\eqref{jobo} (i.e., premise of \ref{FAIR4}), $\phi'_\ell \geq \phi'_i$ due to definition of party $\ell$ and $\phi'_i - \phi_i \geq \phi'_\ell - \phi_\ell$ due to party monotonicity property when $\ell \in N\setminus\{i\}$. Together, it follows that $\phi_\ell \geq \phi_i$. Then,
    \begin{align}
    \phi'_i - \phi_i &\geq \phi'_\ell - \phi_\ell \vspace{1mm}\notag\\
    \phi_\ell(\phi'_i - \phi_i) &\geq \phi_i(\phi'_\ell - \phi_\ell) \vspace{1mm}\notag\\
    \phi_\ell \times \phi'_i &\geq \phi_i \times \phi'_\ell \vspace{1mm}\notag\\
    \phi'_i/\phi'_\ell &\geq \phi_i/\phi_\ell
    \ . \label{haha}
    \end{align}
    If $\phi_\ell > \phi_i$, then $\phi'_i/\phi'_\ell > \phi_i/\phi_\ell$ instead. The equality in~\eqref{haha} holds only if $\phi_\ell = \phi_i$. Let $p\in N$ be the party with the largest Shapley value $\phi_p$ based on $\{v_C\}_{C \in 2^N}$. Then, 
	\[
	\displaystyle
	\begin{array}{l} 
	r_i' \\
	\displaystyle = {\left( {\phi_i'}/ {\phi'_\ell} \right) }^\rho \times v'_N \vspace{1mm}\\
	\displaystyle \geq {\left(  {\phi_i'}/ {\phi'_\ell} \right) }^\rho \times v_N \vspace{1mm}\\
	\displaystyle \geq {\left(  {\phi_i}/ {\phi_\ell} \right) }^\rho \times v_N \vspace{1mm}\\
	\displaystyle \geq {\left( {\phi_i}/ {\phi_p} \right) }^\rho \times v_N \\
	= r_i
	\end{array} 
	\]
	where the first inequality follows from~\eqref{jobo}, the second inequality is due to~\eqref{haha}, and the last inequality is due to the definition of party $p$. 
	For equality in $r'_i \geq r_i$ to possibly hold, it has to be the case that $\phi_i = \phi_\ell = \phi_p$ and $v'_N = v_N$. Based on the definitions of $r_i$ and $r'_i$ in Theorem~\ref{mi-power}, this implies $r'_i = v'_N =v_N= r_i$ which does not satisfy $v'_N > r_i$ in~\eqref{jobo} (i.e., premise of \ref{FAIR4}). For all other cases, $r'_i > r_i$.
\squishend
\section{Details of Experimental Settings}
\label{detail}
In all the experiments, we optimize $\eta_i$ for $i \in N$ (Sec.~\ref{noise}) by finding the root of $\mathbb{I}(\boldsymbol{\theta}; D_i^r) - r^*_i$ using a root-finding algorithm called the TOMS Algorithm $748$ method.\footnote{Alefeld, G. E., Potra, F. A., and Shi, Y. Algorithm $748$: Enclosing zeros of continuous functions. \emph{ACM Trans. Math. Softw.}, $21(3)$, $327$--$344$, $1995$.}
\subsection{Gaussian Process (GP) Regression with Synthetic Friedman  Dataset}\label{synthetic}
The following Friedman function is used in our experiment:
\[
\begin{array}{rl}
  y = \hspace{-2.4mm} & 10\sin(\pi \vect{x}_{[d,0]} \vect{x}_{[d,1]}) + 20 (\vect{x}_{[d,2]} - 0.5)^2 \vspace{1mm}\\
  \hspace{-2.4mm} & + 10\vect{x}_{[d,3]} +5\vect{x}_{[d,4]} + 0\vect{x}_{[d,5]} + \mathcal{N}(0, 1)
\end{array}
\]
where $d$ is the index of data point $(\vect{x},y)$ and 
its $6$ input features (i.e., $\vect{x}\triangleq(\vect{x}_{[d,a]})_{a=0,\ldots,5}$) are independent and uniformly distributed over the input domain $[0, 1]$.
The last input feature $\vect{x}_{[d,5]}$ is an independent variable which does not affect the output $y$.
We standardize the values of output $y$ and train a GP model\footnote{For all GP models, we use \emph{automatic relevance determination} such that each input feature has a different lengthscale parameter.} with a squared exponential kernel. 

For each tested value of $\rho$, we reset the random seed to a fixed value and draw $20$ samples of $\mathbf{z}_i$ from $\mathcal{N}(\mathbf{0}, \eta_i\mathbf{I})$ with an optimized $\eta_i$ (Sec.~\ref{noise}). Though $\eta_i$ differs for varying $\rho$,
the fixed random seed makes the \emph{standardized} samples of $\mathbf{z}_i$ the same across all values of $\rho$, thus allowing a fair comparison.
We evaluate the performance of our reward scheme on a test dataset which consists of $800$ points generated from the Friedman function.
\subsection{Gaussian Process (GP) Regression with DiaP Dataset}
We remove the gender feature and standardize the values of both input matrix $\vect{X}$ and output vector $\vect{y}$.
We train a GP model with a composite kernel comprising the squared exponential kernel and the exponential kernel.

We consider $5$ different train-test (i.e., $80\%$-$20\%$ at random) splits. For each split, we create $10$ different partitions of the training dataset using the partitioning algorithm detailed in Sec.~\ref{experi}.
\subsection{Neural Network (NN) Regression with CaliH Dataset}\label{setting}
We first standardize the values of both input matrix $\vect{X}$ and output vector $\vect{y}$. Then, we use $60\%$ of the CaliH dataset to train a NN with $2$ dense and fully connected hidden layers of $100$ and $50$ units and use the \emph{rectified linear unit} (ReLU) as the activation function.
Subsequently, we perform BLR on the outputs from the last hidden layer.

We consider $5$ different train-test (i.e., $80\%$-$20\%$ at random) splits. For each split, we create $10$ different partitions of the training dataset using the partitioning algorithm detailed in Sec.~\ref{experi}.
\subsection{Sparse GP Regression with Synthetic Friedman Dataset (10 Parties)}
We consider a larger synthetic Friedman dataset of size $5000$ and $3$ different train-test (i.e., $80\%$-$20\%$ at random) splits. For each split, we create $10$ different partitions of the training dataset using the partitioning algorithm detailed in Sec.~\ref{experi}. But, we divide the training dataset into $10$ consecutive blocks instead with the constraint that each party owns at least $5\%$ of the dataset. 

As in Appendix~\ref{synthetic}, we standardize the values of output $y$.
Since the exact Shapley value \eqref{shapley} is expensive to evaluate with a large number of parties, we approximate it using $3000$ samples generated by the \emph{simple random sampling} algorithm \cite{maleki2013}.
\subsection{Calculating IG for BLR and GP Models}
Let $D \triangleq (\vect{X}, \vect{y})$ be the training data where $\vect{X}$ is an input matrix 
and $\vect{y}$ ($\vect{f}$) is the corresponding noisy (latent) output vector such that $\vect{y}$ is perturbed from $\vect{f}$ by an additive Gaussian noise with noise variance $\sigma^2$.

For BLR with model parameters/weights $\boldsymbol{\theta}$ and prior belief $p(\boldsymbol{\theta}) \triangleq \mathcal{N}(\mathbf{0}, \boldsymbol{\Sigma}_0)$, the IG on $\boldsymbol{\theta}$ given $D$ is
\begin{equation*}
  \mathbb{I}(\boldsymbol{\theta};D) = 0.5 \log(|\vect{I} + \vect{\Sigma}_0 \vect{X}^\top\vect{X}/\sigma^2|)\ .
\end{equation*}

We also consider modeling an unknown/latent function $f$ with a full-rank GP and setting $\boldsymbol{\theta} = f$ such that the latent output vector $\vect{f}$ comprises components $f(\vect{x})$ for all $\vect{x}$ in $\vect{X}$.
Then, the IG on $f$ given $D$ is
\begin{equation}
  \mathbb{I}(f; D) 
  =
  \mathbb{I}(\vect{f}; D) = 0.5 \log(|\vect{I} + K_{\vect{X}\vect{X}}/\sigma^2|)
\label{gross}  
\end{equation}
where 
$K_{\vect{X}\vect{X}}$ is a covariance matrix with components $k(\vect{x},\vect{x}')$ for all $\vect{x}, \vect{x}'$ in $\vect{X}$ and $k$ is a kernel function.
When we use a sparse GP model\footnote{We use a sparse GP model called the \emph{deterministic
training conditional} (DTC) approximation of the GP model.} with inducing input matrix $\vect{U}$, we replace $K_{\vect{X}\vect{X}}$ in~\eqref{gross} with 
$K^{\top}_{\vect{X}\vect{U}} K_{\vect{U}\vect{U}}^{-1} K_{\vect{U}\vect{X}}^{}$ where $K_{\vect{U}\vect{U}}$ is a covariance matrix with components $k(\vect{x},\vect{x}')$ for all $\vect{x}, \vect{x}'$ in $\vect{U}$ and $K_{\vect{U}\vect{X}}$ is a covariance matrix with components $k(\vect{x},\vect{x}')$ for all $\vect{x}$ in $\vect{U}$ and $\vect{x}'$ in $\vect{X}$.

When calculating the IG for heteroscedastic data (i.e., data points with different noise variances), instead of dividing by $\sigma^2$, we multiply by the 
diagonal matrix $\vect{K}_\text{noise}^{-1}$ such that 
each diagonal component of $\vect{K}_\text{noise}$ represents the noise variance corresponding to 
a data point in $D$.
\section{Additional Experimental Results}\label{more}
\subsection{Additional Results for Synthetic Friedman and BosH Datasets}\label{ap.bosh}
\textbf{GP regression with synthetic Friedman dataset.}
We also evaluate the performance of our reward scheme on a synthetic Friedman dataset with $2000$ data points.
We standardize the values of output $y$ and train a GP model with a squared exponential kernel. 
We consider $5$ different train-test (i.e., $80\%$-$20\%$ at random) splits. For each split, we create $10$ different partitions of the training dataset using the partitioning algorithm detailed in Sec.~\ref{experi}.

\begin{figure}
        \begin{tabular}{@{}c@{\hspace{1.5mm}}c@{}} \includegraphics[width=0.49\columnwidth]{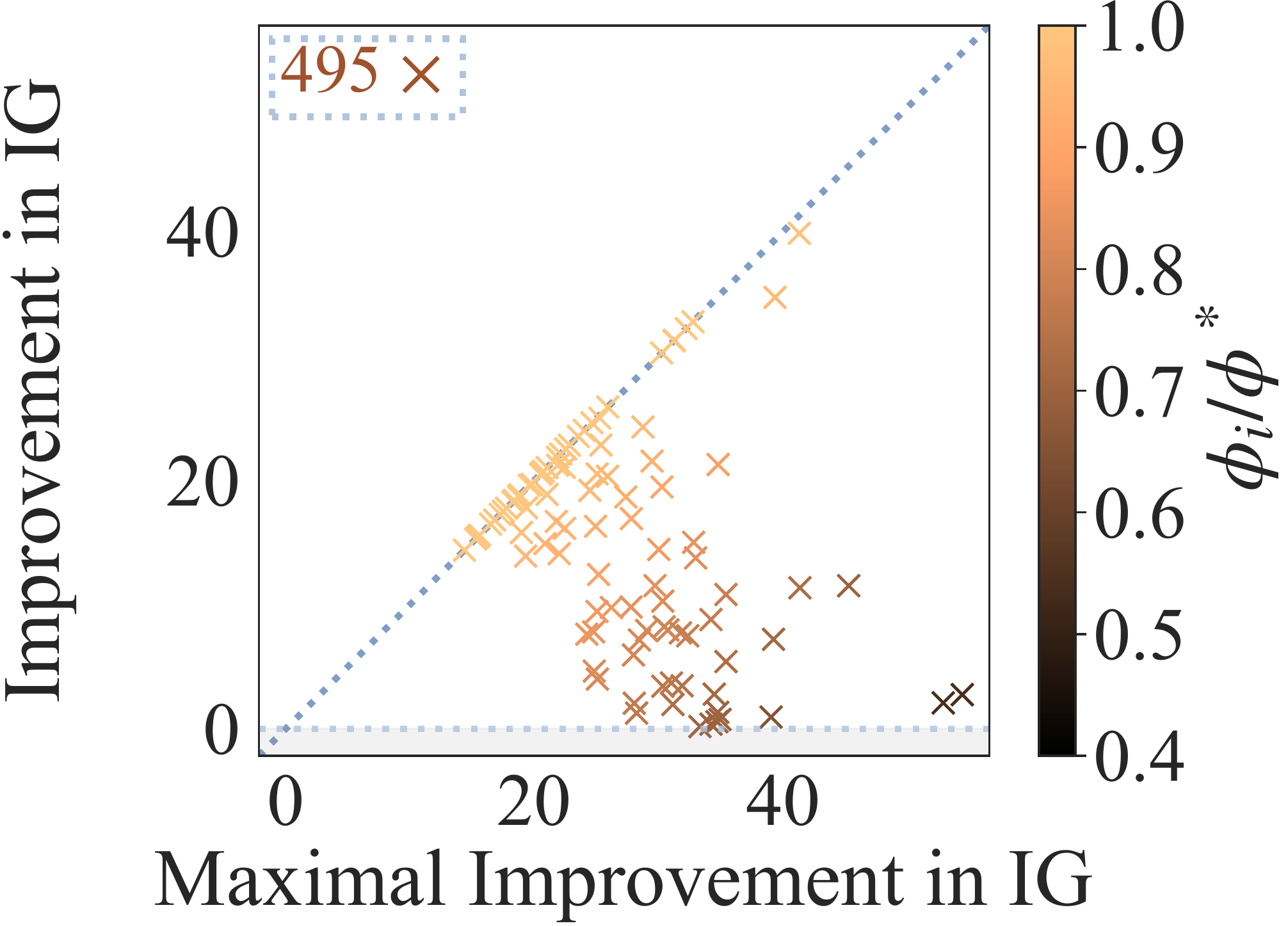} &  \includegraphics[width=0.49\columnwidth]{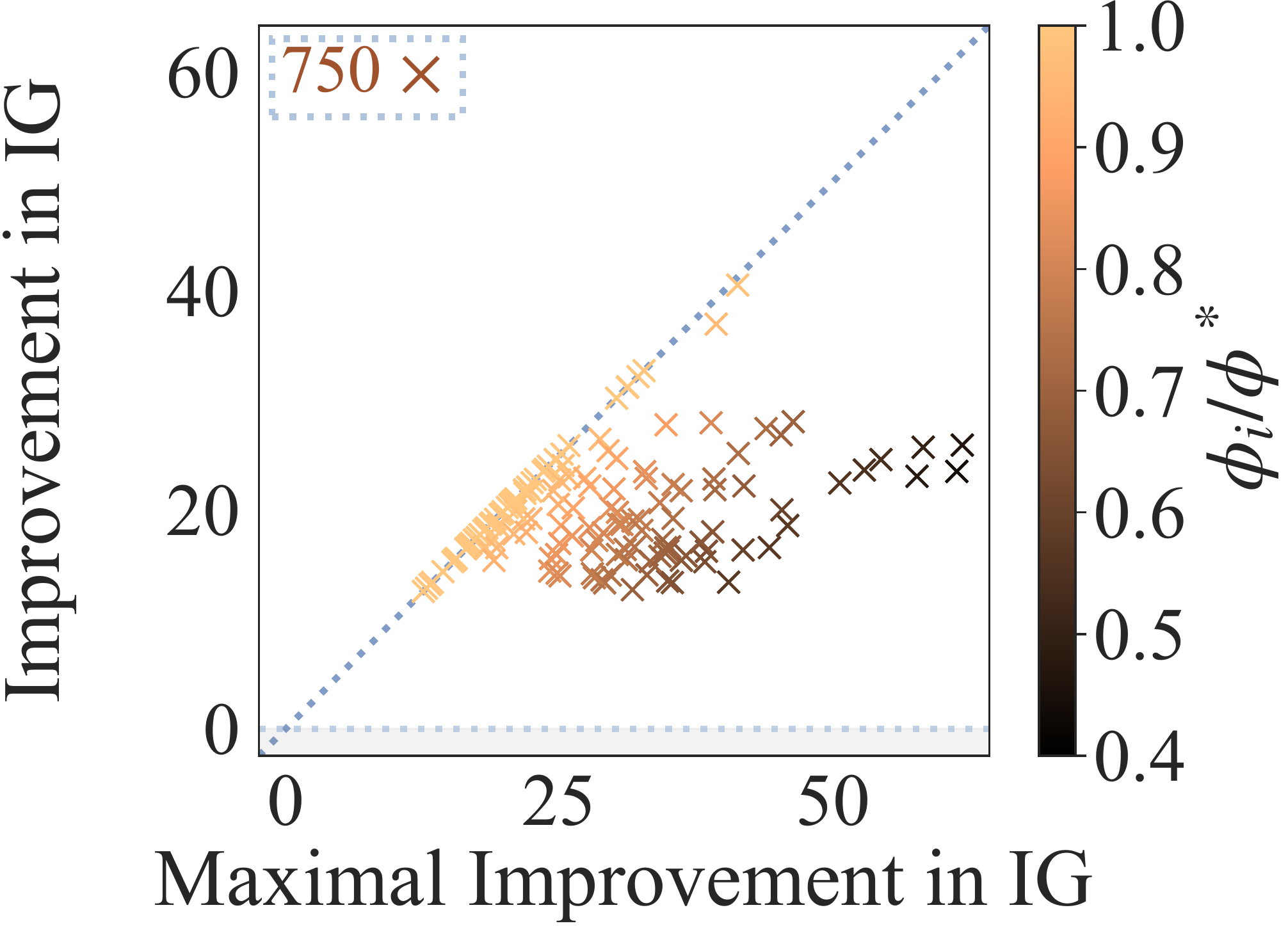} \\
        {(a) IG, $\rho = 1$} & {(b) IG, $\rho = 0.5$} \vspace{1mm}\\
        \includegraphics[width=0.49\columnwidth]{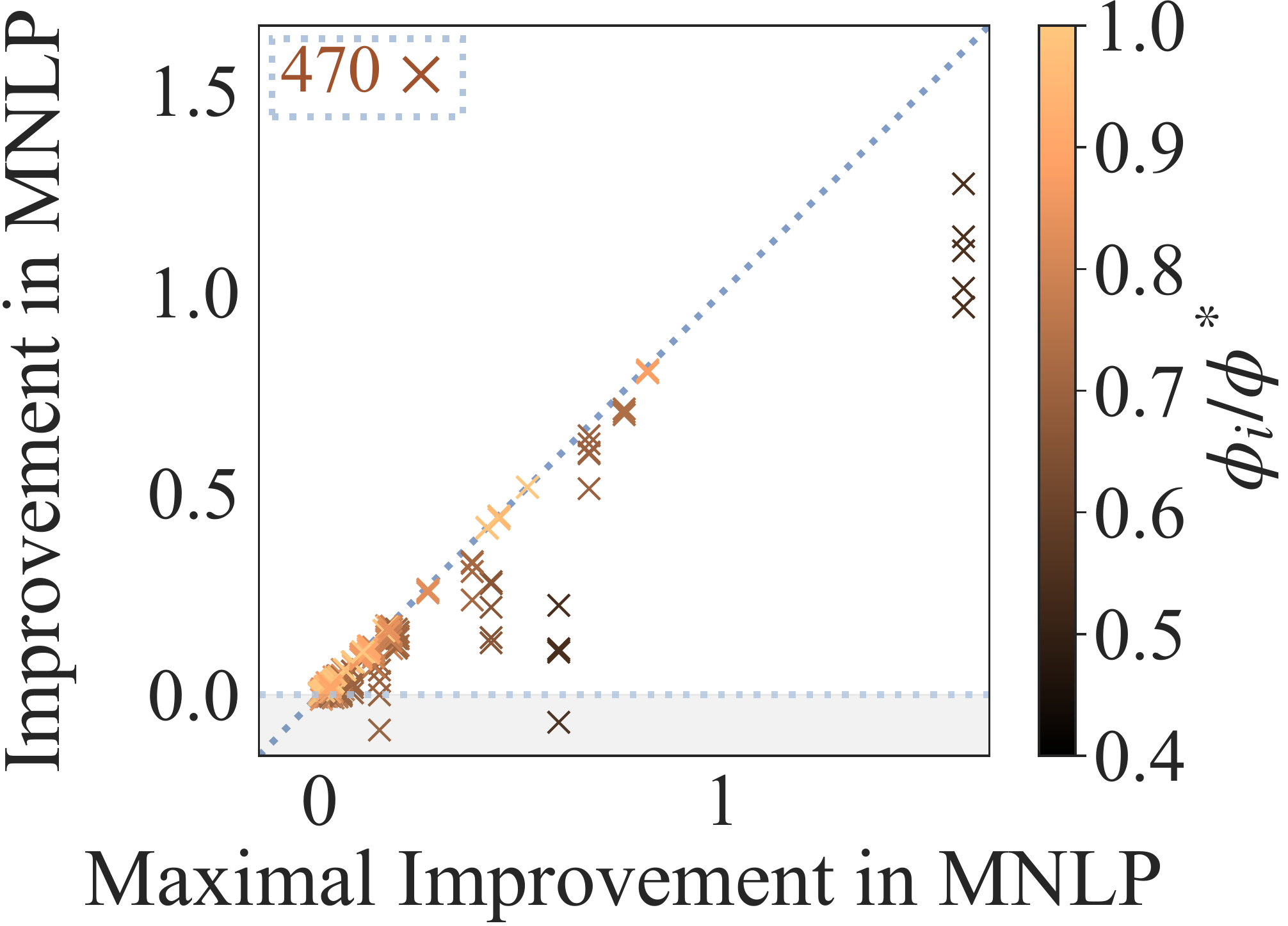} &  \includegraphics[width=0.49\columnwidth]{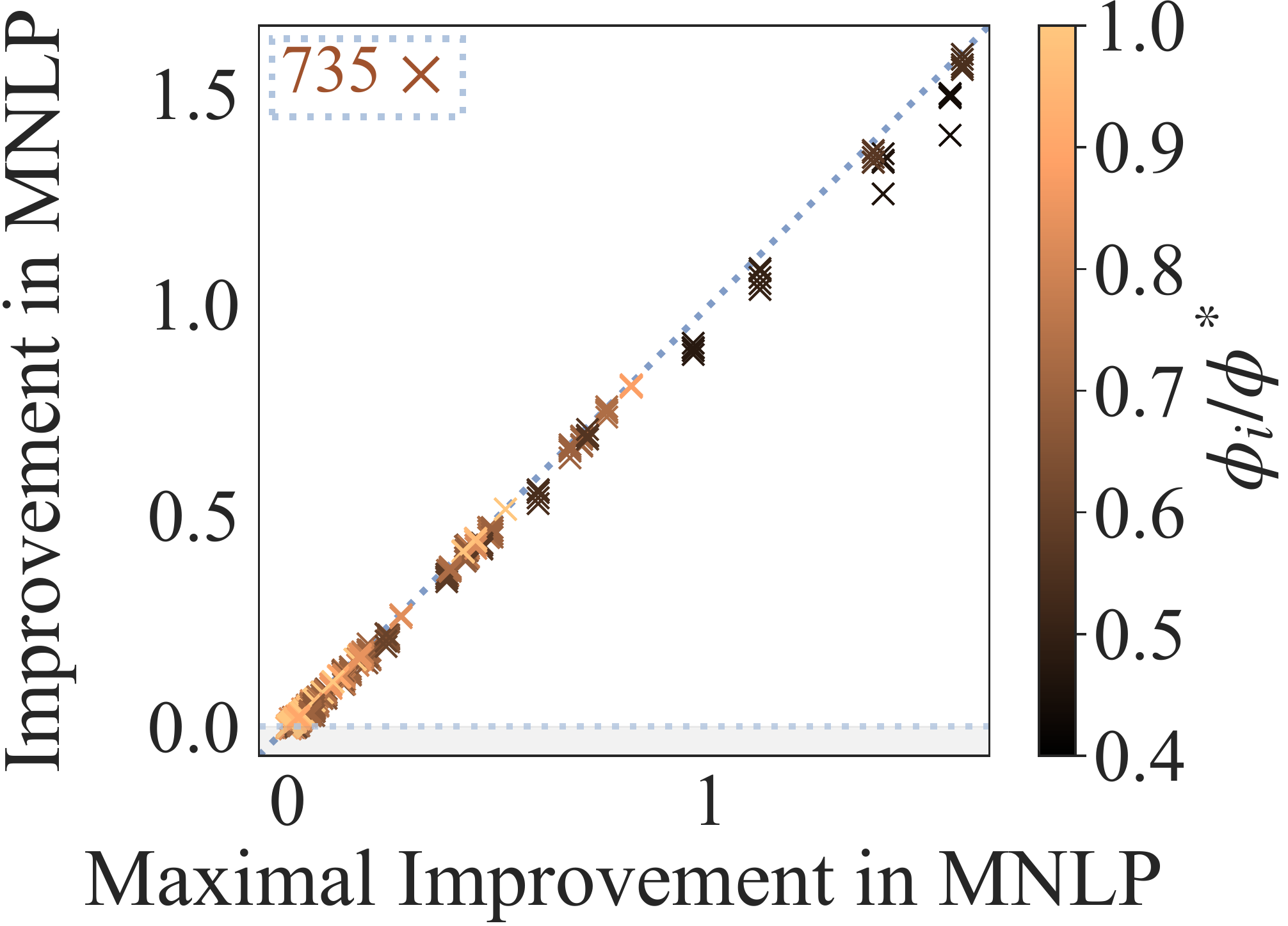} \\
        {(c) MNLP, $\rho = 1$} & {(d) MNLP, $\rho = 0.5$}
        \end{tabular}
        \caption{Scatter graph of the (a-b) improvement in IG ($r_i - v_i$) vs.~maximal improvement in IG ($v_N - v_i$) and (c-d) improvement in MNLP vs.~maximal improvement in MNLP when $\rho=1, 0.5$ for multiple partitions of Friedman dataset among $n=3$ parties.
        }
        \label{fig:friedman-gp}
\end{figure}

The results are shown in Fig.~\ref{fig:friedman-gp}. 
As before, the improvement in MNLP is usually positive. When $\rho=0.5$, most points move closer to the diagonal line, which implies that parties with smaller $\phi_i$ can now receive more valuable model rewards with higher predictive accuracy.

\textbf{NN regression with BosH dataset.}
The \emph{Boston housing} (BosH) dataset contains the value of $506$ houses with $10$ selected input features \cite{boston-dataset}.
We train a NN with $2$ dense and fully connected hidden layers of $50$ units each and use the \emph{rectified linear unit} (ReLU) as the activation function.
Subsequently, we perform BLR on the outputs from the last hidden layer.

We remove $3$ input features (i.e., proportion of residential land zoned for lots over $25,000$ sq.~ft., proportion of non-retail business acres per town, and the Charles River dummy variable) and standardize the values of both input matrix $\vect{X}$ and output vector $\vect{y}$.
Then, we use $80\%$ of the BosH dataset to train the first $2$ layers of the NN and divide the entire BosH dataset into a test dataset and among $3$ parties using our partitioning algorithm.
Due to the small data size, there is overlap between the data used to train the ``public'' NN layers and the data that is divided among parties.

\begin{figure}
    \begin{tabular}{@{}c@{\hspace{1.5mm}}c@{}}  
    \includegraphics[width=0.49\columnwidth]{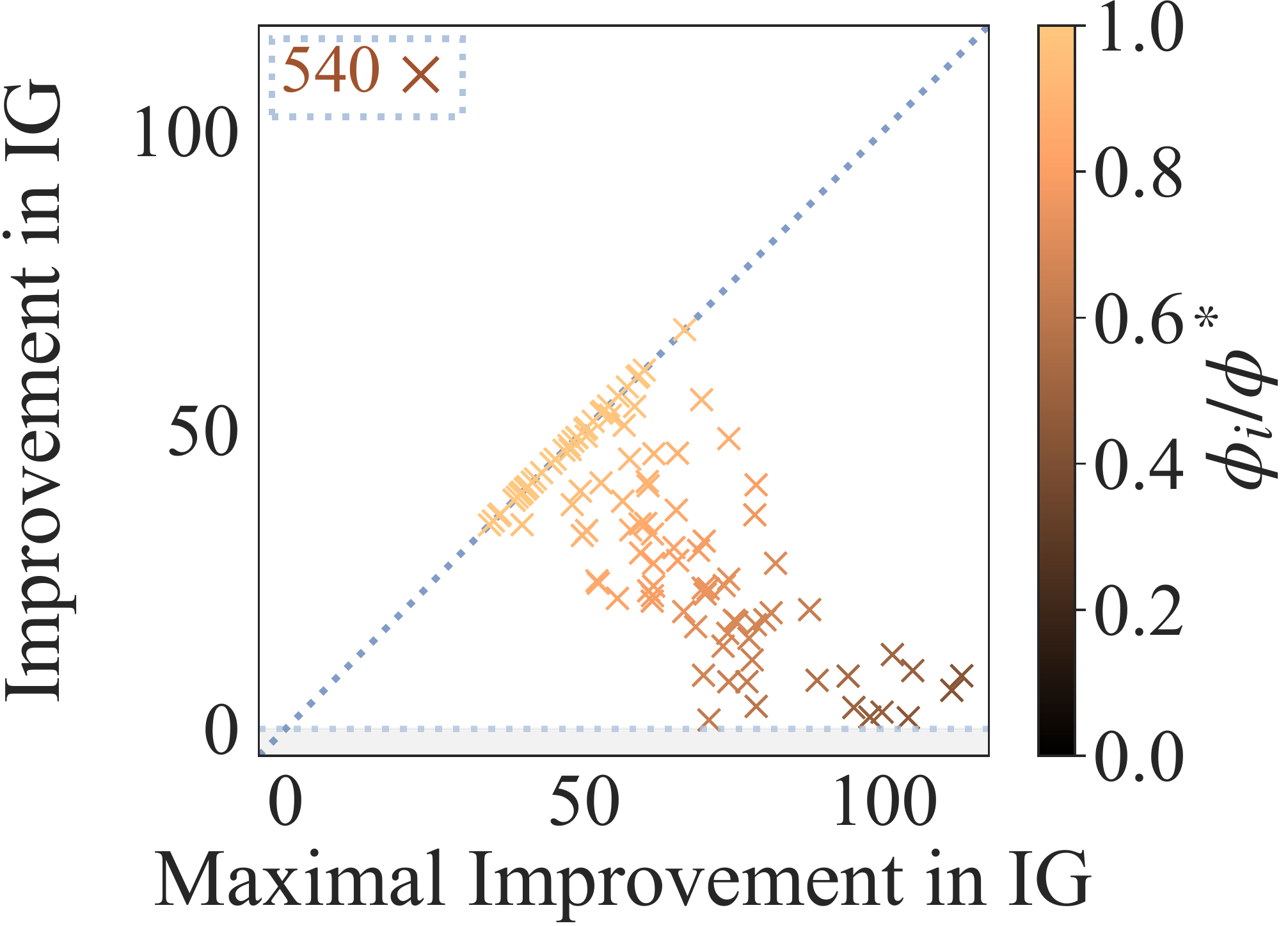} & \includegraphics[width=0.49\columnwidth]{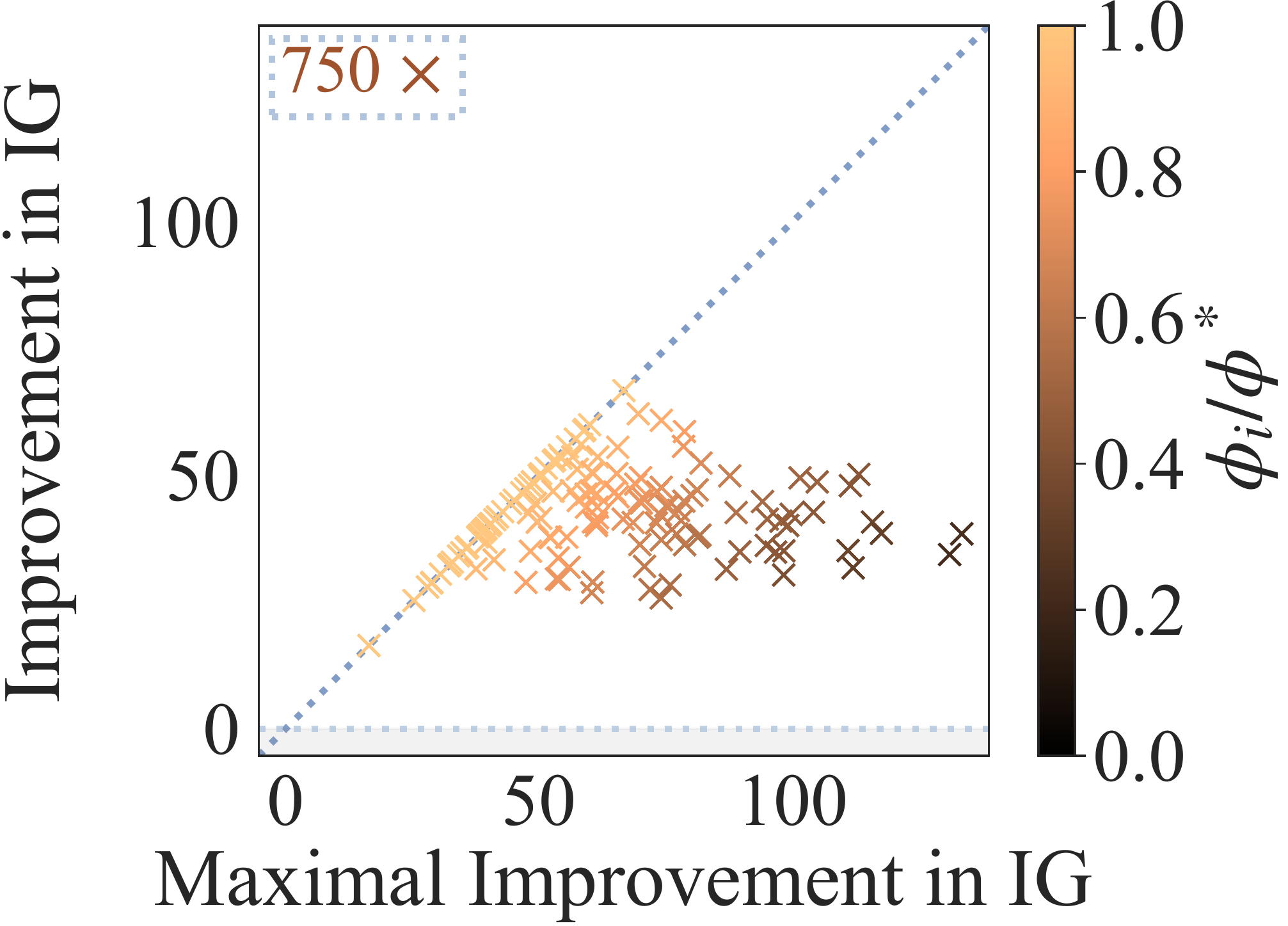} \\
    {(a) IG, $\rho = 1$} & {(b) IG, $\rho = 0.5$} \vspace{1mm}\\
    \includegraphics[width=0.49\columnwidth]{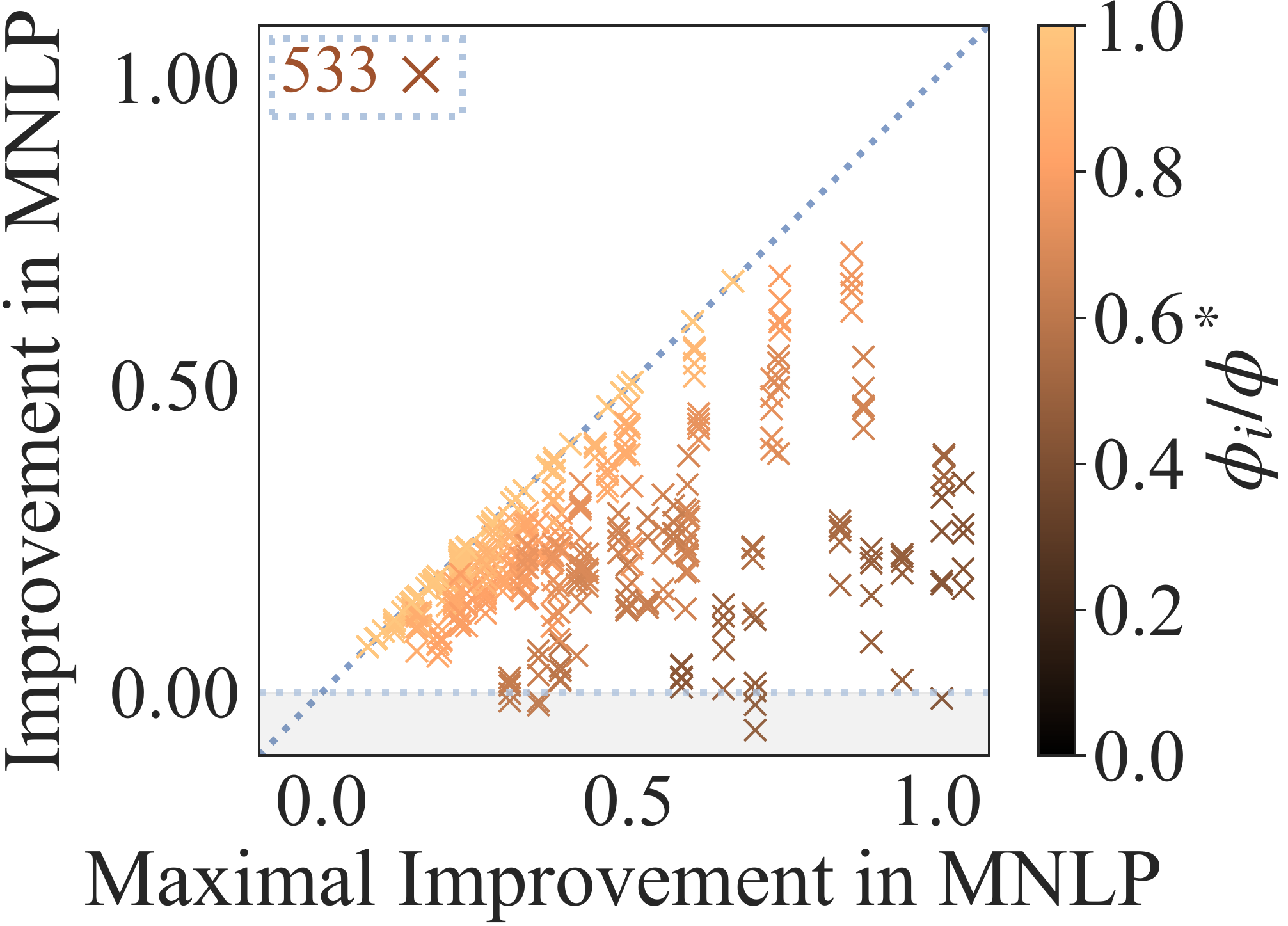} & \includegraphics[width=0.49\columnwidth]{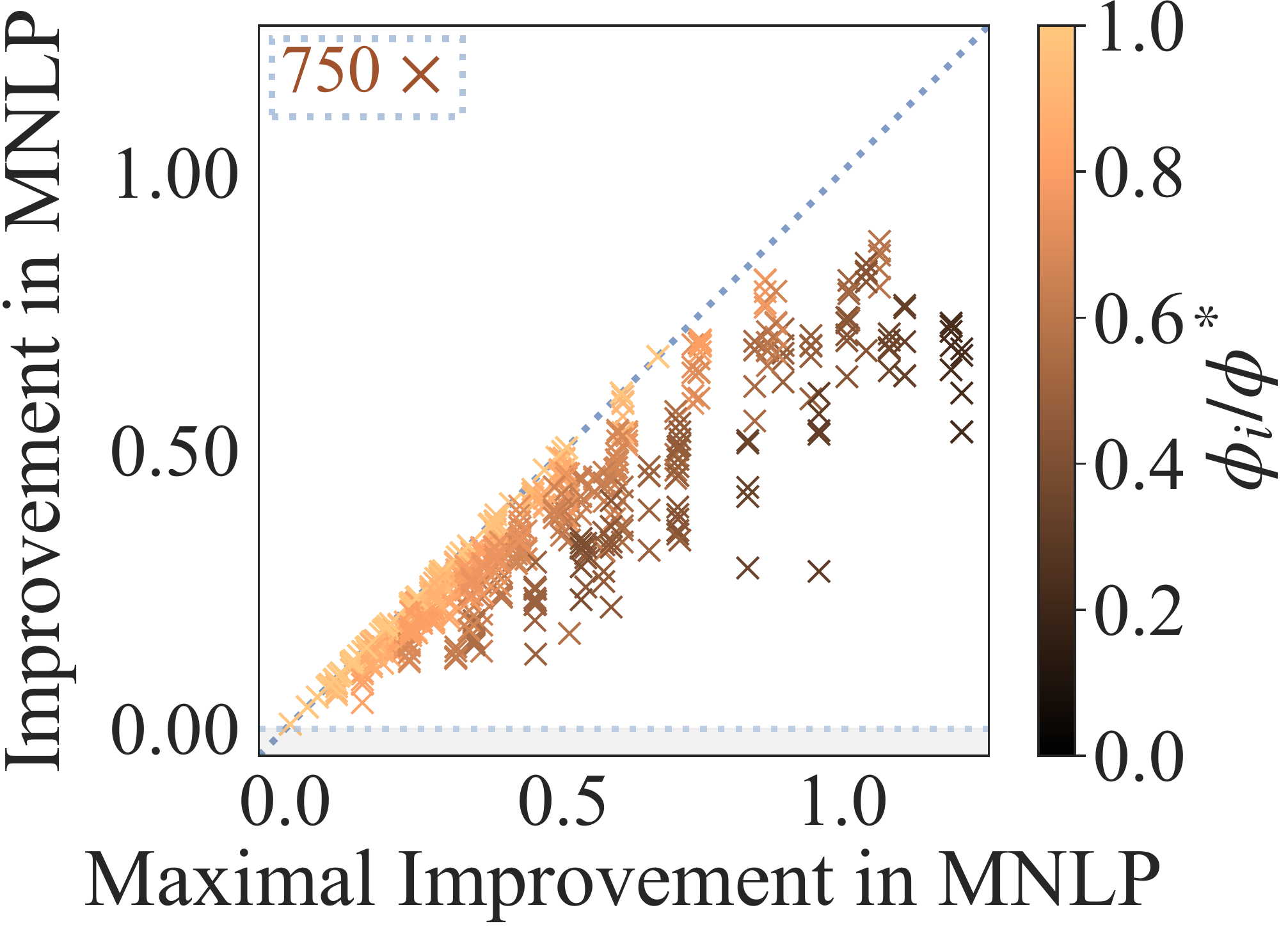} \\
    {(c) MNLP, $\rho = 1$} & {(d) MNLP, $\rho = 0.5$}
    \end{tabular}
    \caption{Scatter graph of the (a-b) improvement in IG ($r_i - v_i$) vs.~maximal improvement in IG ($v_N - v_i$) and (c-d) improvement in MNLP vs.~maximal improvement in MNLP when $\rho=1, 0.5$ for multiple partitions of BosH dataset among $n=3$ parties.
    }
    \label{fig:boston}
\end{figure}

Fig.~\ref{fig:boston} shows results of our reward scheme for multiple partitions of BosH dataset among $n=3$ parties. The same observations follow: Most model rewards achieve positive improvement in MNLP. 

When a smaller $\rho$ is used (i.e., Figs.~\ref{fig:boston}b and~\ref{fig:boston}d) or when the parties have higher Shapley values relative to the others (i.e., lighter color), they will receive more valuable model rewards which translates to lower MNLP (i.e., points lie closer to the diagonal identity line).
\subsection{Empirical Analysis of Relationship between IG and MNLP}\label{ap.ig-mnlp}
In this work, our proposed data valuation method and reward scheme are based on IG. Our experimental results have shown that IG is closely related to the predictive accuracy (i.e., MNLP) of a trained model. 
A model with higher IG usually achieves lower MNLP as it produces smaller predictive variances (i.e., related to the first term of~\eqref{mnlp}).
However, the relationship between IG and MNLP is not strictly monotonic as it is also affected by the second term of \eqref{mnlp}, i.e., the scaled squared error.
In this subsection, we will demonstrate some poor use cases of our scheme where improvement in IG fails to translate to higher predictive accuracy of the trained model more frequently.

\begin{figure}
    \begin{tabular}{@{}c@{\hspace{1.5mm}}c@{}}                               \includegraphics[width=0.49\columnwidth]{Diabetes-GP-fixed-p10-MNLP.pdf} & \includegraphics[width=0.49\columnwidth]{Diabetes-GP-fixed-p05-MNLP.pdf} \\
    {(a) $\rho = 1$, suitable model} & {(b) $\rho = 0.5$, suitable model} \vspace{1mm}\\
    \includegraphics[width=0.49\columnwidth]{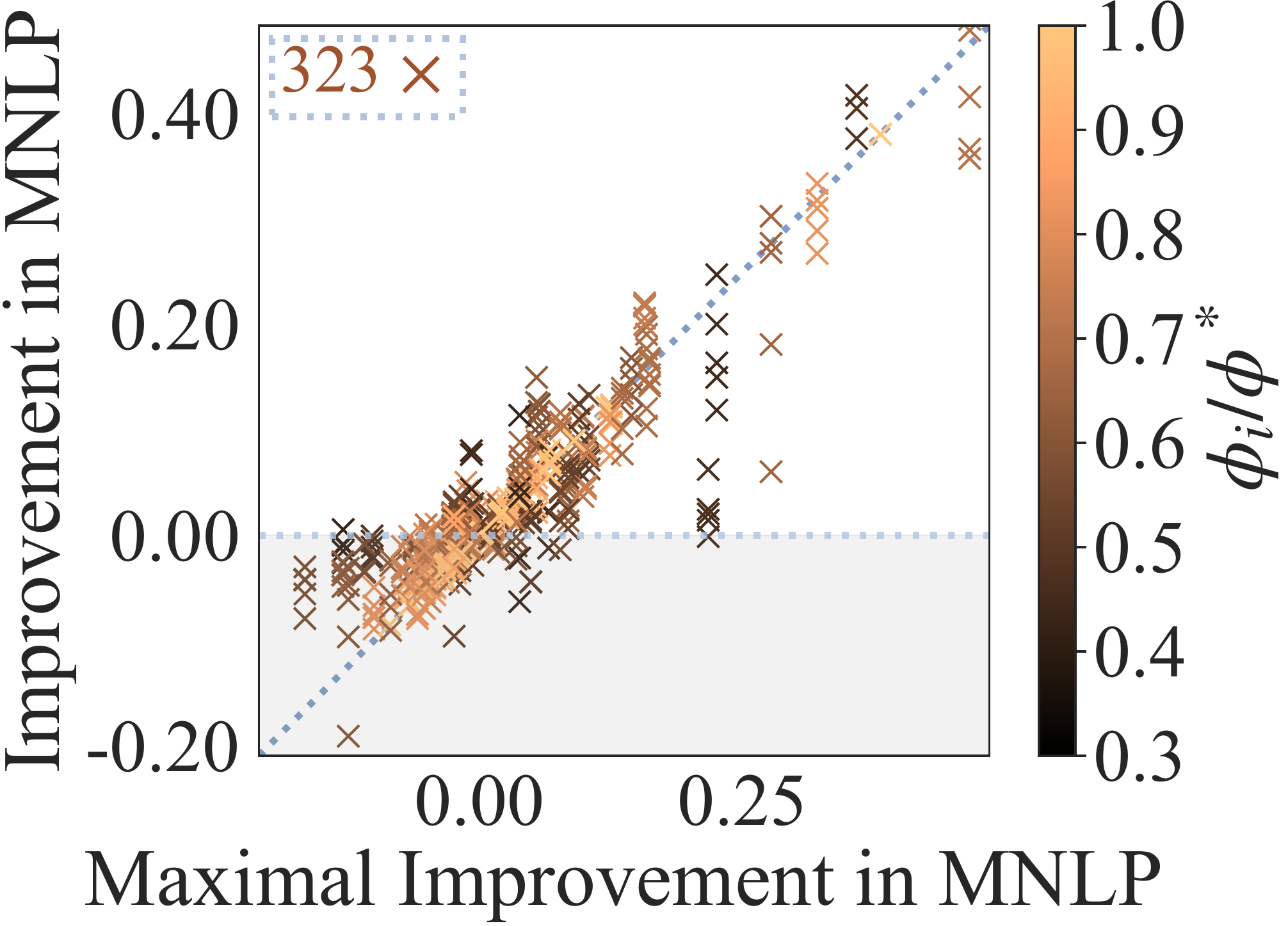} & \includegraphics[width=0.49\columnwidth]{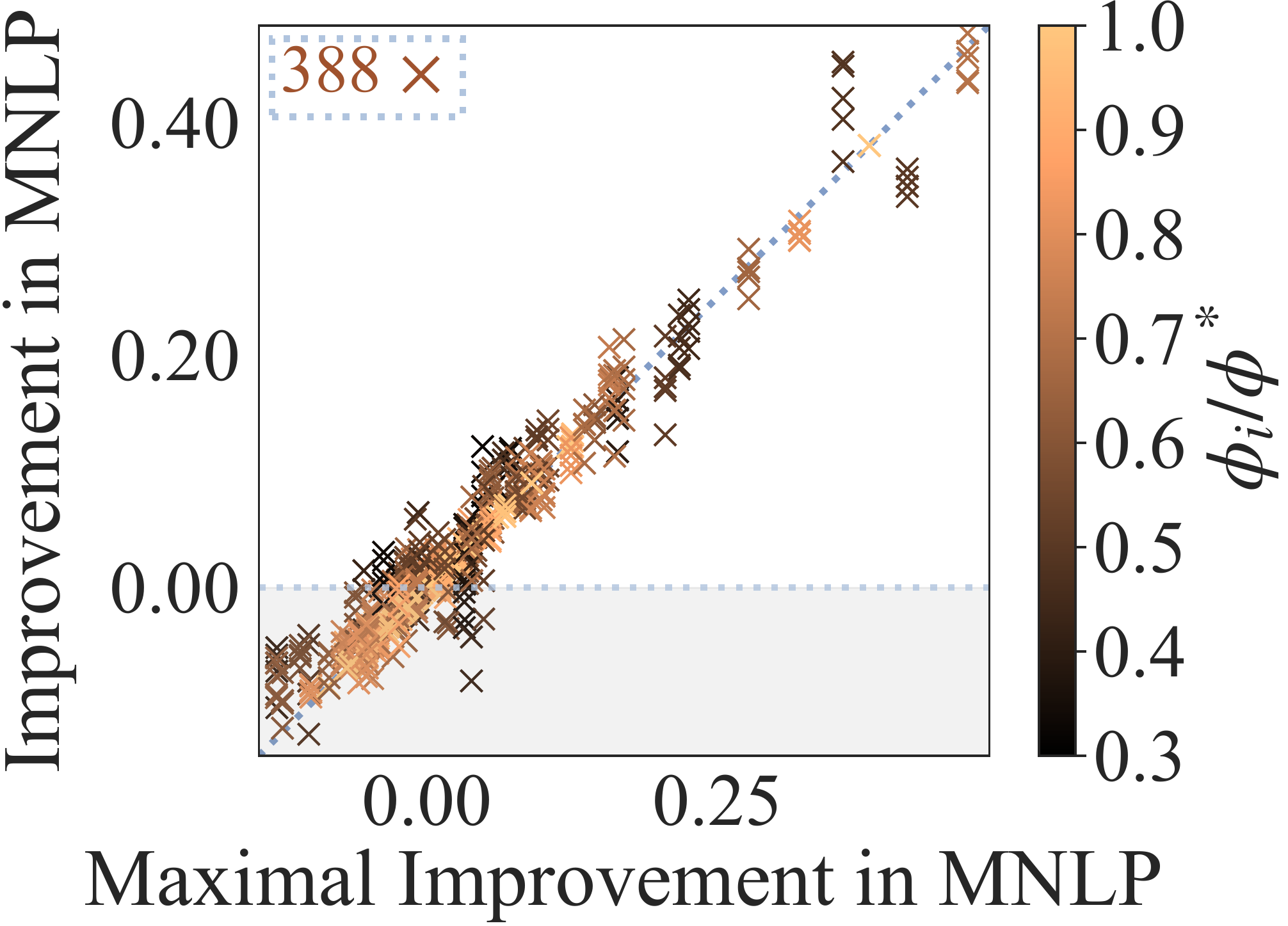} \\
    {(c) $\rho = 1$, unsuitable model} & {(d) $\rho = 0.5$, unsuitable model}
    \end{tabular}
    \caption{Scatter graph of the improvement in MNLP vs.~maximal improvement in MNLP when $\rho=1, 0.5$ for (a-b) a suitable GP model and (c-d) an unsuitable GP model for multiple partitions of DiaP dataset among $n=3$ parties.
    }
    \label{fig:bad-diab}
\end{figure}

The first poor use case is when an unsuitable model is selected.
To demonstrate this, we train an alternative GP regression model with an unsuitable squared exponential kernel on the DiaP dataset. 
Even when ample training data is provided, the squared error remains high relative to the variance in the data.
The results achieved using this unsuitable GP model are shown in Figs.~\ref{fig:bad-diab}c-d. 
For ease of comparison, we copy the results of the GP model with a suitable kernel in Figs.~\ref{fig:diabetes}c-d to Figs.~\ref{fig:bad-diab}a-b.
We can observe that the maximal and actual improvement in MNLP are mostly positive for the suitable model (Figs.~\ref{fig:bad-diab}a-b).
However, for the unsuitable model, fewer positive values of the maximal and actual improvement in MNLP can be observed in Figs.~\ref{fig:bad-diab}c-d.
In Figs.~\ref{fig:bad-diab}b and~\ref{fig:bad-diab}d, there are, respectively, $631$ and $388$ points with positive improvement in MNLP (out of a maximum of $750$ points).
Improvement in IG fails to translate to a higher predictive accuracy of its trained model more frequently when an unsuitable model is selected.

\begin{figure}
    \begin{tabular}{@{}c@{\hspace{1.5mm}}c@{}} 
    \includegraphics[width=0.49\columnwidth]{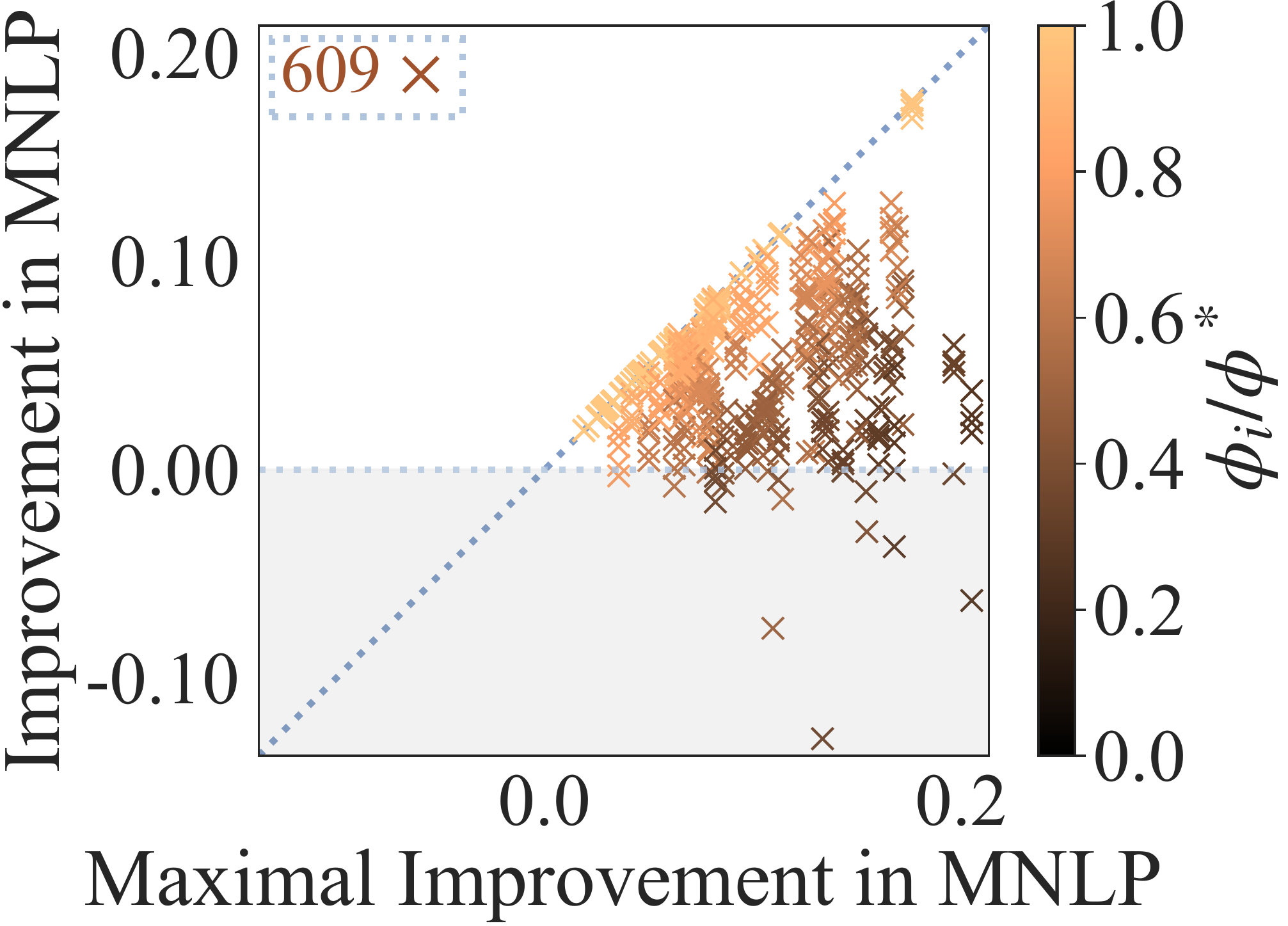} & \includegraphics[width=0.49\columnwidth]{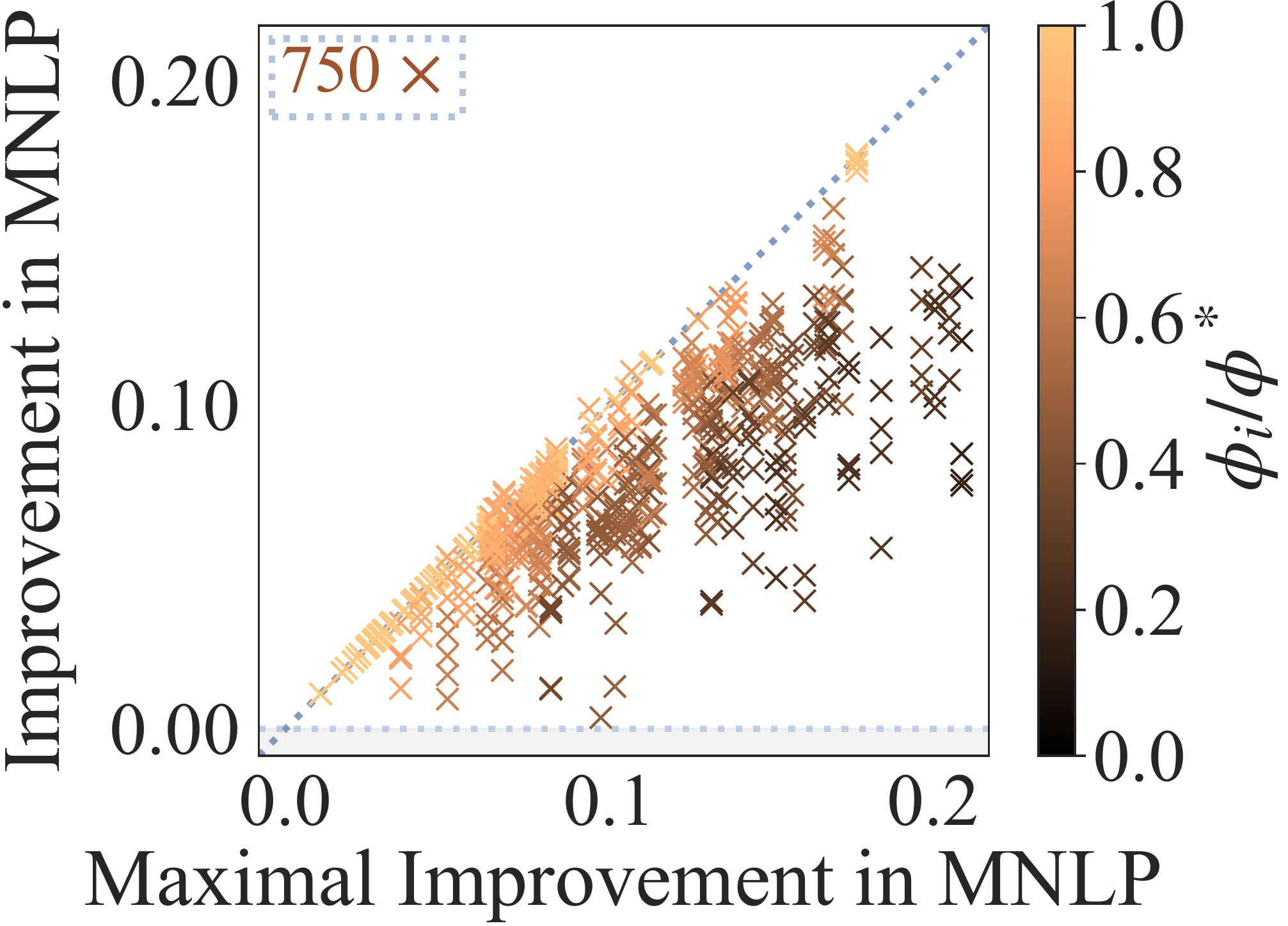} \\
    {(a) MNLP, $\rho = 1$} & {(b) MNLP, $\rho = 0.5$}
    \end{tabular}
    \caption{Scatter graph of the improvement in MNLP vs.~maximal improvement in MNLP when $\rho=1, 0.5$ for multiple partitions of BosH dataset among $n=3$ parties. The prior mean of the model parameters is close to the true mean and the prior variance is small.
    }
    \label{fig:bad-boston}
\end{figure}

The next poor use case is when the model prior, which determines its predictive performance with limited data, is sufficiently informative, hence leading to a high predictive accuracy even if the model is only trained on a small amount of data.
An example of an informative model prior is when the prior mean of the model parameters is close to the true mean and the prior variance is small.
In this case, although the first term of~\eqref{mnlp} is smaller for a model with higher IG, the small predictive variance will magnify the squared error, thus increasing the second term of~\eqref{mnlp}.
So, on the overall, the model with a higher IG may not have a significantly lower MNLP. Instead, it is possible that it has a worse/higher MNLP.
We demonstrate this with the BosH dataset by setting the prior mean of the BLR parameters to be the NN last layer's weights (instead of the zero vector as in Appendix~\ref{ap.bosh}) and prior variance to be a small value of $0.01$ (instead of $1$). The MNLP results are shown in Fig.~\ref{fig:bad-boston}.
It can be observed that the largest maximal improvement in MNLP in Fig.~\ref{fig:bad-boston} is around $0.2$ which is significantly smaller than that (i.e., $1.0$) in Figs.~\ref{fig:boston}c-d. In Fig.~\ref{fig:bad-boston}a, there are more instances with negative improvement in MNLP compared to that in Fig.~\ref{fig:boston}c.

The last poor use case is related to the previous one.
Even if the model prior is insufficiently informative, the phenomenon can also happen if each party has ample data to independently train a model with a reasonable predictive accuracy.
More training data lead to a higher IG and reduce the first term of \eqref{mnlp} but magnify the squared error, thus potentially increasing MNLP.

\textbf{Summary.} 
Ideally, improvement in IG by training on the additional data decided by our reward scheme should translate to a higher predictive accuracy of the model reward.
This should occur when a suitable model is selected and the model prior is not sufficiently informative for any party to achieve a high predictive accuracy by training a model on its own data. In practice, this is reasonable as collaboration precisely happens only when every individual party cannot train a high-quality model alone but can do so from working together.
\subsection{Comparing IG-Based vs.~MNLP-Based Data Valuation Methods}\label{surrogate} 
In Sec.~\ref{MI}, our data valuation method is based on IG
as it can avoid the need and difficulty of selecting a common validation dataset, unlike other data valuation methods (e.g., using MNLP). We also claim that IG is a suitable surrogate measure of the predictive accuracy of a trained model.
In this subsection, we will demonstrate this by comparing the values of data and Shapley values when the data valuation is performed using IG vs.~MNLP.

For the MNLP-based data valuation method, let the value of data $D_C$ for any $C \subseteq N$ be $v'_C = \text{MNLP}_\emptyset - \text{MNLP}_C$ where $\text{MNLP}_C$ is defined in the same way as~\eqref{mnlp} except that $D_i^r$ in~\eqref{mnlp} is replaced by $D_C$.
For each party $i \in N$, we can compute its alternative Shapley value $\phi_i'$ using \eqref{shapley} by replacing $v_C$ with $v'_C$ for all $C \subseteq N$.

\begin{figure}
    \begin{tabular}{@{}c@{\hspace{1.5mm}}c@{}} 
    \includegraphics[width=0.48\columnwidth]{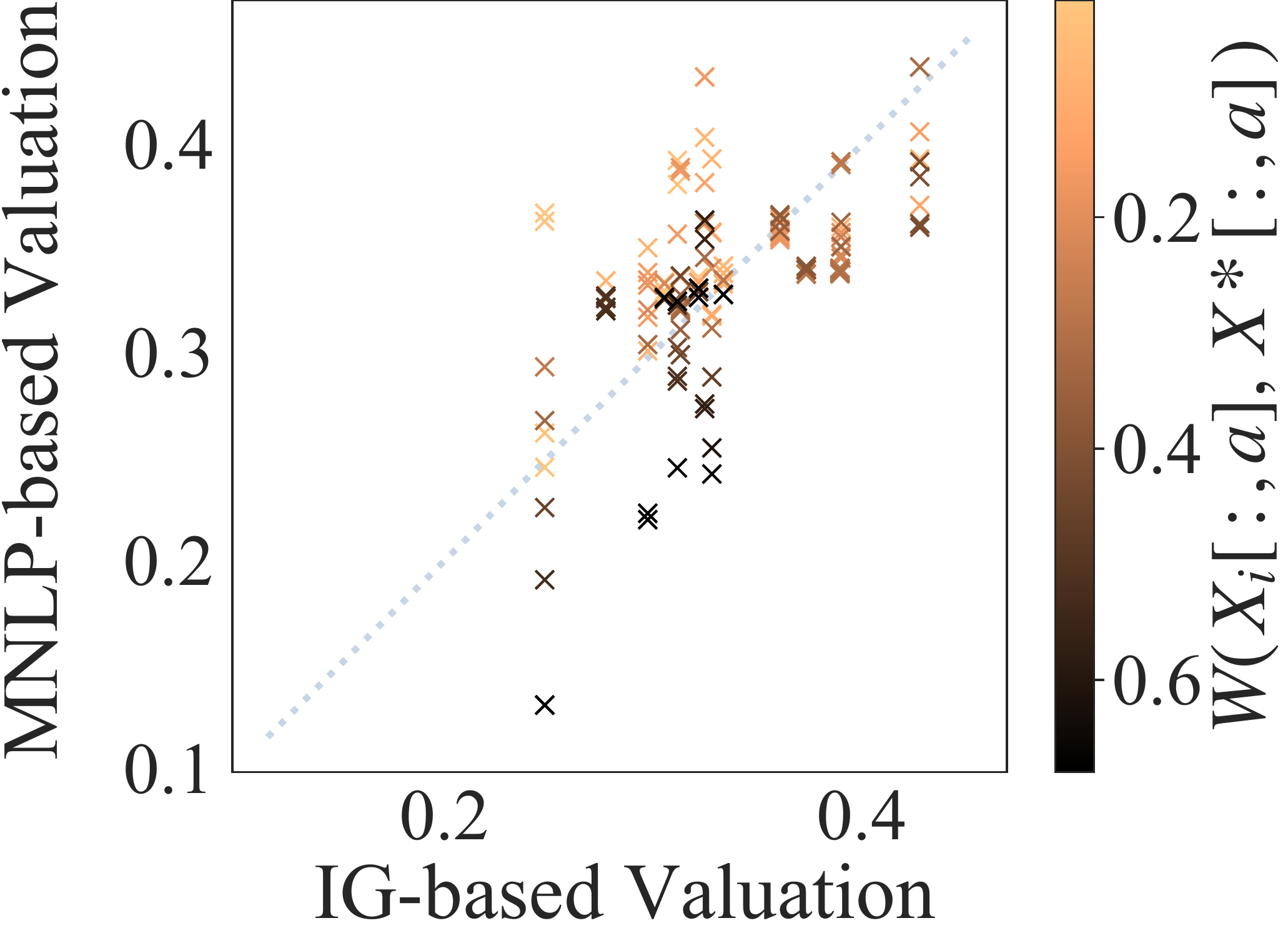} & \includegraphics[width=0.48\columnwidth]{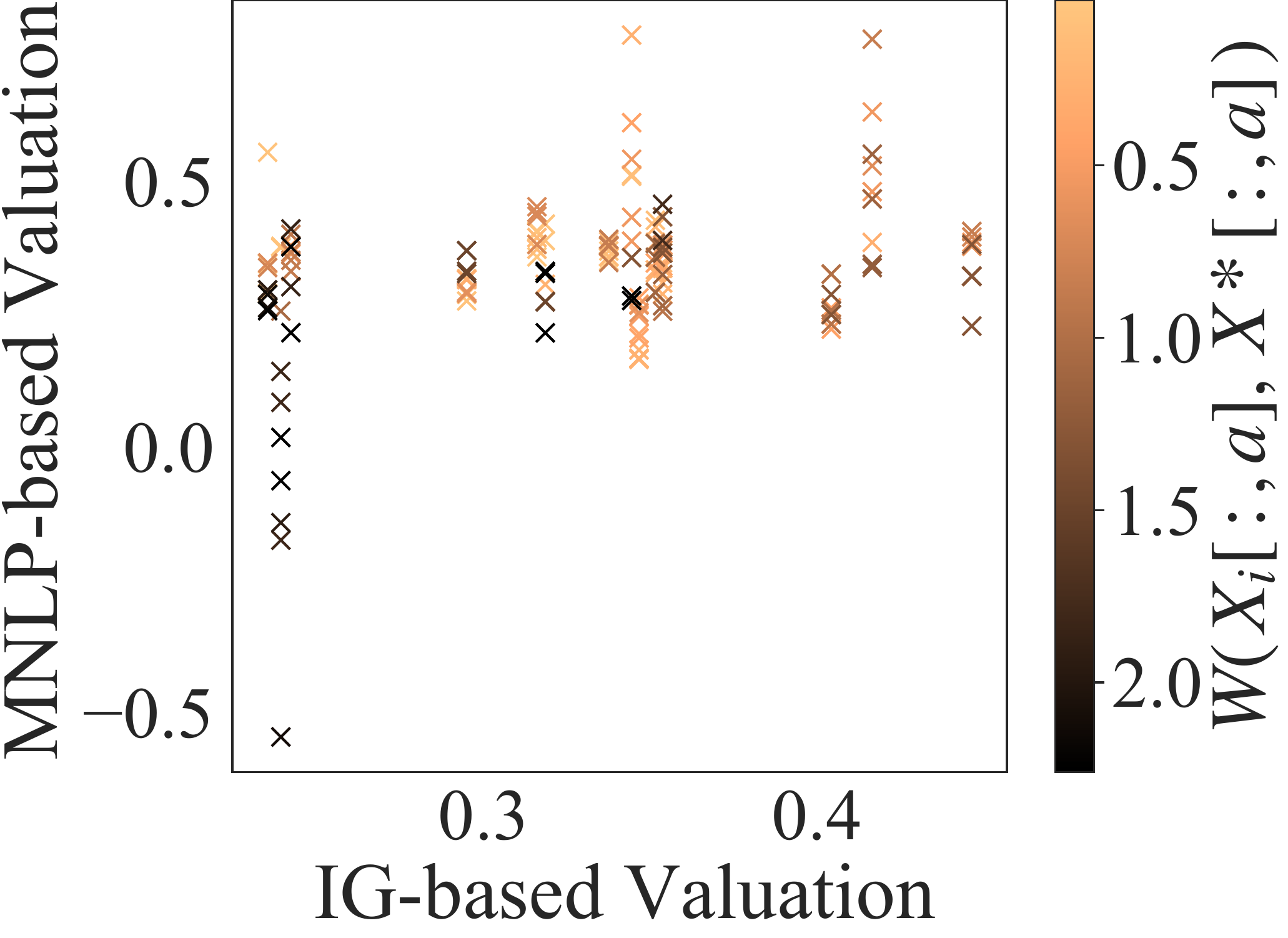} \\
    {(a) GP Friedman} & {(b) NN CaliH} \vspace{1mm}\\
    \includegraphics[width=0.48\columnwidth]{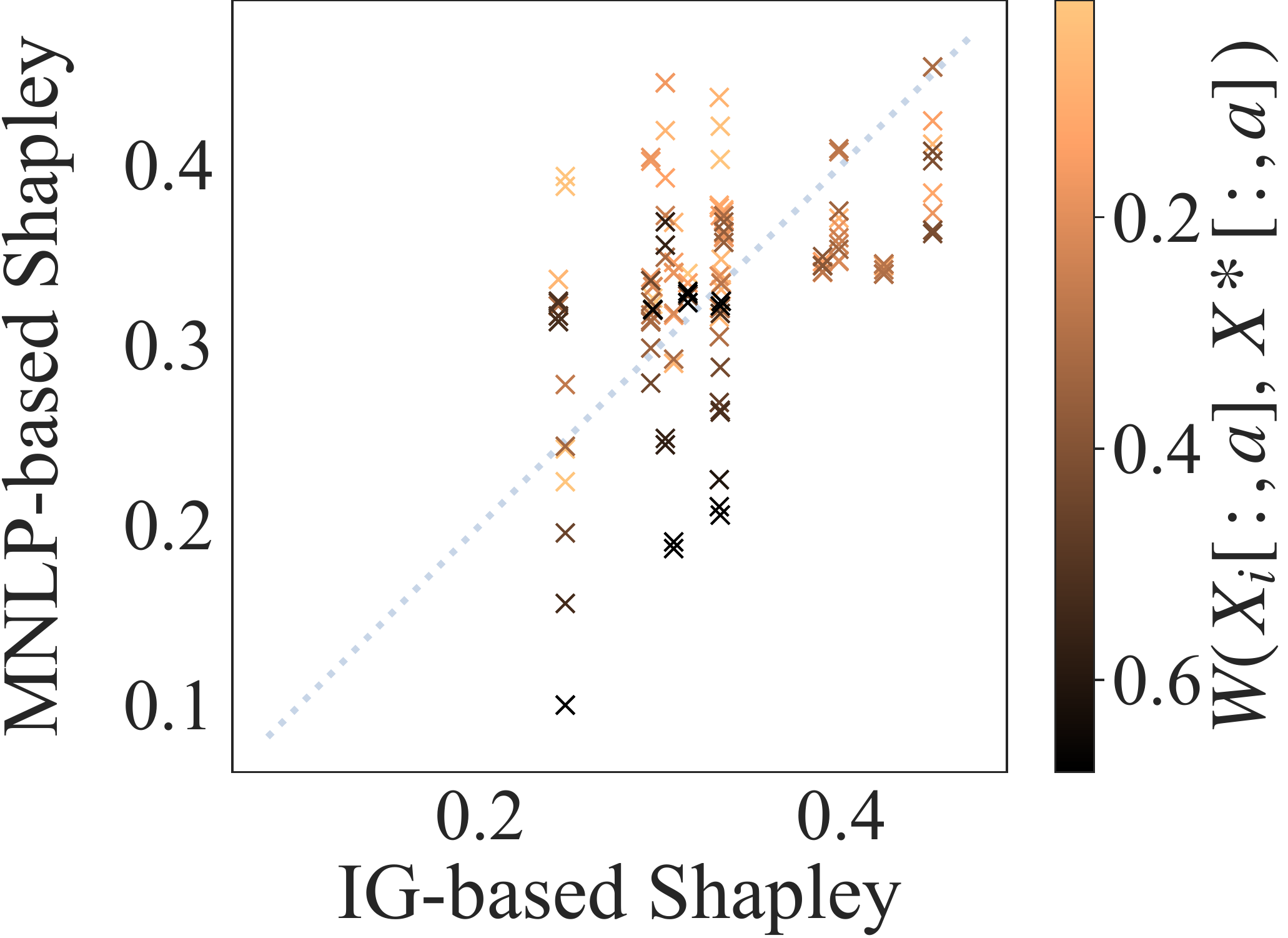} & \includegraphics[width=0.48\columnwidth]{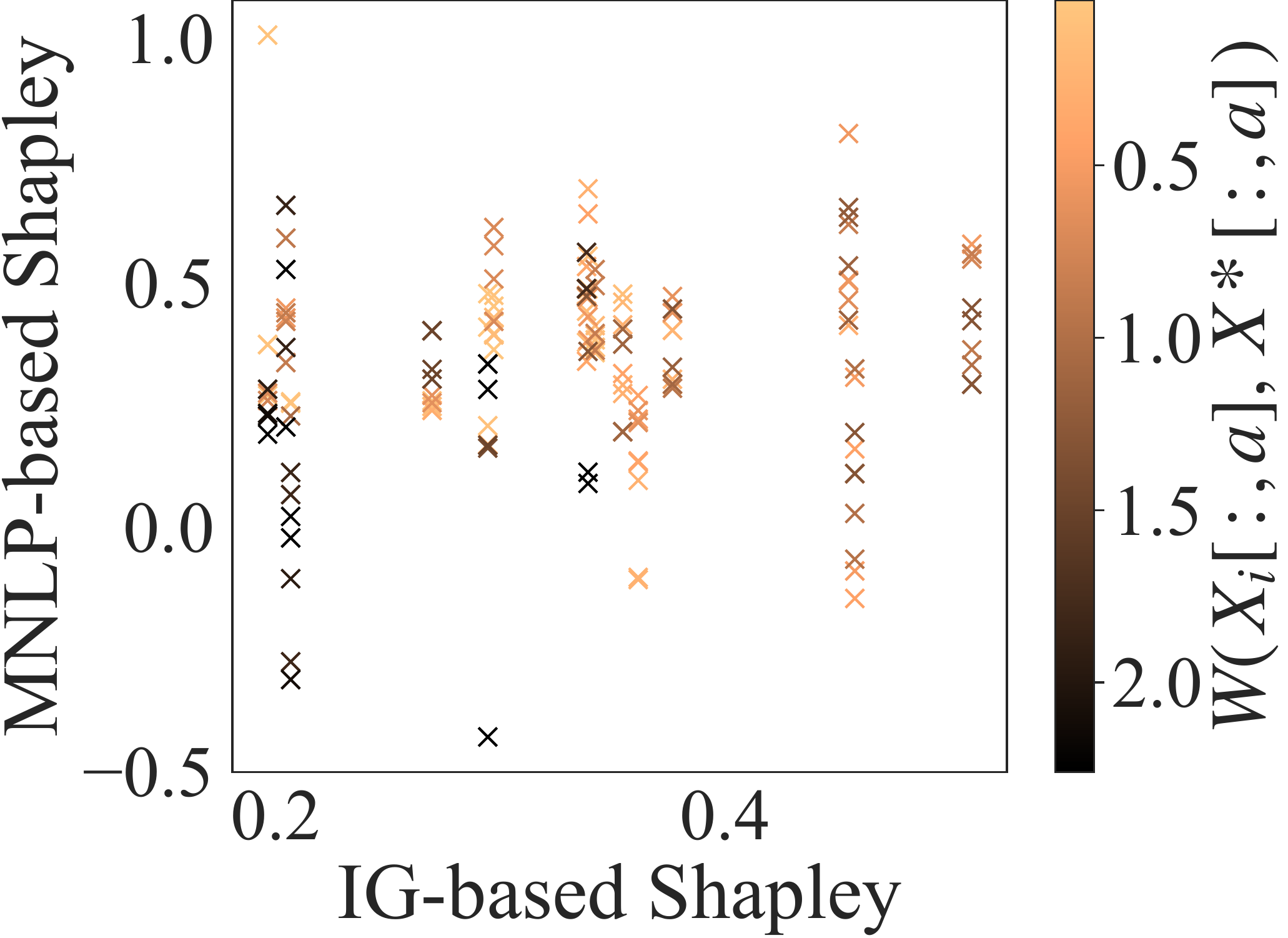} \\
    {(c) GP Friedman} & {(d) NN CaliH}
    \end{tabular}
    \caption{Scatter graph of MNLP-based vs.~IG-based \emph{normalized} (a-b) values of data  and (c-d) Shapley values for Bayesian regression models with various datasets. Each partition and choice of validation dataset generates multiple scatter points (one for each party).
    }
    \label{fig:ig-mnlp-valuation}
\end{figure}

Since we are more concerned about the \emph{relative} values of data and Shapley values between parties, Figs.~\ref{fig:ig-mnlp-valuation}a-b show  results of the IG-based vs.~MNLP-based  \emph{normalized} values of data (respectively, $v_i/\sum_{i\in N}{v_i}$ vs.~$v'_i/\sum_{i\in N}{v'_i}$)
of $5$ different partitions, while
Figs.~\ref{fig:ig-mnlp-valuation}c-d show results of the IG-based vs.~MNLP-based \emph{normalized} Shapley values (respectively, $\phi_i/v_N$ vs.~$\phi'_i/v'_N$) of the same $5$ partitions. 
These partitions are generated using the partitioning algorithm in Sec.~\ref{experi} on a training dataset of $800$ data points.
For each partition, we consider $10$ different validation datasets of size $200$, which results in $10$ possible sets of $\{v'_C\}_{C \in 2^N}$ and $\{\phi'_i\}_{i\in N}$. 
The $10$ validation datasets are either randomly selected or generated using the partitioning algorithm on the larger withheld test dataset (using the same feature $a$) so that it may overlap or exclude any party's data range.
We measure the similarity between the party's data and the validation dataset using the Wasserstein distance $W(\cdot,\cdot)$ between both their values of data or Shapley values for feature $a$. The smaller the distance, the more similar the party's data and the validation dataset.

In Fig.~\ref{fig:ig-mnlp-valuation}, as the validation dataset varies, the normalized values of data and Shapley values for the MNLP-based data valuation method differ significantly (i.e, large spread along the y-axis). This shows that the MNLP-based data valuation method is highly sensitive to the choice of the validation dataset.
Parties are valued significantly lower if their data is very different from the test dataset. This is reflected by the position of the darker colored points below the lighter colored points.

The IG-based data valuation method can serve as a good surrogate for the MNLP-based data valuation method as it will not favor any specific party.
In particular, for the Friedman dataset, the normalized IG-based value of data/Shapley value (i.e., diagonal line) (additionally) closely approximates the normalized MNLP-based values of data/Shapley values over different validation datasets (i.e., scatter points) on average. 



\end{document}